\documentclass[10pt,journal,compsoc]{IEEEtran}

\usepackage[nocompress]{cite}
\usepackage{multirow}
\usepackage[british,UKenglish,USenglish,english,american]{babel}
\usepackage{afterpage}
\usepackage{amsmath}
\usepackage{url}
\usepackage{booktabs} % For formal tables
\usepackage[ruled]{algorithm2e} % For algorithms
\usepackage{subfigure}
\usepackage{tabu}
\usepackage{enumerate}
\usepackage{array}
\usepackage{float}
\usepackage{caption}
\usepackage[nocompress]{cite}
\usepackage[pdftex]{graphicx}
\ifCLASSOPTIONcompsoc
  \usepackage[caption=false,font=normalsize,labelfont=sf,textfont=sf]{subfig}
\else
  \usepackage[caption=false,font=footnotesize]{subfig}
\fi
\usepackage{amsfonts}

\newcolumntype{?}[1]{!{\vrule width #1}}
\newdimen\uulinesep
\newcommand{\etal}{et al. }
\newcommand{\ie}{i.e.}
\newcommand{\eg}{e.g.}
\newcommand{\etc}{etc}

\begin{document}

\title{Neural Style Transfer: A Review}

%\author{
% Yongcheng Jing$^\dagger$ \quad Yezhou Yang$^\ddagger$ \quad Zunlei Feng$^\dagger$ \quad Jingwen Ye$^\dagger$ \quad Mingli Song$^\dagger$\thanks{The corresponding author} \vspace{.5em}\\
% Zhejiang University \vspace{.5em}
%}

\author{Yongcheng~Jing,~%~\IEEEmembership{Member,~IEEE,}
        Yezhou~Yang,~\IEEEmembership{Member,~IEEE,}
        Zunlei~Feng,~%~\IEEEmembership{Member,~IEEE,}
        Jingwen~Ye,~\\%~\IEEEmembership{Member,~IEEE,}\\
        Yizhou~Yu,~\IEEEmembership{Senior~Member,~IEEE,}
        and~Mingli~Song,~\IEEEmembership{Senior~Member,~IEEE}% <-this % stops a space

\IEEEcompsocitemizethanks{
\IEEEcompsocthanksitem \small{Y. Jing, Z. Feng, J. Ye, and M. Song are with Microsoft Visual Perception Laboratory, College of Computer Science and Technology, Zhejiang University, Hangzhou 310027, China. E-mails: \{ycjing, zunleifeng, yejingwen, brooksong\}@zju.edu.cn.}
\IEEEcompsocthanksitem \small{Y. Yang is with School of Computing, Informatics, and Decision Systems Engineering, Arizona State University, Tempe, AZ 85281, USA. E-mail: yz.yang@asu.edu.}% <-this % stops an unwanted space

\IEEEcompsocthanksitem \small{Y. Yu is with the Department of Computer Science, The University of Hong Kong, Pokfulam Road, Hong Kong. E-mail: yizhouy@acm.org.}
}}
%{Shell \MakeLowercase{\textit{et al.}}: Bare Demo of IEEEtran.cls for Computer Society Journals}
\IEEEtitleabstractindextext{%
\begin{abstract}
The seminal work of Gatys et al. demonstrated the power of Convolutional Neural Networks (CNNs) in creating artistic imagery by separating and recombining image content and style. This process of using CNNs to render a content image in different styles is referred to as Neural Style Transfer (NST). Since then, NST has become a trending topic both in academic literature and industrial applications. It is receiving increasing attention and a variety of approaches are proposed to either improve or extend the original NST algorithm. In this paper, we aim to provide a comprehensive overview of the current progress towards NST. We first propose a taxonomy of current algorithms in the field of NST. Then, we present several evaluation methods and compare different NST algorithms both qualitatively and quantitatively. The review concludes with a discussion of various applications of NST and open problems for future research. A list of papers discussed in this review, corresponding codes, pre-trained models and more comparison results are publicly available at: \url{https://github.com/ycjing/Neural-Style-Transfer-Papers}.
\end{abstract}
%, as well as discussing its various applications and open problems for future research
}

\maketitle

\IEEEdisplaynontitleabstractindextext

% \fi
\IEEEpeerreviewmaketitle

\IEEEraisesectionheading{\section{Introduction}\label{sec:intro}}

\IEEEPARstart{P}{ainting} is a popular form of art. For thousands of years, people have been attracted by the art of painting with the advent of many appealing artworks, \eg, van Gogh's ``The Starry Night''. In the past, re-drawing an image in a particular style requires a well-trained artist and lots of time.
%%%%%
%\begin{figure*}
%  \centering
%  \subfigure[Content]{
%    %\label{fig:subfig:a} %% label for first subfigure
%    \includegraphics[width=0.3\textwidth]{figs/content_greatwall.pdf}}
%  \hspace{0.1in}
%  \subfigure[Style]{
%    %\label{fig:subfig:b} %% label for second subfigure
%    \includegraphics[width=0.3\textwidth]{figs/style_fuchun.pdf}}
%    \hspace{0.1in}
%     \subfigure[Content + Style]{
%      \includegraphics[width=0.3\textwidth]{figs/output_fuchun_greatwall.pdf}}
%  \caption{Example of using the Neural Style Transfer algorithm of Gatys \etal to transfer the style of Chinese painting (b) onto The Great Wall photograph (a). The style image is named ``Dwelling in the Fuchun Mountains'' by Gongwang Huang.}
%  \label{fig:exampleGatys} %% label for entire figure
%\end{figure*}
%%%%%
%it is the artist's privilege to create artworks in his happiest mood and condition.

Since the mid-1990s, the art theories behind the appealing artworks have been attracting the attention of not only the artists but many computer science researchers. There are plenty of studies and techniques exploring how to automatically turn images into synthetic artworks. Among these studies, the advances in \emph{non-photorealistic rendering} (NPR) \cite{Gooch2001non, strothotte2002non, rosin2012image} are inspiring, and nowadays, it is a firmly established field in the community of computer graphics. However, most of these NPR stylisation algorithms are designed for particular artistic styles \cite{rosin2012image, gatys2016image} and cannot be easily extended to other styles. In the community of computer vision, style transfer is usually studied as a generalised problem of texture synthesis, which is to extract and transfer the texture from the source to target \cite{efros2001image,drori2003example,frigo2016split,elad2017style}. Hertzmann \etal \cite{hertzmann2001image} further propose a framework named \emph{image analogies} to perform a generalised style transfer by learning the analogous transformation from the provided example pairs of unstylised and stylised images. However, the common limitation of these methods is that they only use low-level image features and often fail to capture image structures effectively.

%However, only low-level features are considered during this process and the results are usually not that impressive.in analogy to

%%%
\begin{figure}
  \centering

  \includegraphics[width=0.48\textwidth]{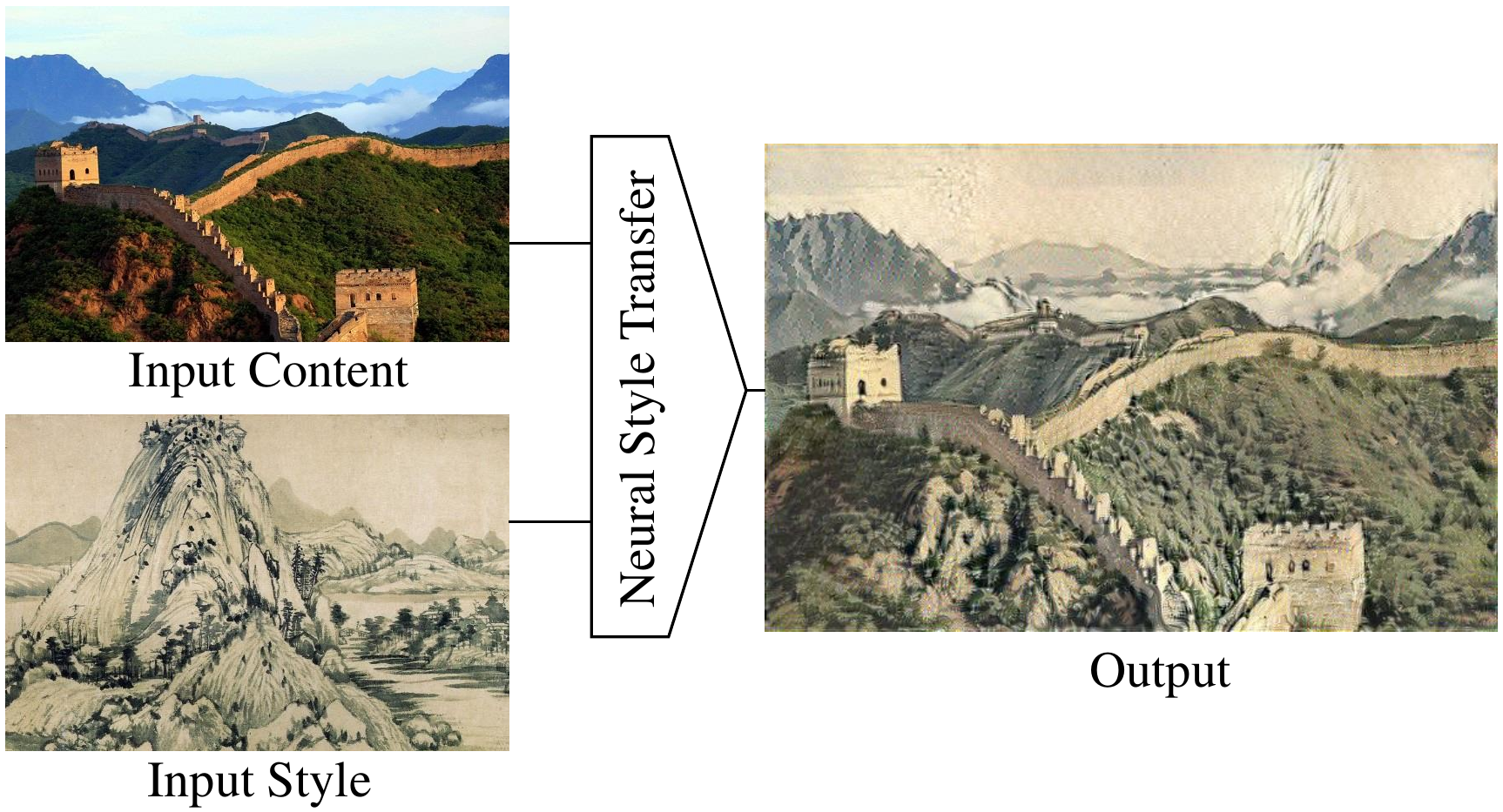}
  \caption{Example of NST algorithm to transfer the style of a Chinese painting onto a given photograph. The style image is named ``Dwelling in the Fuchun Mountains'' by Gongwang Huang.}
  \label{fig:exampleGatys} %% label for entire figure
\end{figure}
%%%

Recently, inspired by the power of \emph{Convolutional Neural Networks (CNNs)}, Gatys \etal \cite{gatys2015neural} first studied how to use a CNN to reproduce famous painting styles on natural images. They proposed to model the \emph{content} of a photo as the feature responses from a pre-trained CNN, and further model the \emph{style} of an artwork as the summary feature statistics. Their experimental results demonstrated that a CNN is capable of extracting \emph{content} information from an arbitrary photograph and \emph{style} information from a well-known artwork. Based on this finding, Gatys \etal \cite{gatys2015neural} first proposed to exploit CNN feature activations to recombine the \emph{content} of a given photo and the \emph{style} of famous artworks. The key idea behind their algorithm is to iteratively optimise an image with the objective of matching desired CNN feature distributions, which involves both the photo's \emph{content} information and artwork's \emph{style} information. Their proposed algorithm successfully produces stylised images with the appearance of a given artwork. Figure~\ref{fig:exampleGatys} shows an example of transferring the style of a Chinese painting ``Dwelling in the Fuchun Mountains'' onto a photo of The Great Wall. Since the algorithm of Gatys \etal does not have any explicit restrictions on the type of style images and also does not need ground truth results for training, it breaks the constraints of previous approaches. The work of Gatys \etal opened up a new field called \emph{Neural Style Transfer (NST)}, which is the process of using \emph{Convolutional Neural Network} to render a content image in different styles.

%the \emph{content} and \emph{style} in a photo were separable, which indicates the probability of changing a photo's \emph{style} while preserving desired \emph{content}.

The seminal work of Gatys \etal has attracted wide attention from both academia and industry. In academia, lots of follow-up studies were conducted to either improve or extend this NST algorithm. The related researches of NST have also led to many successful industrial applications (\eg, \emph{Prisma} \cite{prisma}, \emph{Ostagram} \cite{ostagram}, \emph{Deep Forger} \cite{deepforger}). However, there is no comprehensive survey summarising and discussing recent advances as well as challenges within this new field of Neural Style Transfer.

In this paper, we aim to provide an overview of current advances (up to March 2018) in Neural Style Transfer (NST). Our contributions are threefold. First, we investigate, classify and summarise recent advances in the field of NST. Second, we present several evaluation methods and experimentally compare different NST algorithms. Third, we summarise current challenges in this field and propose possible directions on how to deal with them in future works.

%give an overview of advances

The organisation of this paper is as follows. We start our discussion with a brief review of previous artistic rendering methods without CNNs in Section~\ref{sect:preneural}. Then Section~\ref{sect:derivation} explores the derivations and foundations of NST. Based on the discussions in Section~\ref{sect:derivation}, we categorise and explain existing NST algorithms in Section~\ref{sect:classfication}. Some improvement strategies for these methods and their extensions will be given in Section~\ref{sect:improvementextensions}. Section~\ref{sect:evaluation} presents several methodologies for evaluating NST algorithms and aims to build a standardised benchmark for follow-up studies. Then we demonstrate the commercial applications of NST in Section~\ref{sect:applications}, including both current successful usages and its potential applications. In Section~\ref{sect:challenges}, we summarise current challenges in the field of NST, as well as propose possible directions on how to deal with them in future works. Finally, Section~\ref{sect:conclusion} concludes the paper and delineates several promising directions for future research.

% and Section~\ref{sect:futurework}
%In Section~\ref{sect:loss}, we give an overview of the different style transfer loss functions used by the various techniques.

%%%%%%%%%%%%%%%%%%%%%%%%%%%%
%%%%%%%%%%%%%%%%%%%%%%%%%%%%
%%%%%%%%%%%%%%%%%%%%%%%%%%%%

%%%%

%\begin{figure*}[!t]
%  \centering
%  \captionsetup{justification=centering}
%  \includegraphics[width=\textwidth]{figs/framework_n2.pdf}
%  \centering\caption{A taxonomy of artistic style transfer techniques.}
%  \label{fig:taxonomy} %% label for entire figure
%\end{figure*}

\begin{figure*}[!t]
  %\centering
  %\captionsetup{justification=centering}
  \includegraphics[width=\textwidth]{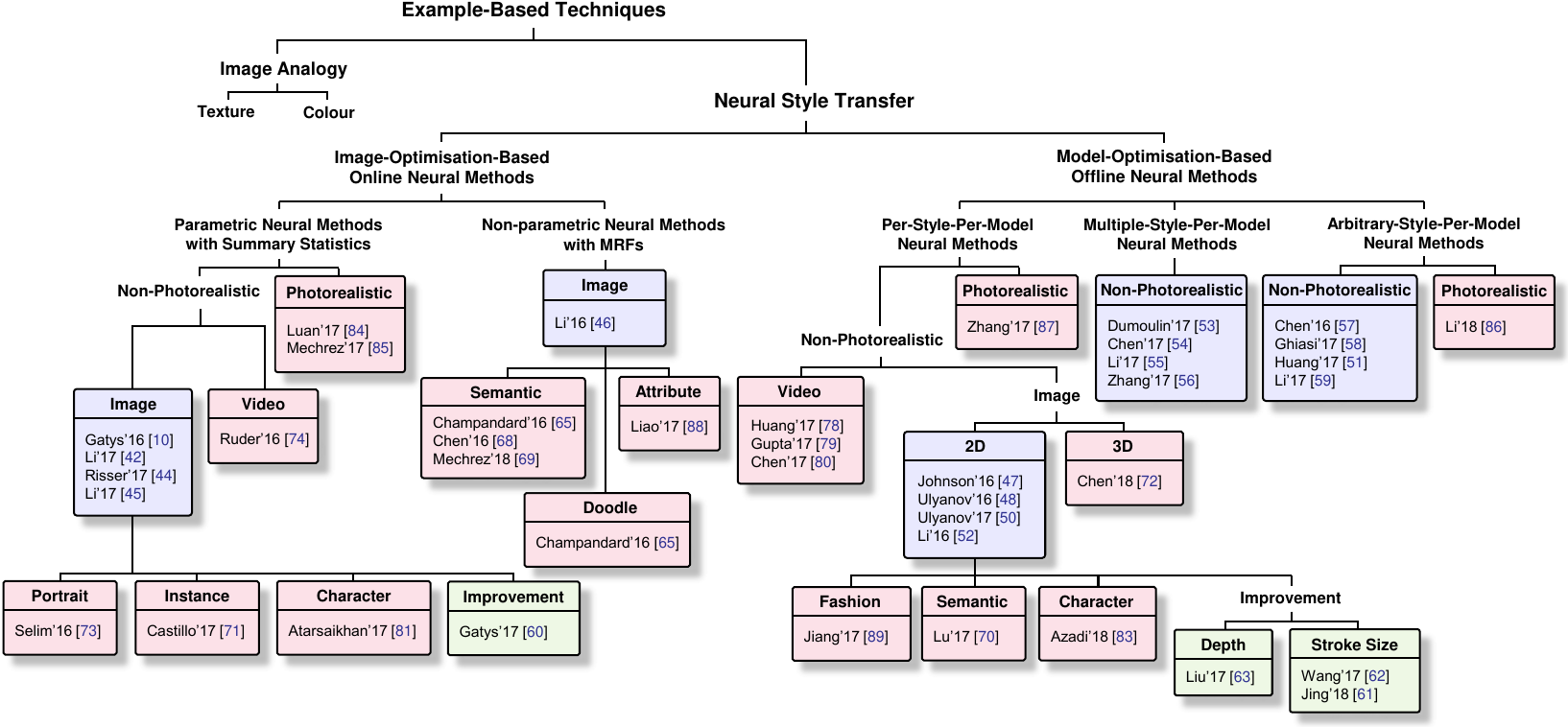}
  \caption{A taxonomy of NST techniques. Our proposed NST taxonomy extends the IB-AR taxonomy proposed by Kyprianidis \etal \cite{kyprianidis2013state}.}
  \label{fig:taxonomy} %% label for entire figure
\end{figure*}

%%%%

\section{Style Transfer Without Neural Networks}
\label{sect:preneural}

Artistic stylisation is a long-standing research topic. Due to its wide variety of applications, it has been an important research area for more than two decades. Before the appearance of NST, the related researches have expanded into an area called \emph{non-photorealistic rendering} (NPR). In this section, we briefly review some of these \emph{artistic rendering} (AR) algorithms without CNNs. Specifically, we focus on artistic stylization of 2D images, which is called \emph{image-based artistic rendering} (IB-AR) in \cite{kyprianidis2013state}. For a more comprehensive overview of IB-AR techniques, we recommend \cite{rosin2012image,kyprianidis2013state,SID17}. Following the IB-AR taxonomy defined by Kyprianidis \etal \cite{kyprianidis2013state}, we first introduce each category of IB-AR techniques without CNNs and then discuss their strengths and weaknesses.

%in computer graphics In the field of computer vision, style transfer is often considered as a generalised problem of texture synthesis.

\textbf{Stroke-Based Rendering.} Stroke-based rendering (SBR) refers to a process of placing virtual strokes (\eg, brush strokes, tiles, stipples) upon a digital canvas to render a photograph with a particular style \cite{hertzmann1998painterly}. The process of SBR is generally starting from a source photo, incrementally compositing strokes to match the photo, and finally producing a non-photorealistic imagery, which looks like the photo but with an artistic style. During this process, an objective function is designed to guide the greedy or iterative placement of strokes.

The goal of SBR algorithms is to faithfully depict a prescribed style. Therefore, they are generally effective at simulating certain types of styles (\eg, oil paintings, watercolours, sketches). However, each SBR algorithm is carefully designed for only one particular style and not capable of simulating an arbitrary style, which is not flexible.

%SBR algorithms are generally effective at simulating certain types of styles, since each SBR algorithm is carefully designed with the goal of faithfully depicting a prescribed style. However, they are not that flexible and each algorithm usually.

%since each SBR algorithm is usually designed only for a particular style (\eg, oil paintings, watercolours, sketches).

%Despite the effectiveness of a large body of SBR algorithms, they are usually designed for a particular style (\eg, oil paintings, watercolours, sketches), which is not that flexible.

\textbf{Region-Based Techniques.} Region-based rendering is to incorporate region segmentation to enable the adaption of rendering based on the content in regions. Early region-based IB-AR algorithms exploit the shape of regions to guide the stroke placement \cite{kolliopoulos2005image,gooch2002artistic}. In this way, different stroke patterns can be produced in different semantic regions in an image. Song \etal \cite{song2008arty} further propose a region-based IB-AR algorithm to manipulate geometry for artistic styles. Their algorithm creates simplified shape rendering effects by replacing regions with several canonical shapes.

Considering regions in rendering allows the local control over the level of details. However, the problem in SBR persists: one region-based rendering algorithm is not capable of simulating an arbitrary style.

\textbf{Example-Based Rendering.} The goal of example-based rendering is to learn the mapping between an exemplar pair. This category of IB-AR techniques is pioneered by Hertzmann et al., who propose a framework named image analogies \cite{hertzmann2001image}. Image analogies aim to learn a mapping between a pair of source images and target stylised images in a supervised manner. The training set of image analogy comprises pairs of unstylised source images and the corresponding stylised images with a particular style. Image analogy algorithm then learns the analogous transformation from the example training pairs and creates analogous stylised results when given a test input photograph. Image analogy can also be extended in various ways, \eg, to learn stroke placements for portrait painting rendering \cite{zhao2011portrait}.

In general, image analogies are effective for a variety of artistic styles. However, pairs of training data are usually unavailable in practice. Another limitation is that image analogies only exploit low-level image features. Therefore, image analogies typically fail to effectively capture content and style, which limits the performance.

\textbf{Image Processing and Filtering.} Creating an artistic image is a process that aims for image simplification and abstraction. Therefore, it is natural to consider adopting and combining some related image processing filters to render a given photo. For example, in \cite{winnemoller2006real}, Winnem{\"o}ller \etal for the first time exploit bilateral \cite{tomasi1998bilateral} and difference of Gaussians filters \cite{gooch2004human} to automatically produce cartoon-like effects.

Compared with other categories of IB-AR techniques, image-filtering based rendering algorithms are generally straightforward to implement and efficient in practice. At an expense, they are very limited in style diversity.

\textbf{Summary.} Based on the above discussions, although some IB-AR algorithms without CNNs are capable of faithfully depicting certain prescribed styles, they typically have the limitations in flexibility, style diversity, and effective image structure extractions. Therefore, there is a demand for novel algorithms to address these limitations, which gives birth to the field of NST.

\section{Derivations of Neural Style Transfer}
\label{sect:derivation}

For a better understanding of the NST development, we start by introducing its derivations. To automatically transfer an artistic style, the first and most important issue is how to model and extract \emph{style} from an image. Since \emph{style} is very related to \emph{texture}\footnote{We clarify that style is very related to texture but not limited to texture. Style also involves a large degree of simplification and shape abstraction effects, which falls back to the composition or alignment of texture features.}, a straightforward way is to relate \emph{Visual Style Modelling} back to previously well-studied \emph{Visual Texture Modelling} methods. After obtaining the style representation, the next issue is how to reconstruct an image with desired style information while preserving its content, which is addressed by the \emph{Image Reconstruction} techniques.

\subsection{Visual Texture Modelling}
\label{sect:texturemodelling}

Visual texture modelling \cite{wei2009state} is previously studied as the heart of texture synthesis \cite{efros1999texture,wei2000fast}. Throughout the history, there are two distinct approaches to model visual textures, which are \emph{Parametric Texture Modelling with Summary Statistics} and \emph{Non-parametric Texture Modelling with Markov Random Fields (MRFs)}.

%\cite{julesz1981textons}\cite{zhu2005textons} \cite{zhu2000exploring} introduce mathematical model of texture.

\textbf{1) Parametric Texture Modelling with Summary Statistics.} One path towards texture modelling is to capture image statistics from a sample texture and exploit summary statistical property to model the texture. The idea is first proposed by Julesz \cite{julesz1962visual}, who models textures as pixel-based $N$-th order statistics. Later, the work in \cite{heeger1995pyramid} exploits filter responses to analyze textures, instead of direct pixel-based measurements. After that, Portilla and Simoncelli \cite{portilla2000parametric} further introduce a texture model based on multi-scale orientated filter responses and use gradient descent to improve synthesised results. A more recent parametric texture modelling approach proposed by Gatys \etal \cite{gatys2015texture} is the first to measure summary statistics in the domain of a CNN. They design a Gram-based representation to model textures, which is the correlations between filter responses in different layers of a pre-trained classification network (VGG network) \cite{simonyan2014very}. More specifically, the Gram-based representation encodes the second order statistics of the set of CNN filter responses. Next, we will explain this representation in detail for the usage of the following sections.

Assume that the feature map of a sample texture image $I_s$ at layer $l$ of a pre-trained deep classification network is $\mathcal{F}^{l}(I_s) \in \mathbb{R}^{C \times H \times W}$, where $C$ is the number of channels, and $H$ and $W$ represent the height and width of the feature map $\mathcal{F}(I_s)$. Then the Gram-based representation can be obtained by computing the Gram matrix $\mathcal{G}(\mathcal{F}^{l}(I_s)') \in \mathbb{R}^{C \times C}$ over the feature map $\mathcal{F}^{l}(I_s)' \in \mathbb{R}^{C \times (HW)}$ (a reshaped version of $\mathcal{F}^{l}(I_s)$):
\begin{equation}
\mathcal{G}(\mathcal{F}^{l}(I_s)')= [\mathcal{F}^{l}(I_s)'] [\mathcal{F}^{l}(I_s)']^T.
\label{eg:gram}
\end{equation}
This Gram-based texture representation from a CNN is effective at modelling wide varieties of both natural and non-natural textures. However, the Gram-based representation is designed to capture global statistics and tosses spatial arrangements, which leads to unsatisfying results for modelling regular textures with long-range symmetric structures. To address this problem, Berger and Memisevic \cite{berger2016incorporating} propose to horizontally and vertically translate feature maps by $\delta$ pixels to correlate the feature at position $(i,j)$ with those at positions $(i+\delta, j)$ and $(i, j+\delta)$. In this way, the representation incorporates spatial arrangement information and is therefore more effective at modelling textures with symmetric properties.

\textbf{2) Non-parametric Texture Modelling with MRFs.} Another notable texture modelling methodology is to use non-parametric resampling. A variety of non-parametric methods are based on MRFs model, which assumes that in a texture image, each pixel is entirely characterised by its spatial neighbourhood. Under this assumption, Efros and Leung \cite{efros1999texture} propose to synthesise each pixel one by one by searching similar neighbourhoods in the source texture image and assigning the corresponding pixel. Their work is one of the earliest non-parametric algorithms with MRFs. Following their work, Wei and Levoy \cite{wei2000fast} further speed up the neighbourhood matching process by always using a fixed neighbourhood.

\subsection{Image Reconstruction}

In general, an essential step for many vision tasks is to extract an abstract representation from the input image. Image reconstruction is a reverse process, which is to reconstruct the whole input image from the extracted image representation. It is previously studied to analyse a particular image representation and discover what information is contained in the abstract representation. Here our major focus is on CNN representation based image reconstruction algorithms, which can be categorised into \emph{Image-Optimisation-Based Online Image Reconstruction} (IOB-IR) and \emph{Model-Optimisation-Based Offline Image Reconstruction} (MOB-IR).

\textbf{1) Image-Optimisation-Based Online Image Reconstruction.} The first algorithm to reverse CNN representations is proposed by Mahendran and Vedaldi \cite{mahendran2015understanding,mahendran2016visualizing}. Given a CNN representation to be reversed, their algorithm iteratively optimises an image (generally starting from random noise) until it has a similar desired CNN representation. The iterative optimisation process is based on gradient descent in image space. Therefore, the process is time-consuming especially when the desired reconstructed image is large.

\textbf{2) Model-Optimisation-Based Offline Image Reconstruction.} To address the efficiency issue of \cite{mahendran2015understanding,mahendran2016visualizing}, Dosovitskiy and Brox \cite{dosovitskiy2016inverting} propose to train a feed-forward network in advance and put the computational burden at training stage. At testing stage, the reverse process can be simply done with a network forward pass. Their algorithm significantly speeds up the image reconstruction process. In their later work \cite{dosovitskiy2016generating}, they further combine Generative Adversarial Network (GAN) \cite{goodfellow2014generative} to improve the results.

%\cite{nguyen2017plug}application generate image \cite{nguyen2016synthesizing}

%%%%%%%%%%%%%%%%%%%%%%%%%%%%
%%%%%%%%%%%%%%%%%%%%%%%%%%%%
%%%%%%%%%%%%%%%%%%%%%%%%%%%%

\section{A Taxonomy of Neural Style Transfer Algorithms}
\label{sect:classfication}
%large body of
% It denotes the group \emph{Style Transfer via Neural Network}. One can also say that NST is a combination of \emph{Style Transfer via Texture Synthesis} and \emph{Convolutional Neural Network}.
NST is a subset of the aforementioned example-based IB-AR techniques. In this section, we first provide a categorisation of NST algorithms and then explain major 2D image based non-photorealistic NST algorithms (Figure~\ref{fig:taxonomy}, purple boxes) in detail. More specifically, for each algorithm, we start by introducing the main idea and then discuss its weaknesses and strengths. Since it is complex to define the notion of style \cite{rosin2012image,xie2007feature} and therefore very subjective to define what criteria are important to make a successful style transfer algorithm \cite{ashikhmin2003fast}, here we try to evaluate these algorithms in a more structural way by only focusing on \emph{details, semantics, depth and variations in brush strokes}\footnote{We claim that the visual criteria with respect to a successful style transfer are definitely not limited to these factors.}. We will discuss more about the problem of aesthetic evaluation criterion in Section~\ref{sect:challenges} and also present more evaluation results in Section~\ref{sect:evaluation}.
% (but definitely not limited to these criterion)

Our proposed taxonomy of NST techniques is shown in Figure~\ref{fig:taxonomy}. We keep the taxonomy of IB-AR techniques proposed by Kyprianidis \etal \cite{kyprianidis2013state} unaffected and extend it by NST algorithms. Current NST methods fit into one of two categories, \emph{Image-Optimisation-Based Online Neural Methods} (IOB-NST) and \emph{Model-Optimisation-Based Offline Neural Methods} (MOB-NST). The first category transfers the style by iteratively optimising an image, \ie, algorithms belong to this category are built upon IOB-IR techniques. The second category optimises a generative model offline and produces the stylised image with a single forward pass, which exploits the idea of MOB-IR techniques.

%\textbf{for inside each category, explain and demonstrate prons and cons in terms of what visual criterion is important}

%Generally speaking, Neural Style Transfer methods can be broadly divided into two groups based on the type of object to update iteratively.

%Detailed descriptions and evaluations of these two groups as well as some improvements and extensions will be given in this paper.

%\stdcomment{For each subsection, first introduce the method in general including motivation, show the mathematical model, and then its pros and cons.}
%
%\myLcomment{The key idea behind style transfer is to perform a gradient descent from random noise minimising the deviation from content of the target image and the deviation from the style representation of the style image. (Exploring the neural algorithm of artistic style)}

%\subsection{Slow Optimisation based stylisation}
%\subsection{Slow Neural Method based on Online Image-Optimisation-Based Online Neural Methods}
\subsection{Image-Optimisation-Based Online Neural Methods}
\label{sect:slowMethod}
%\myScomment{
%\begin{itemize}
%\item Leon A. Gatys 2015- A Neural Algorithm of Artistic Style \cite{gatys2015neural} 1
%  = Leon A. Gatys 2016- Image Style Transfer Using Convolutional Neural Networks- CVPR \cite{gatys2016image} 1
%\item Chuan Li 2016- Combining Markov Random Fields and Convolutional Neural Networks for image synthesis \cite{li2016combining} 2
%\item Champandard 2016- \textbf{Neural Doodle} Semantic Style Transfer and Turning Two-Bit Doodles into Fine Artwork \cite{champandard2016semantic} 2
%\item Yi-Lei Chen 2016- Towards Deep Style Transfer A Content-Aware Perspective- BMVC \cite{chen2016towards} 1
%\item Rujie Yin 2016- Content-Aware Neural Style Transfer \cite{yin2016content} 1
%\item Carlos Castillo 2017- Son of Zorn's Lemma Targeted Style Transfer Using Instance-aware Semantic Segmentation \cite{castillo2017zorn} 1
%\item Yanghao Li 2017- Demystifying Neural Style Transfer \cite{li2017demystifying} 1
%\item Eric Risser 2017- Stable and Controllable Neural Texture Synthesis and Style Transfer Using Histogram Losses \cite{wilmot2017stable}
%\end{itemize}
%}

DeepDream \cite{deepdream} is the first attempt to produce artistic images by reversing CNN representations with IOB-IR techniques. By further combining \emph{Visual Texture Modelling} techniques to model style, IOB-NST algorithms are subsequently proposed, which build the early foundations for the field of NST. Their basic idea is to first model and extract style and content information from the corresponding style and content images, recombine them as the target representation, and then iteratively reconstruct a stylised result that matches the target representation. In general, different IOB-NST algorithms share the same IOB-IR technique, but differ in the way they model the visual style, which is built on the aforementioned two categories of \emph{Visual Texture Modelling} techniques. The common limitation of IOB-NST algorithms is that they are computationally expensive, due to the iterative image optimisation procedure.

%The Descriptive Neural Method is the first proposed neural method to transfer the style between two images. The idea is to update pixels in the (yet unknown) stylised image iteratively through backpropagation, which starts from random noise, until the desired statistics is satisfied. The objective of image iteration is to minimise the total loss such that the content representation of the stylised image matches that of the content image, and meanwhile the style representation matches that of the style image.
%
%One of the keys to Neural Style Transfer is the representation of style, \ie, the pre-defined style loss function. The style loss function is optimised to match the feature statistics of the style image. Depending on the different adopted style loss functions, Descriptive Neural Methods can be further divided into methods based on \emph{Maximum Mean Discrepancy (MMD)} and methods based on \emph{Markov Random Fields (MRF)}. For simplicity, we call them \emph{MMD-based} and \emph{MRF-based} methods.

\subsubsection{Parametric Neural Methods with Summary Statistics}

%The algorithm proposed by Gatys \etal is actually a combination of image reconstruction algorithm (inverting image representations) and texture synthesis method (generating new samples of specific texture), \ie,, the overall procedure of iterating images using gradient descent is similar to that of image reconstruction algorithm proposed by Mahendran \etal while the style representation is inspired by texture synthesis method proposed by Gatys \etal. Given a content image $x$ and style image $y$, Gatys \etal

The first subset of IOB-NST methods is based on \emph{Parametric Texture Modelling with Summary Statistics}. The style is characterised as a set of spatial summary statistics.

We start by introducing the first NST algorithm proposed by Gatys \etal \cite{gatys2015neural,gatys2016image}. By reconstructing representations from intermediate layers of the VGG-19 network, Gatys \etal observe that a deep convolutional neural network is capable of extracting image content from an arbitrary photograph and some appearance information from the well-known artwork. According to this observation, they build the content component of the newly stylised image by penalising the difference of high-level representations derived from content and stylised images, and further build the style component by matching Gram-based summary statistics of style and stylised images, which is derived from their proposed texture modelling technique \cite{gatys2015texture} (Section \ref{sect:texturemodelling}). The details of their algorithm are as follows.

Given a content image $I_c$ and a style image $I_s$, the algorithm in \cite{gatys2016image} tries to seek a stylised image $I$ that minimises the following objective:
\begin{align}
\begin{split}
I^* &= \mathop {\arg \min }\limits_{I} \mathcal{L}_{total}(I_c,I_s,I) \\ & = \mathop {\arg \min }\limits_{I} \ \alpha \mathcal{L}_{c}(I_c,I) + \beta \mathcal{L}_{s}(I_s,I), %+ \gamma \mathcal{L}_{reg}.
\label{eq:totalloss}
\end{split}
\end{align}
where $\mathcal{L}_c$ compares the content representation of a given content image to that of the stylised image, and $\mathcal{L}_s$ compares the Gram-based style representation derived from a style image to that of the stylised image. $\alpha$ and $\beta$ are used to balance the content component and style component in the stylised result.

The content loss $\mathcal{L}_c$ is defined by the squared Euclidean distance between the feature representations $\mathcal{F}^{l}$ of the content image $I_c$ in layer $l$ and that of the stylised image $I$ which is initialised with a noise image:
\begin{equation}
\mathcal{L}_{c} =  \sum\mathop{}_{l \in \{l_c\}}\lVert \mathcal{F}^{l}(I_c) - \mathcal{F}^{l}(I) \rVert^2,
\end{equation}
where $\{l_c\}$ denotes the set of VGG layers for computing the content loss. For the style loss $\mathcal{L}_s$, \cite{gatys2016image} exploits Gram-based visual texture modelling technique to model the style, which has already been explained in Section~\ref{sect:texturemodelling}. Therefore, the style loss is defined by the squared Euclidean distance between the Gram-based style representations of $I_s$ and $I$:
\begin{equation}
\mathcal{L}_s =  \sum\mathop{}_{l \in \{l_s\}}\lVert  \mathcal{G}(\mathcal{F}^{l}(I_s)') - \mathcal{G}(\mathcal{F}^{l}(I)') \rVert^2,
\end{equation}
where $\mathcal{G}$ is the aforementioned Gram matrix to encode the second order statistics of the set of filter responses. $\{l_s\}$ represents the set of VGG layers for calculating the style loss.

The choice of content and style layers is an important factor in the process of style transfer. Different positions and numbers of layers can result in very different visual experiences. Given the pre-trained VGG-19 \cite{simonyan2014very} as the loss network, Gatys et al.'s choice of $\{l_s\}$ and $\{l_c\}$ in \cite{gatys2016image} is $\{l_s\}=\{ relu1\_1, relu2\_1, relu3\_1, relu4\_1, relu5\_1 \}$ and $\{l_c\}=\{relu4\_2\}$. For $\{l_s\}$, the idea of combining multiple layers (up to higher layers) is critical for the success of Gatys et al.'s NST algorithm. Matching the multi-scale style representations leads to a smoother and more continuous stylisation, which gives the visually most appealing results \cite{gatys2016image}. For the content layer $\{l_c\}$, matching the content representations on a lower layer preserves the undesired fine structures (\eg, edges and colour map) of the original content image during stylisation. In contrast, by matching the content on a higher layer of the network, the fine structures can be altered to agree with the desired style while preserving the content information of the content image.
Also, using VGG-based loss networks for style transfer is not the only option. Similar performance can be achieved by selecting other pre-trained classification networks, \eg, ResNet \cite{he2016deep}.
%Comparisons among different layer choices will be demonstrated in the experiment part.
%The choice of $\{l_c\}$ and $\{l_c\}$ empirically follows the principle that the usage of lower layer tends to retain low-level features (\eg, colours), while the usage of higher layer generally preserves more high-level content information. Therefore, $\mathcal{L}_{s}$ is usually computed with lower layers and $\mathcal{L}_{c}$ is computed with higher layers.

In Equation (\ref{eq:totalloss}), both $\mathcal{L}_c$ and $\mathcal{L}_s$ are differentiable. Thus, with random noise as the initial $I$, Equation (\ref{eq:totalloss}) can be minimised by using gradient descent in image space with backpropagation. In addition, a total variation denoising term is usually added in practice to encourage the smoothness in the stylised result.

The algorithm of Gatys \etal does not need ground truth data for training and also does not have explicit restrictions on the type of style images, which addresses the limitations of previous IB-AR algorithms without CNNs (Section~\ref{sect:preneural}). However, the algorithm of Gatys \etal does not perform well in preserving the coherence of fine structures and details during stylisation since CNN features inevitably lose some low-level information. Also, it generally fails for photorealistic synthesis, due to the limitations of Gram-based style representation. Moreover, it does not consider the variations of brush strokes and the semantics and depth information contained in the content image, which are important factors in evaluating the visual quality.

In addition, a Gram-based style representation is not the only choice to statistically encode style information. There are also some other effective statistical style representations, which are derived from a Gram-based representation. Li \etal \cite{li2017demystifying} derive some different style representations by considering style transfer in the domain of transfer learning, or more specifically, \emph{domain adaption} \cite{patel2015visual}. Given that training and testing data are drawn from different distributions, the goal of domain adaption is to adapt a model trained on labelled training data from a source domain to predict labels of unlabelled testing data from a target domain. One way for domain adaption is to match a sample in the source domain to that in the target domain by minimising their distribution discrepancy, in which \emph{Maximum Mean Discrepancy (MMD)} is a popular choice to measure the discrepancy between two distributions. Li \etal prove that matching Gram-based style representations between a pair of style and stylised images is intrinsically minimising MMD with a quadratic polynomial kernel. Therefore, it is expected that other kernel functions for MMD can be equally applied in NST, \eg, the linear kernel, polynomial kernel and Gaussian kernel. Another related representation is the batch normalisation (BN) statistic representation, which is to use mean and variance of the feature maps in VGG layers to model style:
\begin{multline}
\mathcal{L}_s  =  \sum\mathop{}_{l \in \{l_s\}} \frac{1}{C^l}\sum_{c =1}^{C^l} \  \lVert \mu(\mathcal{F}^{l}_{c}(I_s)) - \mu(\mathcal{F}^{l}_{c}(I)) \rVert^2 + \\ \lVert \sigma(\mathcal{F}^{l}_{c}(I_s)) - \sigma(\mathcal{F}^{l}_{c}(I)) \rVert^2,
\end{multline}
where $\mathcal{F}^{l}_{c} \in \mathbb{R}^{H \times W}$ is the $c$-th feature map channel at layer $l$ of VGG network, and $C^l$ is the number of channels.

The main contribution of Li et al.'s algorithm is to theoretically demonstrate that the Gram matrices matching process in NST is equivalent to minimising MMD with the second order polynomial kernel, thus proposing a timely interpretation of NST and making the principle of NST clearer. However, the algorithm of Li \etal does not resolve the aforementioned limitations of Gatys et al.'s algorithm.

One limitation of the Gram-based algorithm is its instabilities during optimisations. Also, it requires manually tuning the parameters, which is very tedious. Risser \etal \cite{wilmot2017stable} find that feature activations with quite different means and variances can still have the same Gram matrix, which is the main reason of instabilities. Inspired by this observation, Risser \etal introduce an extra histogram loss, which guides the optimisation to match the entire histogram of feature activations. They also present a preliminary solution to automatic parameter tuning, which is to explicitly prevent gradients with extreme values through extreme gradient normalisation.

By additionally matching the histogram of feature activations, the algorithm of Risser \etal achieves a more stable style transfer with fewer iterations and parameter tuning efforts. However, its benefit comes at an expense of a high computational complexity. Also, the aforementioned weaknesses of Gatys et al.'s algorithm still exist, \eg, a lack of consideration in depth and the coherence of details.

%argue that the cause of instabilities is that Gram matrices are actually built on feature activations rather than image intensities. But feature activations with quite different means and variances are shown to still have the same Gram matrix (Figure 4 in \cite{wilmot2017stable}).

All these aforementioned neural methods only compare content and stylised images in the CNN feature space to make the stylised image semantically similar to the content image. But since CNN features inevitably lose some low-level information contained in the image, there are usually some unappealing distorted structures and irregular artefacts in the stylised results. To preserve the coherence of fine structures during stylisation, Li \etal \cite{li2017laplacian} propose to incorporate additional constraints upon low-level features in pixel space. They introduce an additional Laplacian loss, which is defined as the squared Euclidean distance between the Laplacian filter responses of a content image and stylised result. Laplacian filter computes the second order derivatives of the pixels in an image and is widely used for edge detection.

The algorithm of Li \etal has a good performance in preserving the fine structures and details during stylisation. But it still lacks considerations in semantics, depth, variations in brush strokes, \etc.

\subsubsection{Non-parametric Neural Methods with MRFs}
\label{sect:MRF}

Non-parametric IOB-NST is built on the basis of \emph{Non-parametric Texture Modelling with MRFs}. This category considers NST at a local level, \ie, operating on patches to match the style.

Li and Wand \cite{li2016combining} are the first to propose an MRF-based NST algorithm. They find that the parametric NST method with summary statistics only captures the per-pixel feature correlations and does not constrain the spatial layout, which leads to a less visually plausible result for photorealistic styles. Their solution is to model the style in a non-parametric way and introduce a new style loss function which includes a patch-based MRF prior:
\begin{equation}
\mathcal{L}_s = \sum \mathop{}_{l \in \{l_s\}} \sum\limits_{i = 1}^m \lVert {\Psi _i}(\mathcal{F}^{l} (I)) - {\Psi _{NN(i)}}(\mathcal{F}^{l} (I_s)) \rVert^2,
\end{equation}
where $\Psi(\mathcal{F}^l(I))$ is the set of all local patches from the feature map $\mathcal{F}^l(I)$. $\Psi _i$ denotes the $i^{th}$ local patch and $\Psi _{NN(i)}$ is the most similar style patch with the $i$-th local patch in the stylised image $I$. The best matching $\Psi _{NN(i)}$ is obtained by calculating normalised cross-correlation over all style patches in the style image $I_s$. $m$ is the total number of local patches. Since their algorithm matches a style in the patch-level, the fine structure and arrangement can be preserved much better.

The advantage of the algorithm of Li and Wand is that it performs especially well for photorealistic styles, or more specifically, when the content photo and the style are similar in shape and perspective, due to the patch-based MRF loss. However, it generally fails when the content and style images have strong differences in perspective and structure since the image patches could not be correctly matched. It is also limited in preserving sharp details and depth information.

%Given a photograph as the content, their algorithm achieves superior results, especially for photorealistic styles.
%
%Their design of content-aware transfer makes the results remarkable for photorealistic styles.

%The algorithm of Li and Wand is actually a content-aware Neural Style Transfer algorithm which, in a sense, considers the semantic content (Figure 7 in \cite{li2016combining}). Their design of content-aware transfer makes the results remarkable for photorealistic styles.

%\stdcomment{\textbf{page 2 & page 3}}
%\subsection{Fast feed-forward stylisation}
%\subsection{Fast Neural Method based on Offline Model Optimisation}
\subsection{Model-Optimisation-Based Offline Neural Methods}
\label{sect:fastMethod}

Although IOB-NST is able to yield impressive stylised images, there are still some limitations. The most concerned limitation is the efficiency issue. The second category MOB-NST addresses the speed and computational cost issue by exploiting MOB-IR to reconstruct the stylised result, \ie, a feed-forward network $g$ is optimised over a large set of images $I_c$ for one or more style images $I_s$:
%\begin{equation}
%g(I_c)=I, I \sim p(I| I_c, I_s)
%\end{equation}
\begin{equation}
\theta^*=\mathop {\arg \min }\limits_{\theta} \mathcal{L}_{total}(I_c,I_s,g_{\theta^*}(I_c)), \ I^* = g_{\theta^*}(I_c).
\label{eq:fastloss}
\end{equation}
Depending on the number of artistic styles a single $g$ can produce, MOB-NST algorithms are further divided into \emph{Per-Style-Per-Model} (PSPM) MOB-NST methods , \emph{Multiple-Style-Per-Model} (MSPM) MOB-NST Methods, and \emph{Arbitrary-Style-Per-Model} (ASPM) MOB-NST Methods.

\subsubsection{Per-Style-Per-Model Neural Methods}
\label{sect:PSPM}
\textbf{1) Parametric PSPM with Summary Statistics.} The first two MOB-NST algorithms are proposed by Johnson \etal \cite{Johnson2016perceptual} and Ulyanov \etal \cite{ulyanov2016texture} respectively. These two methods share a similar idea, which is to pre-train a feed-forward style-specific network and produce a stylised result with a single forward pass at testing stage. They only differ in the network architecture, for which Johnson \etal's design roughly follows the network proposed by Radford \etal \cite{radford2015unsupervised} but with residual blocks as well as fractionally strided convolutions, and Ulyanov \etal use a multi-scale architecture as the generator network. The objective function is similar to the algorithm of Gatys \etal \cite{gatys2016image}, which indicates that they are also \emph{Parametric Methods with Summary Statistics}.

The algorithms of Johnson \etal and Ulyanov \etal achieve a real-time style transfer. However, their algorithm design basically follows the algorithm of Gatys et al., which makes them suffer from the same aforementioned issues as Gatys et al.'s algorithm (\eg, a lack of consideration in the coherence of details and depth information).

Shortly after \cite{Johnson2016perceptual,ulyanov2016texture}, Ulyanov \etal \cite{ulyanov2017improved} further find that simply applying normalisation to every single image rather than a batch of images (precisely \emph{batch normalization (BN)}) leads to a significant improvement in stylisation quality. This single image normalisation is called \emph{instance normalisation} (IN), which is equivalent to batch normalisation when the batch size is set to $1$. The style transfer network with IN is shown to converge faster than BN and also achieves visually better results. One interpretation is that IN is a form of style normalisation and can directly normalise the style of each content image to the desired style \cite{huang2017arbitrary}. Therefore, the objective is easier to learn as the rest of the network only needs to take care of the content loss.

%Johnson \etal \cite{Johnson2016perceptual} introduce a fast approach based on the algorithm proposed by Gatys \etal They first train an equivalent feed-forward generator network for each specific style. When there is a content image to be stylised, the result can be produced by just a single forward pass. Their presented system has two components, \ie, image generator network and loss network (Figure 2 in \cite{Johnson2016perceptual}). The architecture of generator network  Rather than defining a per-pixel loss function as many image transformation algorithms do, they propose a concept of perceptual loss function or so-called loss network, which are themselves deep convolutional neural network (VGG-16 network is chosen in \cite{Johnson2016perceptual}). The objective function is similar to the algorithm of Gatys et al. but with an extra denoising term.

%Another concurrent work of fast generative method is the texture network proposed by Ulyanov \etal \cite{ulyanov2016texture}. The key idea behind Ulyanov et al.'s algorithm is similar to Johnson et al.'s algorithm, except for the architecture of generator network. Ulyanov \etal use a multi-scale architecture as the generator network.

\textbf{2) Non-parametric PSPM with MRFs.} Another work by Li and Wand \cite{li2016precomputed} is inspired by the MRF-based NST \cite{li2016combining} algorithm in Section~\ref{sect:MRF}. They address the efficiency issue by training a Markovian feed-forward network using adversarial training. Similar to \cite{li2016combining}, their algorithm is a \emph{Patch-based Non-parametric Method with MRFs}. Their method is shown to outperform the algorithms of Johnson \etal and Ulyanov \etal in the preservation of coherent textures in complex images, thanks to their patch-based design. However, their algorithm has a less satisfying performance with non-texture styles (\eg, face images), since their algorithm lacks a consideration in semantics. Other weaknesses of their algorithm include a lack of consideration in depth information and variations of brush strokes, which are important visual factors.

%The algorithm of Li and Wand \cite{li2016precomputed} basically has the aforementioned benefits of their previous work in \cite{li2016combining} with their proposed deep Markovian model and they further speed up the style transfer process in \cite{li2016combining}.

\subsubsection{Multiple-Style-Per-Model Neural Methods}
\label{sect:MSPM}
Although the above PSPM approaches can produce stylised images two orders of magnitude faster than previous IOB-NST methods, separate generative networks have to be trained for each particular style image, which is quite time-consuming and inflexible. But many paintings (\eg, impressionist paintings) share similar paint strokes and only differ in their colour palettes. Intuitively, it is redundant to train a separate network for each of them. MSPM is therefore proposed, which improves the flexibility of PSPM by further incorporating multiple styles into one single model. There are generally two paths towards handling this problem: 1) tying only a small number of parameters in a network to each style (\cite{dumoulin2016learned,chen2017stylebank}) and 2) still exploiting only a single network like PSPM but combining both style and content as inputs (\cite{li2017diverse,zhang2017multi}).
%many style images may share some computations and

\textbf{1) Tying only a small number of parameters to each style.} An early work by Dumoulin \etal \cite{dumoulin2016learned} is built on the basis of the proposed IN layer in PSPM algorithm \cite{ulyanov2017improved} (Section~\ref{sect:PSPM}). They surprisingly find that using the same convolutional parameters but only scaling and shifting parameters in IN layers is sufficient to model different styles. Therefore, they propose an algorithm to train a conditional multi-style transfer network based on conditional instance normalisation (CIN), which is defined as:
\begin{equation}
\textrm{CIN}(\mathcal{F}(I_c), s)= \gamma^{s}\left(\frac{\mathcal{F}(I_c)-\mu(\mathcal{F}(I_c))}{\sigma(\mathcal{F}(I_c))}\right)+\beta^{s},
\label{eq:CIN}
\end{equation}
where $\mathcal{F}$ is the input feature activation and $s$ is the index of the desired style from a set of style images. As shown in Equation (\ref{eq:CIN}), the conditioning for each style $I_s$ is done by scaling and shifting parameters $\gamma^s$ and $\beta^s$ after normalising feature activation $\mathcal{F}(I_c)$, \ie, each style $I_s$ can be achieved by tuning parameters of an affine transformation. The interpretation is similar to that for \cite{ulyanov2017improved} in Section~\ref{sect:PSPM}, \ie, the normalisation of feature statistics with different affine parameters can normalise input content image to different styles. Furthermore, the algorithm of Dumoulin \etal can also be extended to combine multiple styles in a single stylised result by combining affine parameters of different styles.

Another algorithm which follows the first path of MSPM is proposed by Chen \etal \cite{chen2017stylebank}. Their idea is to explicitly decouple style and content, \ie, using separate network components to learn the corresponding content and style information. More specifically, they use mid-level convolutional filters (called ``StyleBank'' layer) to individually learn different styles. Each style is tied to a set of parameters in ``StyleBank'' layer. The rest components in the network are used to learn content information, which is shared by different styles. Their algorithm also supports flexible incremental training, which is to fix the content components in the network and only train a ``StyleBank'' layer for a new style.

In summary, both the algorithms of Dumoulin \etal and Chen \etal have the benefits of little efforts needed to learn a new style and a flexible control over style fusion. However, they do not address the common limitations of NST algorithms, \eg, a lack of details, semantics, depth and variations in brush strokes.

\textbf{2) Combining both style and content as inputs.} One disadvantage of the first category is that the model size generally becomes larger with the increase of the number of learned styles. The second path of MSPM addresses this limitation by fully exploring the capability of one single network and combining both content and style into the network for style identification. Different MSPM algorithms differ in the way to incorporate style into the network.

In \cite{li2017diverse}, given $N$ target styles, Li \etal design a selection unit for style selection, which is a $N$-dimensional one-hot vector. Each bit in the selection unit represents a specific style $I_s$ in the set of target styles. For each bit in the selection unit, Li \etal first sample a corresponding noise map $f(I_s)$ from a uniform distribution and then feed $f(I_s)$ into the style sub-network to obtain the corresponding style encoded features $\mathcal{F}(f(I_s))$. By feeding the concatenation of the style encoded features $\mathcal{F}(f(I_s))$ and the content encoded features $Enc(I_c)$ into the decoder part $Dec$ of the style transfer network, the desired stylised result can be produced: $I=Dec( \ \mathcal{F}(f(I_s)) \ \oplus \ Enc(I_c) \ )$.

%with a convolutional layer
%
%style sub-network which is responsible for style selection and style encoding. The style sub-network takes a selection unit as input, which is a one-hot vector. Each bit in the selection unit represents a specific style in the target style set. For each bit in the selection unit (\ie, each style $I_s$), Li \etal sample a corresponding noise map $f(I_s)$ from a uniform distribution. Then,  the style sub-network as style encoded feature activations $\mathcal{F}(f(I_s))$  The set of noise maps are then

%Given $N$ target styles, Li \etal \cite{li2017diverse} propose to first sample a set of noise maps from a uniform distribution and establish a one-to-one mapping $f$ between each style and noise map. For clarity, we divide the style transfer network into an encoder ($Enc$) and decoder ($Dec$) pair. For each style, the corresponding noise map $f(I_s)$ is concatenated ($\oplus$) with the encoded feature activations $Enc(I_c)$ and then feeded into the decoder to get the stylised result: $I=Dec( \ f(I_s) \ \oplus \ Enc(I_c) \ )$.

Another work by Zhang and Dana \cite{zhang2017multi} first forwards each style image in the style set through the pre-trained VGG network and obtain multi-scale feature activations $\mathcal{F}(I_s)$ in different VGG layers. Then multi-scale $\mathcal{F}(I_s)$ are combined with multi-scale encoded features $Enc(I_c)$ from different layers in the encoder through their proposed inspiration layers. The inspiration layers are designed to reshape $\mathcal{F}(I_s)$ to match the desired dimension, and also have a learnable weight matrix to tune feature maps to help minimise the objective function.

The second type of MSPM addresses the limitation of the increased model size in the first type of MSPM. At an expense, the style scalability of the second type of MSPM is much smaller, since only one single network is used for multiple styles. We will quantitatively compare the style scalability of different MSPM algorithms in Section~\ref{sect:evaluation}. In addition, some aforementioned limitations in the first type of MSPM still exist, \ie, the second type of MSPM algorithms are still limited in preserving the coherence of fine structures and also depth information.

\subsubsection{Arbitrary-Style-Per-Model Neural Methods}
%\myScomment{
%\begin{itemize}
%%  \item Justin Johnson 2016- Perceptual Losses for Real-Time Style Transfer and Super-Resolution- ECCV \cite{Johnson2016perceptual}
%%  \item Dmitry Ulyanov 2016- Texture Networks Feed-forward Synthesis of Textures and stylised Images- ICML \cite{ulyanov2016texture}
%%  \item Chuan Li 2016- Precomputed Real-Time Texture Synthesis with Markovian Generative Adversarial Networks \cite{li2016precomputed}
%%  \item Vincent Dumoulin 2016- A Learned Representation for Artistic Style \cite{dumoulin2016learned}
%\item Tian Qi Chen 2016- Fast Patch-based Style Transfer of Arbitrary Style \cite{chen2016fast}
%\end{itemize}
%}

The third category, ASPM-MOB-NST, aims at one-model-for-all, \ie, one single trainable model to transfer arbitrary artistic styles. There are also two types of ASPM, one built upon \emph{Non-parametric Texture Modelling with MRFs} and the other one built upon \emph{Parametric Texture Modelling with Summary Statistics}.

\textbf{1) Non-parametric ASPM with MRFs.}
The first ASPM algorithm is proposed by Chen and Schmidt \cite{chen2016fast}. They first extract a set of activation patches from content and style feature activations computed in pre-trained VGG network. Then they match each content patch to the most similar style patch and swap them (called ``Style Swap'' in \cite{chen2016fast}). The stylised result can be produced by reconstructing the resulting activation map after ``Style Swap'', with either IOB-IR or MOB-IR techniques. The algorithm of Chen and Schmidt is more flexible than the previous approaches due to its characteristic of one-model-for-all-style. But the stylised results of \cite{chen2016fast} are less appealing since the content patches are typically swapped with the style patches which are not representative of the desired style. As a result, the content is well preserved while the style is generally not well reflected.

%Addressing the same problem as \cite{dumoulin2016learned}, very recently Chen and Schmidt \cite{chen2016fast} propose a fast patch-based style transfer algorithm. They first propose a fast method based on image iteration in their paper and then, based on the proposed method, they train an inverse network to further speed up this process. For their method based on image iteration, they first extract a set of patches for both content image and style image and swap the activation of each content patch with its closet style patch, which is the process of so-called ``swapping the style'' in their paper. The activation is constructed in a single layer and thus the computation complexity is reduced and the process is faster. Then based on above algorithm, Chen and Schmidt further train an inverse network to directly invert the swapped activation. Compared with \cite{Johnson2016perceptual,ulyanov2016texture,li2016precomputed}, their algorithm is capable of producing stylised image for any new style images with only a single trained inverse network.

\textbf{2) Parametric ASPM with Summary Statistics.} Considering \cite{dumoulin2016learned} in Section~\ref{sect:MSPM}, the simplest approach for arbitrary style transfer is to train a separate parameter prediction network $P$ to predict $\gamma^s$ and $\beta^s$ in Equation (\ref{eq:CIN}) with a number of training styles \cite{ghiasi2017exploring}. Given a test style image $I_s$, CIN layers in the style transfer network take affine parameters $\gamma^s$ and $\beta^s$ from $P(I_s)$, and normalise the input content image to the desired style with a forward pass.

Another similar approach based on \cite{dumoulin2016learned} is proposed by Huang and Belongie \cite{huang2017arbitrary}. Instead of training a parameter prediction network, Huang and Belongie propose to modify conditional instance normalisation (CIN) in Equation (\ref{eq:CIN}) to adaptive instance normalisation (AdaIN):
\begin{multline}
\textrm{AdaIN}(\mathcal{F}(I_c), \mathcal{F}(I_s))= \\ \sigma(\mathcal{F}(I_s))\left(\frac{\mathcal{F}(I_c)-\mu(\mathcal{F}(I_c))}{\sigma(\mathcal{F}(I_c))}\right)+  \mu(\mathcal{F}(I_s)).
\label{eq:AdaIN}
\end{multline}
AdaIN transfers the channel-wise mean and variance feature statistics between content and style feature activations, which also shares a similar idea with \cite{chen2016fast}. Different from \cite{dumoulin2016learned}, the encoder in the style transfer network of \cite{huang2017arbitrary} is fixed and comprises the first few layers in pre-trained VGG network. Therefore, $\mathcal{F}$ in \cite{huang2017arbitrary} is the feature activation from a pre-trained VGG network. The decoder part needs to be trained with a large set of style and content images to decode resulting feature activations after AdaIN to the stylised result: $I=Dec( \ \textrm{AdaIN}(\mathcal{F}(I_c), \mathcal{F}(I_s)) \ )$.

The algorithm of Huang and Belongie \cite{huang2017arbitrary} is the first ASPM algorithm that achieves a real-time stylisation. However, the algorithm of Huang and Belongie \cite{huang2017arbitrary} is data-driven and limited in generalising on unseen styles. Also, simply adjusting the mean and variance of feature statistics makes it hard to synthesise complicated style patterns with rich details and local structures.

A more recent work by Li \etal \cite{li2017universal} attempts to exploit a series of feature transformations to transfer arbitrary artistic style in a style learning free manner. Similar to \cite{huang2017arbitrary}, Li \etal use the first few layers of pre-trained VGG as the encoder and train the corresponding decoder. But they replace the AdaIN layer \cite{huang2017arbitrary} in between the encoder and decoder with a pair of whitening and colouring transformations (WCT): $I=Dec( \ \textrm{WCT}(\mathcal{F}(I_c), \mathcal{F}(I_s)) \ )$.  Their algorithm is built on the observation that the whitening transformation can remove the style related information and preserve the structure of content. Therefore, receiving content activations $\mathcal{F}(I_c)$ from the encoder, whitening transformation can filter the original style out of the input content image and return a filtered representation with only content information. Then, by applying colouring transformation, the style patterns contained in $\mathcal{F}(I_s)$ are incorporated into the filtered content representation, and the stylised result $I$ can be obtained by decoding the transformed features. They also extend this single-level stylisation to multi-level stylisation to further improve visual quality.

The algorithm of Li \etal is the first ASPM algorithm to transfer artistic styles in a learning-free manner. Therefore, compared with \cite{huang2017arbitrary}, it does not have the limitation in generalisation capabilities. But the algorithm of Li \etal is still not effective at producing sharp details and fine strokes. The stylisation results will be shown in Section~\ref{sect:evaluation}. Also, it lacks a consideration in preserving depth information and variations in brush strokes.

%%%%%%%%%%%%%%%%%%%%%%%%%%%%
%%%%%%%%%%%%%%%%%%%%%%%%%%%%
%%%%%%%%%%%%%%%%%%%%%%%%%%%%

%\textbf{mention the style scability when introducing each algorithm}

%exploring \cite{ghiasi2017exploring}

%%%%%
\begin{figure*}
  \centering
  \subfigure[Content]{
    %\label{fig:subfig:a} %% label for first subfigure
    \includegraphics[width=0.23\textwidth]{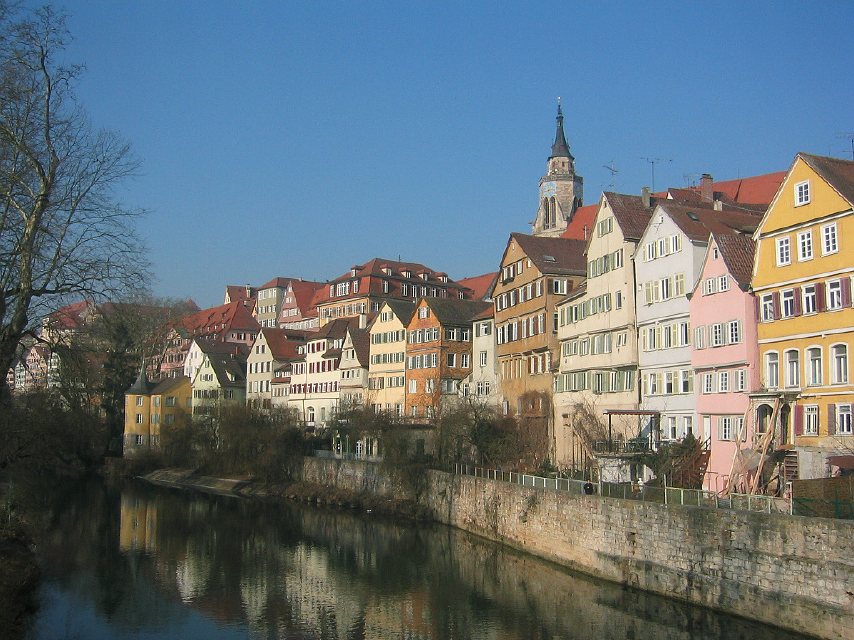}}
  %\hspace{0.05in}
  \subfigure[Style]{
    %\label{fig:subfig:a} %% label for first subfigure
    \includegraphics[width=0.23\textwidth]{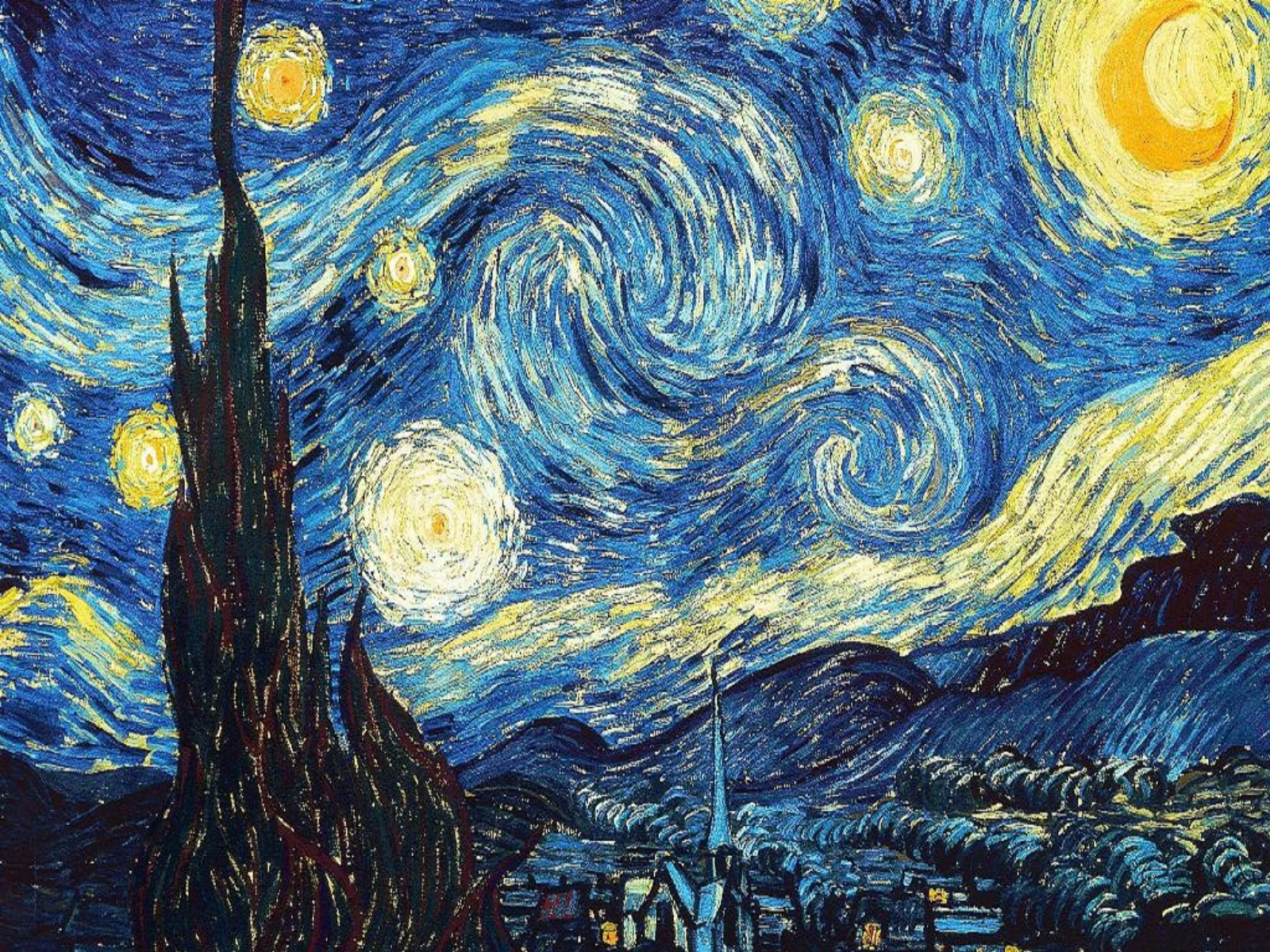}}
  %\hspace{0.05in}
  \subfigure[Small Stroke Size]{
    %\label{fig:subfig:b} %% label for second subfigure
    \includegraphics[width=0.23\textwidth]{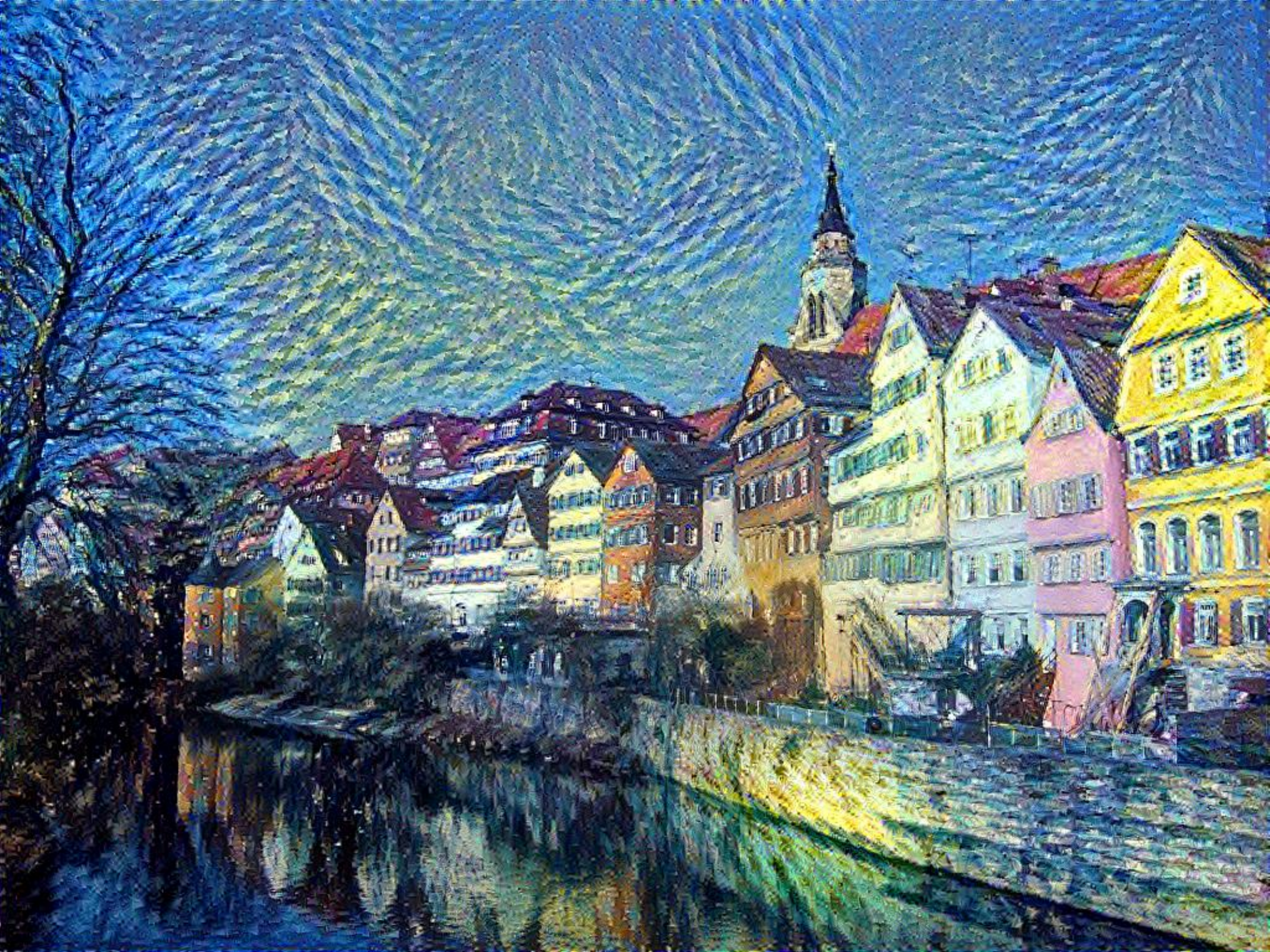}}
    %\hspace{0.05in}
     \subfigure[Large Stroke Size]{
      \includegraphics[width=0.23\textwidth]{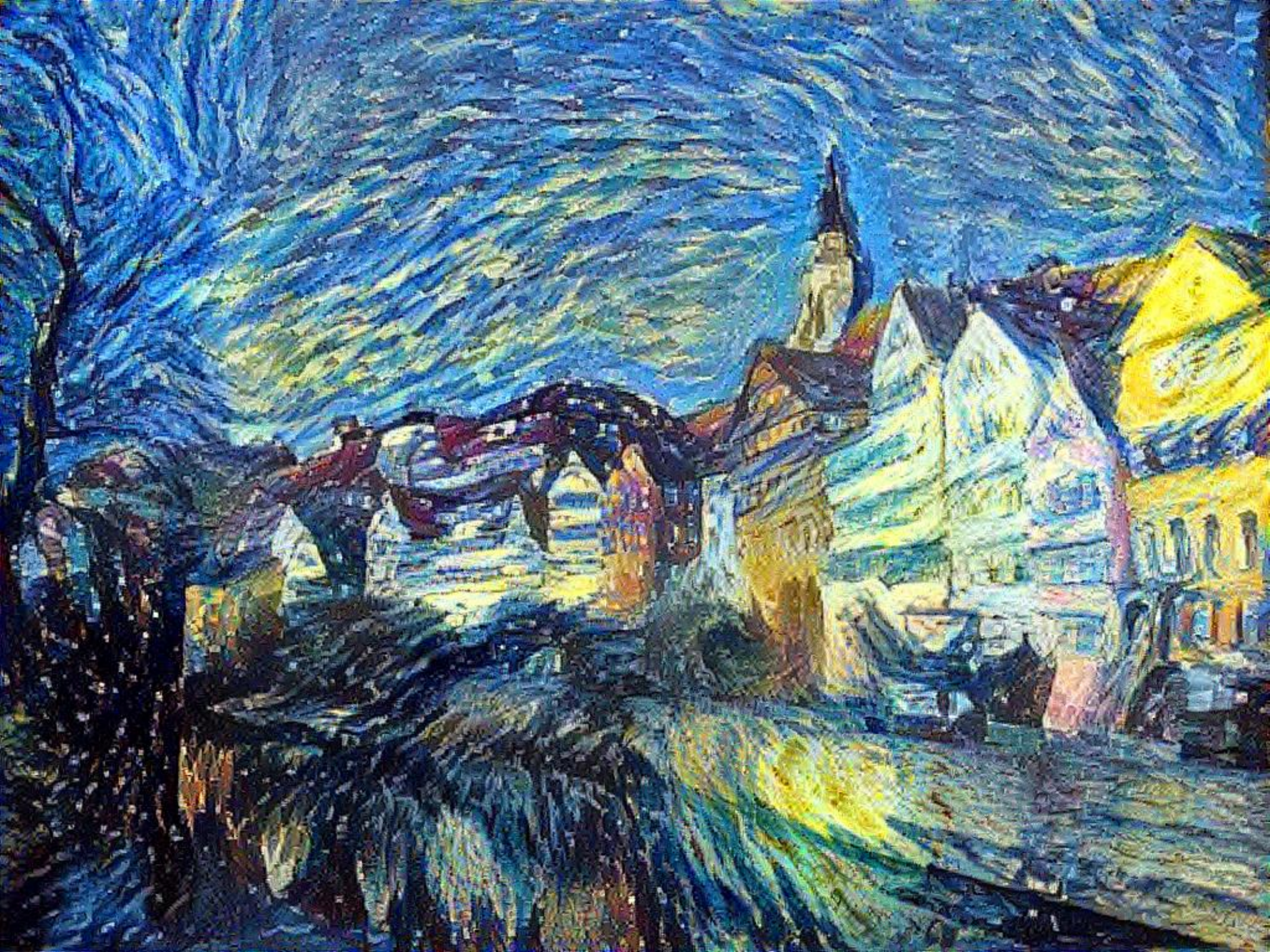}}
  \caption{Control the brush stroke size in NST. (c) is the output with smaller brush size and (d) with larger brush size. The style image is ``The Starry Night'' by Vincent van Gogh.}
  \label{fig:examplebrushSize} %% label for entire figure
\end{figure*}
%%%%%

\section{Improvements and Extensions}
\label{sect:improvementextensions}

Since the emergence of NST algorithms, there are also some researches devoted to improving current NST algorithms by controlling perceptual factors (\eg, stroke size control, spatial style control, and colour control) (Figure~\ref{fig:taxonomy}, green boxes). Also, all of aforementioned NST methods are designed for general still images. They may not be appropriate for specialised types of images and videos (\eg, doodles, head portraits, and video frames). Thus, a variety of follow-up studies (Figure~\ref{fig:taxonomy}, pink boxes) aim to extend general NST algorithms to these particular types of images and even extend them beyond artistic image style (\eg, audio style).

\textbf{Controlling Perceptual Factors in Neural Style Transfer.} Gatys \etal themselves \cite{gatys2016controlling} propose several slight modifications to improve their previous algorithm \cite{gatys2016image}. They demonstrate a spatial style control strategy to control the style in each region of the content image. Their idea is to define guidance channels for the feature activations for both content and style image. The guidance channel has values in $[0,1]$ specifying which style should be transferred to which content region, \ie, the content regions where the content guidance channel is $1$ should be rendered with the style where the style guidance channel is equal to $1$.
While for the colour control, the original NST algorithm produces stylised images with the colour distribution of the style image. However, sometimes people prefer a colour-preserving style transfer, \ie, preserving the colour of the content image during style transfer. The corresponding solution is to first transform the style image's colours to match the content image's colours before style transfer, or alternatively perform style transfer only in the luminance channel.

%where the desired region (getting the style) is assigned $1$ and otherwise, $0$.

For stroke size control, the problem is much more complex. We show sample results of stroke size control in Figure~\ref{fig:examplebrushSize}. The discussions of stroke size control strategy need to be split into several cases \cite{jing2018stroke}:

\emph{1) IOB-NST with non-high-resolution images:} Since current style statistics (\eg, Gram-based and BN-based statistics) are scale-sensitive \cite{jing2018stroke}, to achieve different stroke sizes, the solution is simply resizing a given style image to different scales.

\emph{2) MOB-NST with non-high-resolution images:} One possible solution is to resize the input image to different scales before the forward pass, which inevitably hurts stylisation quality. Another possible solution is to train multiple models with different scales of a style image, which is space and time consuming. Also, the possible solution fails to preserve \emph{stroke consistency} among results with different stroke sizes, \ie, the results vary in stroke orientations, stroke configurations, \etc. However, users generally desire to only change the stroke size but not others. To address this problem, Jing \etal \cite{jing2018stroke} propose a stroke controllable PSPM algorithm. The core component of their algorithm is a \emph{StrokePyramid} module, which learns different stroke sizes with adaptive receptive fields. Without trading off quality and speed, their algorithm is the first to exploit one single model to achieve flexible continuous stroke size control while preserving \emph{stroke consistency}, and further achieve spatial stroke size control to produce new artistic effects. Although one can also use ASPM algorithm to control stroke size, ASPM trades off quality and speed. As a result, ASPM is not effective at producing fine strokes and details compared with \cite{jing2018stroke}.

\emph{3) IOB-NST with high-resolution images:} For high-resolution images (\eg, $3000 \times 3000$ pixels in \cite{gatys2016controlling}), a large stroke size cannot be achieved by simply resizing style image to a large scale. Since only the region in the content image with a receptive field size of VGG can be affected by a neuron in the loss network, there is almost no visual difference between a large and larger brush strokes in a small image region with receptive field size. Gatys \etal \cite{gatys2016controlling} tackle this problem by proposing a coarse-to-fine IOB-NST procedure with several steps of downsampling, stylising, upsampling and final stylising.
%
%For the scale control, they separately study the scale control for style mixing and for high resolution (the outputs are with small brush size for high-resolution content image given that the output has the same resolution). The scale control for high resolution is essentially a strategy to control the brush size of the stylised image through a coarse-to-fine procedure with down-sampling and up-sampling. Figure~\ref{fig:examplebrushSize} shows the results of our reimplementation of the scale control strategy in controlling the brush size of the outputs. All these strategies make the process of style transfer more controllable and some of these modifications can be generally applied equally to the Generative Neural Methods.

\emph{4) MOB-NST with high-resolution images:} Similar to 3), stroke size in stylised result does not vary with style image scale for high-resolution images. The solution is also similar to Gatys \etal's algorithm in \cite{gatys2016controlling}, which is a coarse-to-fine stylisation procedure \cite{wang2016multimodal}. The idea is to exploit a multimodel, which comprises multiple subnetworks. Each subnetwork receives the upsampled stylised result of the previous subnetwork as the input, and stylises it again with finer strokes.

Another limitation of current NST algorithms is that they do not consider the depth information contained in the image. To address this limitation, the depth preserving NST algorithm \cite{liu2017depthaware} is proposed. Their approach is to add a depth loss function based on \cite{Johnson2016perceptual} to measure the depth difference between the content image and the stylised image. The image depth is acquired by applying a single-image depth estimation algorithm (\eg, Chen et al.'s work in \cite{chen2016single}).

\textbf{Semantic Style Transfer.} Given a pair of style and content images which are similar in content, the goal of semantic style transfer is to build a semantic correspondence between the style and content, which maps each style region to a corresponding semantically similar content region. Then the style in each style region is transferred to the semantically similar content region.

\emph{1) Image-Optimisation-Based Semantic Style Transfer.} Since the patch matching scheme naturally meets the requirements of the region-based correspondence, Champandard \cite{champandard2016semantic} proposes to build a semantic style transfer algorithm based on the aforementioned patch-based algorithm \cite{li2016combining} (Section~\ref{sect:MRF}). Although the result produced by the algorithm of Li and Wand \cite{li2016combining} is close to the target of semantic style transfer, \cite{li2016combining} does not incorporate an accurate segmentation mask, which sometimes leads to a wrong semantic match. Therefore, Champandard augments an additional semantic channel upon \cite{li2016combining}, which is a downsampled semantic segmentation map. The segmentation map can be either manually annotated or from a semantic segmentation algorithm \cite{ye2018finer,zhang2018context}. Despite the effectiveness of \cite{champandard2016semantic}, MRF-based design is not the only choice. Instead of combining MRF prior, Chen and Hsu \cite{chen2016towards} provide an alternative way for semantic style transfer, which is to exploit masking out process to constrain the spatial correspondence and also a higher order style feature statistic to further improve the result. More recently, Mechrez \etal \cite{mechrez2018contextual} propose an alternative contextual loss to realise semantic style transfer in a segmentation-free manner.

%Following the work of Li and Wand, Champandard \cite{champandard2016semantic} introduces the semantic map into the algorithm. By annotating the input image with a semantic map either manually authored or from the pixel labeling algorithms, the algorithm proposed by Champandard offers the user more control over the stylised result and thus improves the quality of stylisation.

\emph{2) Model-Optimisation-Based Semantic Style Transfer.} As before, the efficiency issue is always a big issue. Both \cite{champandard2016semantic} and \cite{chen2016towards} are based on IOB-NST algorithms and therefore leave much room for improvement. Lu \etal \cite{lu2017decoder} speed up the process by optimising the objective function in feature space, instead of in pixel space. More specifically, they propose to do feature reconstruction, instead of image reconstruction as previous algorithms do. This optimisation strategy reduces the computation burden, since the loss does not need to propagate through a deep network. The resulting reconstructed feature is decoded into the final result with a trained decoder. Since the speed of \cite{lu2017decoder} does not reach real-time, there is still big room for further research.

%semantic style transfer \cite{lu2017decoder}
%Although aforementioned algorithm achieves remarkable results, they do not consider the semantic content of the image, \ie, the transfer process is not content-aware.

\textbf{Instance Style Transfer.} Instance style transfer is built on instance segmentation and aims to stylise only a single user-specified object within an image. The challenge mainly lies in the transition between a stylised object and non-stylised background. Castillo \etal \cite{castillo2017zorn} tackle this problem by adding an extra MRF-based loss to smooth and anti-alias boundary pixels.% the transitions between stylised object and background.

%\textbf{Neural Style Transfer for Single User-specified Object.} A Targeted Style Transfer algorithm which is the process to stylize only a single user-specified object within an image is proposed by Castillo \etal in \cite{castillo2017zorn}. The idea is to segment the target object from the stylised image using a semantic segmentation algorithm and then merge the extracted stylised object with the non-stylised background.

\textbf{Doodle Style Transfer.} An interesting extension can be found in \cite{champandard2016semantic}, which is to exploit NST to transform rough sketches into fine artworks. The method is simply discarding content loss term and using doodles as segmentation map to do semantic style transfer.

%using high-level annotations of the input image.

%which has been introduced in Section~\ref{sect:MRF}. Other than aforementioned contribution which is introducing the semantic map into the Neural Style Transfer algorithm, \cite{champandard2016semantic} can also be used to transform a rough sketch into fine artwork using high-level annotations of the input image.

\textbf{Stereoscopic Style Transfer.} Driven by the demand of AR/VR, Chen \etal \cite{chen2018stereoscopic} propose a stereoscopic NST algorithm for stereoscopic images. They propose a disparity loss to penalise the bidirectional disparity. Their algorithm is shown to produce more consistent strokes for different views.

\textbf{Portrait Style Transfer.} Current style transfer algorithms are usually not optimised for head portraits. As they do not impose spatial constraints, directly applying these existing algorithms to head portraits will deform facial structures, which is unacceptable for the human visual system. Selim \etal \cite{selim2016painting} address this problem and extend \cite{gatys2016image} to head portrait painting transfer. They propose to use the notion of gain maps to constrain spatial configurations, which can preserve the facial structures while transferring the texture of the style image.

\textbf{Video Style Transfer.} NST algorithms for video sequences are substantially proposed shortly after Gatys et al.'s first NST algorithm for still images \cite{gatys2016image}. Different from still image style transfer, the design of video style transfer algorithm needs to consider the smooth transition between adjacent video frames. Like before, we divide related algorithms into Image-Optimisation-Based and Model-Optimisation-Based Video Style Transfer.

%introduces a temporal loss function
%use a temporal constraint to preserve smooth transition between individual frames, \ie, penalizing deviation along the point trajectories. Their algorithm is shown to be able to eliminate most temporal artifacts and produce smooth stylised videos.

\emph{1) Image-Optimisation-Based Online Video Style Transfer.} The first video style transfer algorithm is proposed by Ruder \etal \cite{ruder2016artistic,Ruder2018}. They introduce a temporal consistency loss based on optical flow to penalise the deviations along point trajectories. The optical flow is calculated by using novel optical flow estimation algorithms \cite{weinzaepfel2013deepflow,revaud2015epicflow}. As a result, their algorithm eliminates temporal artefacts and produces smooth stylised videos. However, they build their algorithm upon \cite{gatys2016image} and need several minutes to process a single frame.

\emph{2) Model-Optimisation-Based Offline Video Style Transfer.} Several follow-up studies are devoted to stylising a given video in real-time. Huang \etal \cite{huang2017real} propose to augment Ruder et al.'s temporal consistency loss \cite{ruder2016artistic} upon current PSPM algorithm. Given two consecutive frames, the temporal consistency loss is directly computed using two corresponding outputs of style transfer network to encourage pixel-wise consistency, and a corresponding two-frame synergic training strategy is introduced for the computation of temporal consistency loss. Another concurrent work which shares a similar idea with \cite{huang2017real} but with an additional exploration of style instability problem can be found in \cite{gupta2017characterizing}. Different from \cite{huang2017real,gupta2017characterizing}, Chen \etal \cite{chen2017coherent} propose a flow subnetwork to produce feature flow and incorporate optical flow information in feature space. Their algorithm is built on a pre-trained style transfer network (an encoder-decoder pair) and wraps feature activations from the pre-trained stylisation encoder using the obtained feature flow.

%video sequences is given in \cite{ruder2016artistic} by Ruder et al., which is referred to as Neural Video Style Transfer in our paper. Given a style image, the algorithm of Ruder \etal introduces a temporal loss function to transfer its artistic style to the entire video. The key idea behind their algorithm is to use a temporal constraint to preserve smooth transition between individual frames, \ie, penalizing deviation along the point trajectories. Their algorithm is shown to be able to eliminate most temporal artifacts and produce smooth stylised videos. Another concurrent work is the algorithm proposed by Anderson \etal \cite{anderson2016deepmovie} which is able to render a movie by exploiting optical flow to initialize the style transfer Optimisation.

\textbf{Character Style Transfer.} Given a style image containing multiple characters, the goal of \emph{Character Style Transfer} is to apply the idea of NST to generate new fonts and text effects. In \cite{atarsaikhan2017neural}, Atarsaikhan \etal directly apply the algorithm in \cite{gatys2016image} to font style transfer and achieve visually plausible results. While Yang \etal \cite{yang2017awesome} propose to first characterise style elements and exploit extracted characteristics to guide the generation of text effects. A more recent work \cite{azadi2018multi} designs a conditional GAN model for glyph shape prediction, and also an ornamentation network for colour and texture prediction. By training these two networks jointly, font style transfer can be realised in an end-to-end manner.

%In \cite{azadi2018multi}, Azadi \etal design a conditional GAN model to generate ornamented glyphs
%
%font style transfer and text effect transfer \cite{azadi2018multi}
%
%font style transfer \cite{atarsaikhan2017neural} \cite{azadi2018multi}

\textbf{Photorealistic Style Transfer.} Photorealistic style transfer (also known as colour style transfer) aims to transfer the style of colour distributions. The general idea is to build upon current semantic style transfer but to eliminate distortions and preserve the original structure of the content image.

\emph{1) Image-Optimisation-Based Photorealistic Style Transfer.} The earliest photorealistic style transfer approach is proposed by Luan \etal \cite{luan2017deep}. They propose a two-stage optimisation procedure, which is to initialise the optimisation by stylising a given photo with non-photorealistic style transfer algorithm \cite{champandard2016semantic} and then penalise image distortions by adding a photorealism regularization. But since Luan et al.'s algorithm is built on the \emph{Image-Optimisation-Based Semantic Style Transfer} method \cite{champandard2016semantic}, their algorithm is computationally expensive. Similar to \cite{luan2017deep}, another algorithm proposed by Mechrez \etal \cite{mechrez2017photorealistic} also adopts a two-stage optimisation procedure. They propose to refine the non-photorealistic stylised result by matching the gradients in the output image to those in the content photo. Compared to \cite{luan2017deep}, the algorithm of Mechrez \etal achieves a faster photorealistic stylisation speed.

% They propose to add a photorealism regularization upon \cite{champandard2016semantic} to penalise image distortions.
% refines stylized results by matching the gradients in the output photo to those in the content photo in the output image to those

\emph{2) Model-Optimisation-Based Photorealistic Style Transfer.} Li \etal \cite{li2018closed} address the efficiency issue of \cite{luan2017deep} by handling this problem with two steps, the stylisation step and smoothing step. The stylisation step is to apply the NST algorithm in \cite{li2017universal} but replace upsampling layers with unpooling layers to produce the stylised result with fewer distortions. Then the smoothing step further eliminates structural artefacts. These two aforementioned algorithms \cite{luan2017deep,li2018closed} are mainly designed for natural images. Another work in \cite{zhang2017style} proposes to exploit GAN to transfer the colour from human-designed anime images to sketches. Their algorithm demonstrates a promising application of Photorealistic Style Transfer, which is the automatic image colourisation.

%for photo, for non-natural images such as anime style transfer for animes \cite{zhang2017style} photo style transfer \cite{luan2017deep} \cite{li2018closed}

\textbf{Attribute Style Transfer.} Image attributes are generally referred to image colours, textures, \etc. Previously, image attribute transfer is accomplished through image analogy \cite{hertzmann2001image} in a supervised manner (Section~\ref{sect:preneural}). Derived from the idea of patch-based NST \cite{li2016combining}, Liao \etal \cite{liao2017visual} propose a deep image analogy to study image analogy in the domain of CNN features. Their algorithm is based on a patch matching technique and realises a weakly supervised image analogy, \ie, their algorithm only needs a single pair of source and target images instead of a large training set.

\textbf{Fashion Style Transfer.} Fashion style transfer receives fashion style image as the target and generates clothing images with desired fashion styles. The challenge of Fashion Style Transfer lies in the preservation of similar design with the basic input clothing while blending desired style patterns. This idea is first proposed by Jiang and Fu \cite{jiang2017fashion}. They tackle this problem by proposing a pair of fashion style generator and discriminator.

\textbf{Audio Style Transfer.} In addition to transferring image styles, \cite{verma2018neural,mital2017time} extend the domain of image style to audio style, and synthesise new sounds by transferring the desired style from a target audio. The study of audio style transfer also follows the route of image style transfer, \ie, \emph{Audio-Optimisation-Based Online Audio Style Transfer} and then \emph{Model-Optimisation-Based Offline Audio Style Transfer}. Inspired by image-based IOB-NST, Verma and Smith \cite{verma2018neural} propose a \emph{Audio-Optimisation-Based Online Audio Style Transfer} algorithm based on online audio optimisation. They start from a noise signal and optimise it iteratively using backpropagation. \cite{mital2017time} improves the efficiency by transferring an audio in a feed-forward manner and can produce the result in real-time.

\newcommand\m{0.09}

\begin{figure}[!tbp]
\setlength\tabcolsep{1.5 pt}
{\renewcommand{\arraystretch}{0.9}
\begin{tabular}{c c c c c }
\centering

 \includegraphics[width=\m\textwidth]{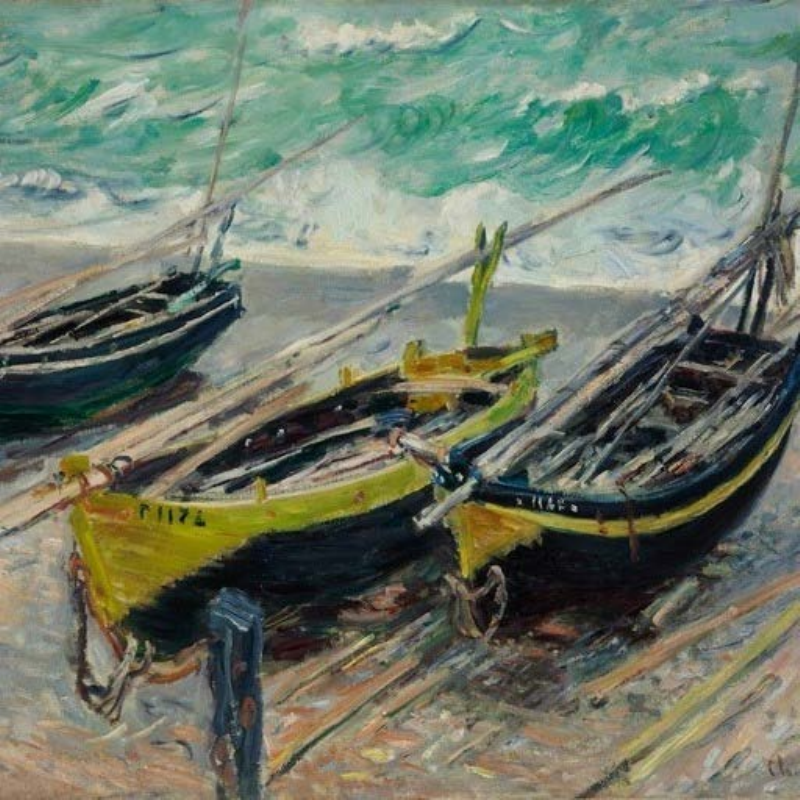} &\includegraphics[width=\m\textwidth]{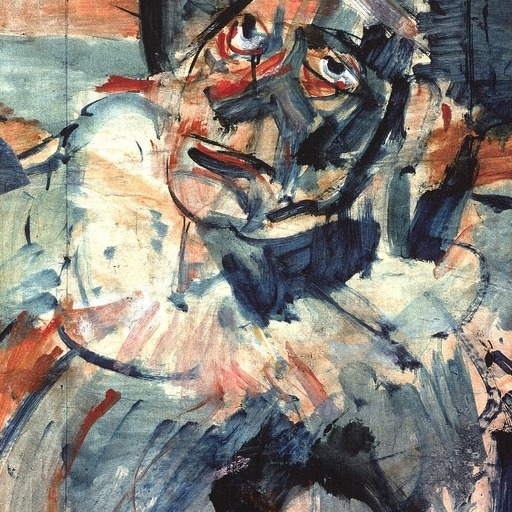}  & \includegraphics[width=\m\textwidth]{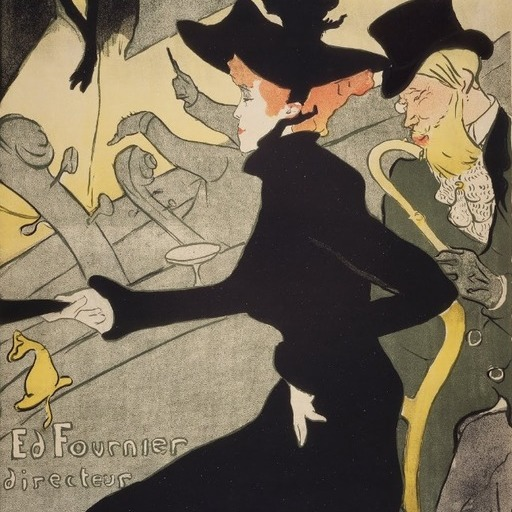}&\includegraphics[width=\m\textwidth]{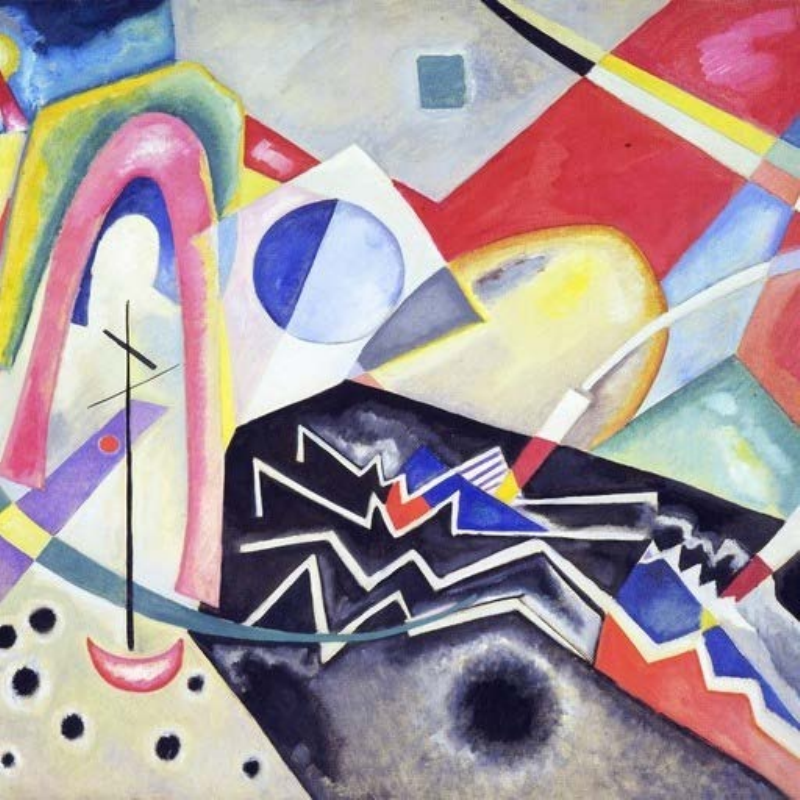} & \includegraphics[width=\m\textwidth]{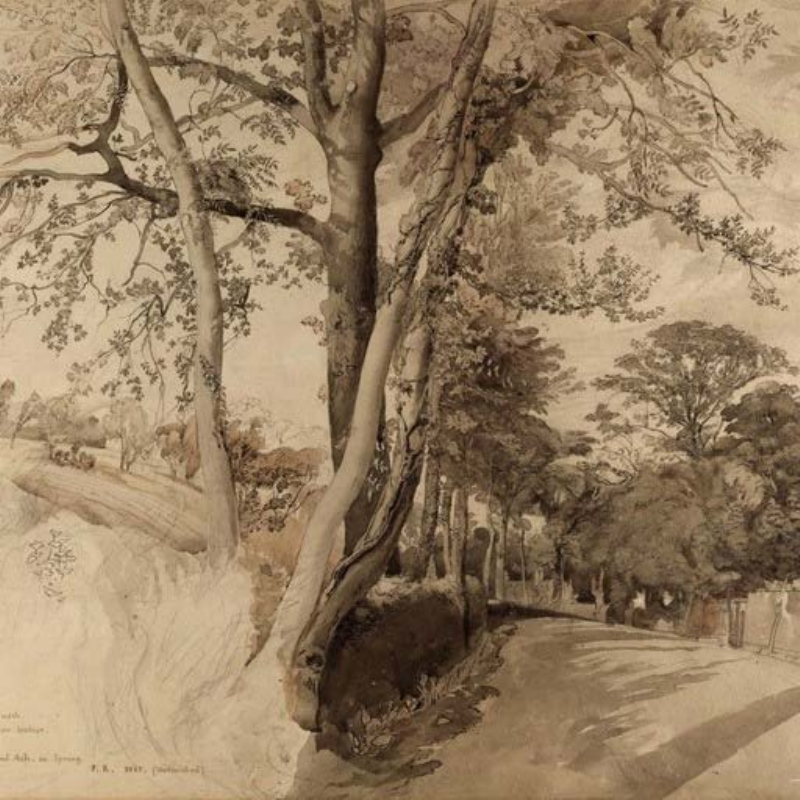} \\
 (1) & (2)&(3)&(4)&(5)\\
 \includegraphics[width=\m\textwidth]{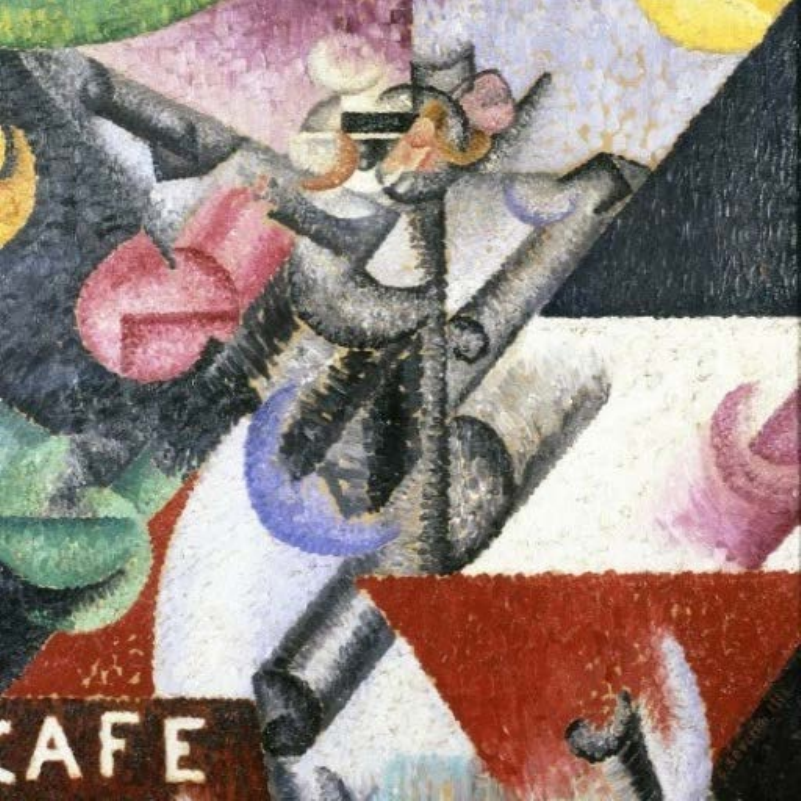} & \includegraphics[width=\m\textwidth]{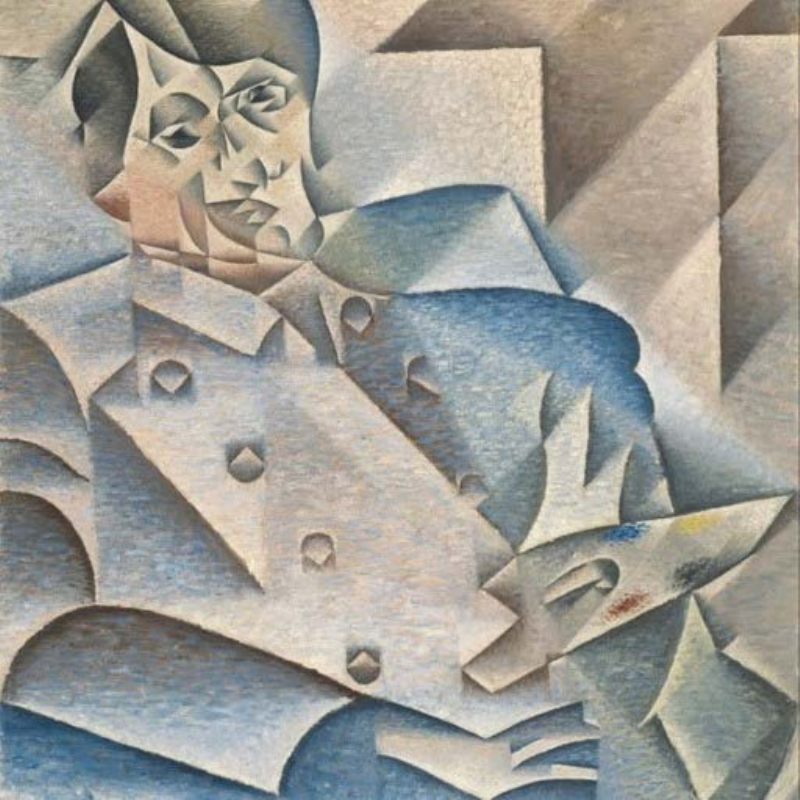} &\includegraphics[width=\m\textwidth]{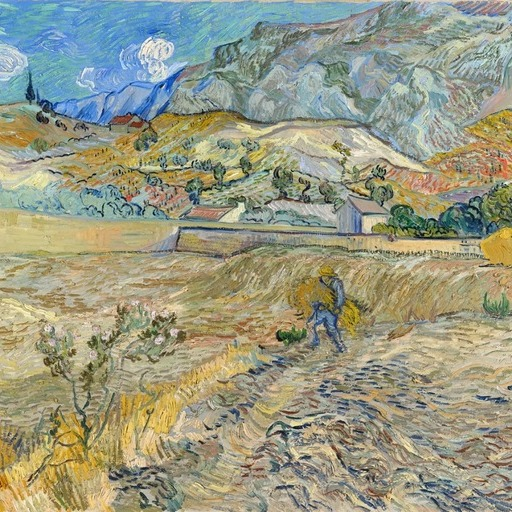}& \includegraphics[width=\m\textwidth]{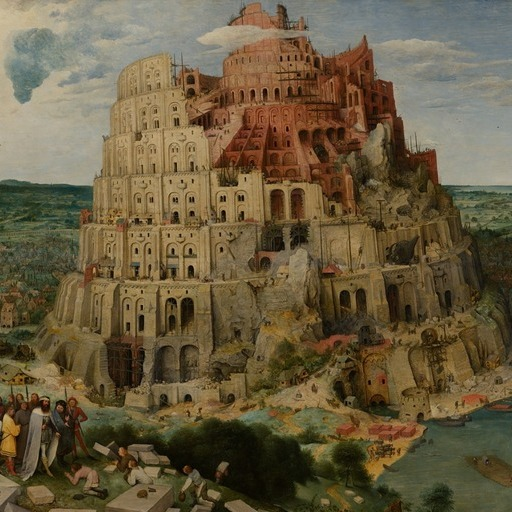}& \includegraphics[width=\m\textwidth]{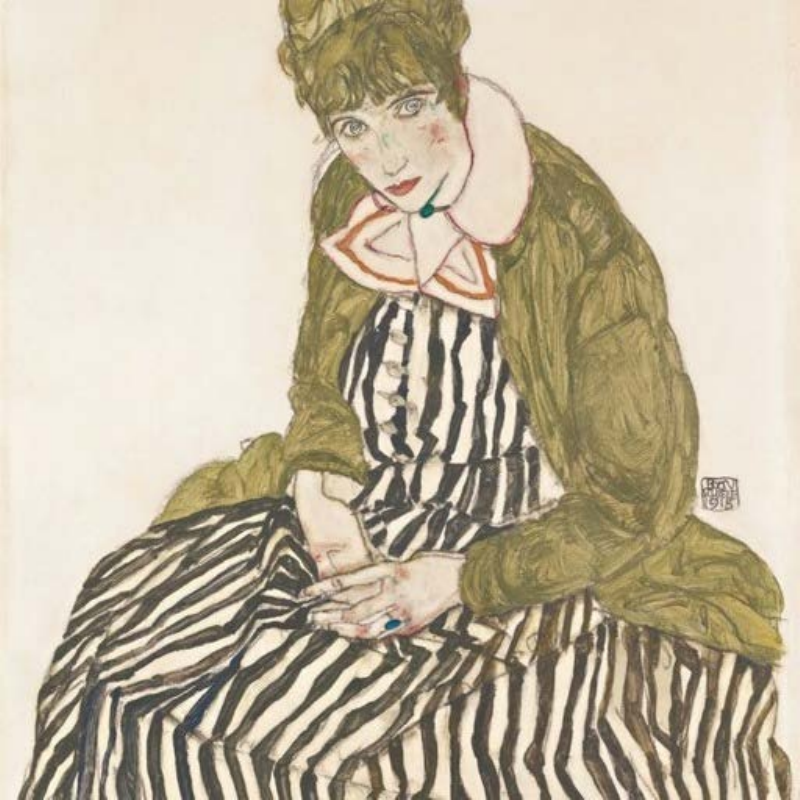}\\
 (6)&(7)&(8)&(9)&(10)
\end{tabular}
}
\caption{Diversified style images used in our experiment.}

\label{fig:styleimage} %% label for entire figure
\end{figure}

\begin{table}
\renewcommand\arraystretch{1.3}
\caption{Detailed information of our style images.}
%\begin{minipage}{\columnwidth}
\begin{center}
\begin{tabular}{  |c | p{3.33cm}| p{3.9cm}|  }
    %\toprule[1pt]
    %\hline
    \hline
    \textbf{No.} &  \textbf{Author}&  \textbf{Name \& Year}\\
    \hline
    1 & Claude Monet & \emph{Three Fishing Boats} (1886) \\
    2 & Georges Rouault& \emph{Head of a Clown}  (1907)\\
    3 & Henri de Toulouse-Lautrec& \emph{Divan Japonais} (1893)\\
    4 & Wassily Kandinsky & \emph{White Zig Zags} (1922)\\
    5 & John Ruskin & \emph{Trees in a Lane} (1847)\\
    6 & Severini Gino & \emph{Ritmo plastico del 14 luglio} (1913)\\
    7 & Juan Gris & \emph{Portrait of Pablo Picasso} (1912)\\
    8 & Vincent van Gogh & \emph{Landscape at Saint-R\'emy} (1889) \\
    9 & Pieter Bruegel the Elder& \emph{The Tower of Babel} (1563)\\
    10 & Egon Schiele & \emph{Edith with Striped Dress} (1915)\\
    \hline
    %\midrule[1pt]
     \end{tabular}
\end{center}
%\bigskip\centering
\footnotesize
%\begin{flushleft}
\smallskip

\qquad\qquad\emph{Note:} All our style images are in the public domain. \hfill
%\end{flushleft}
\label{table:styleimage}
%\end{minipage}
\end{table}
%%%%%%%%%%%%%%%%

%%%%%%%%%%%%%%%%%%%%%%%%%%%%
%%%%%%%%%%%%%%%%%%%%%%%%%%%%
%%%%%%%%%%%%%%%%%%%%%%%%%%%%

\section{Evaluation Methodology}
\label{sect:evaluation}

%\footnote{\url{https://github.com/anishathalye/neural-style}}
%\footnote{\url{https://github.com/jcjohnson/fast-neural-style}}
%\footnote{\url{https://github.com/DmitryUlyanov/texture_nets}}
%\footnote{\url{https://github.com/rtqichen/style-swap}}
%\footnote{\url{https://github.com/chuanli11/MGANs}}
%\footnote{\url{https://github.com/tensorflow/magenta/tree/master/magenta/models/image_stylisation}}
%
%Since there is no ground truth for NST, it is difficult to quantify ``aesthetic beauty'' of different results.

The evaluations of NST algorithms remain an open and important problem in this field. In general, there are two major types of evaluation methodologies that can be employed in the field of NST, \ie, qualitative evaluation and quantitative evaluation. Qualitative evaluation relies on the aesthetic judgements of observers. The evaluation results are related to lots of factors (\eg, age and occupation of participants). While quantitative evaluation focuses on the precise evaluation metrics, which include time complexity, loss variation, \etc. In this section, we experimentally compare different NST algorithms both qualitatively and quantitatively.

%We hope our study can build a \emph{standardised benchmark} for this area.
%Therefore, the evaluation of visual results produced by Neural Style Transfer algorithms remains an open and important problem.
%Currently, Fast Style Transfer has become a trending topic in that the speed issue is one of the major concerns for industrial applications. However, none of previous researches run all of these popular Fast Style Transfer algorithms under the same experimental settings and compare them both qualitatively and quantitatively.

%In this section, we experimentally compare current Fast Style Transfer algorithms, along with Gatys \etal's Slow Neural Style Transfer as the baseline

%\newpage
%\thispagestyle{empty}
\newcommand\x{0.14}
\newcommand\y{2.5cm}
\newcommand\z{2.5cm}
\begin{figure*}[!tbp]
\setlength\tabcolsep{1.5 pt}
{\renewcommand{\arraystretch}{0.9}
\begin{tabular}{m{\y} >{\centering}m{\z} >{\centering}m{\z} >{\centering}m{\z} >{\centering}m{\z} >{\centering}m{\z} >{\centering\arraybackslash}m{2.5cm}}
\centering

& \textbf{\footnotesize{Group \uppercase\expandafter{\romannumeral1}}} & \textbf{\footnotesize{Group \uppercase\expandafter{\romannumeral2}}} & \textbf{\footnotesize{Group \uppercase\expandafter{\romannumeral3}}} & \textbf{\footnotesize{Group \uppercase\expandafter{\romannumeral4}}} & \textbf{\footnotesize{Group \uppercase\expandafter{\romannumeral5}}} & \textbf{\footnotesize{Group \uppercase\expandafter{\romannumeral6}}}\\ %\vspace{1cm}

%\textbf{\small{Style:}}  & \includegraphics[width=\x\textwidth]{figs/style/1.pdf} & \includegraphics[width=\x\textwidth]{figs/style/2.pdf}& \includegraphics[width=\x\textwidth]{figs/style/4.pdf} & \includegraphics[width=\x\textwidth]{figs/style/5.pdf} & \includegraphics[width=\x\textwidth]{figs/style/6.pdf} & \includegraphics[width=\x\textwidth]{figs/style/7.pdf} & \includegraphics[width=\x\textwidth]{figs/style/9.pdf} & \includegraphics[width=\x\textwidth]{figs/style/10.pdf}\\

\textbf{\footnotesize{Content \& Style:}} & \includegraphics[width=\x\textwidth]{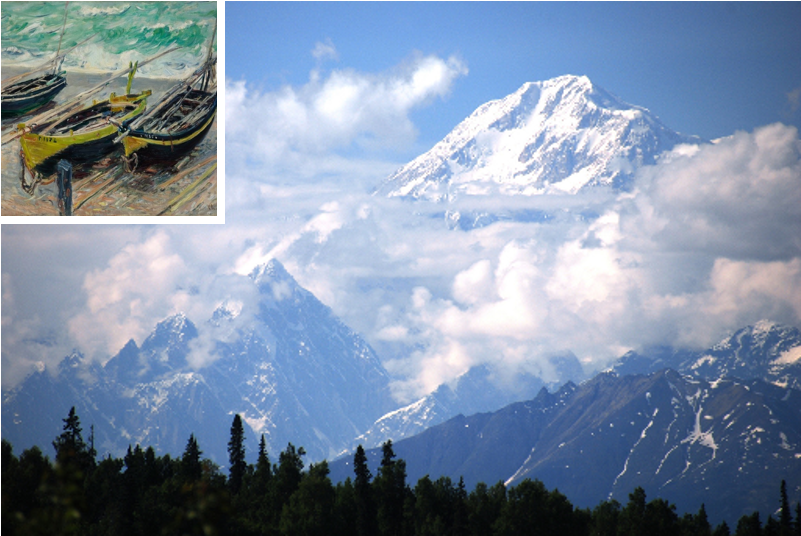} &\includegraphics[width=\x\textwidth]{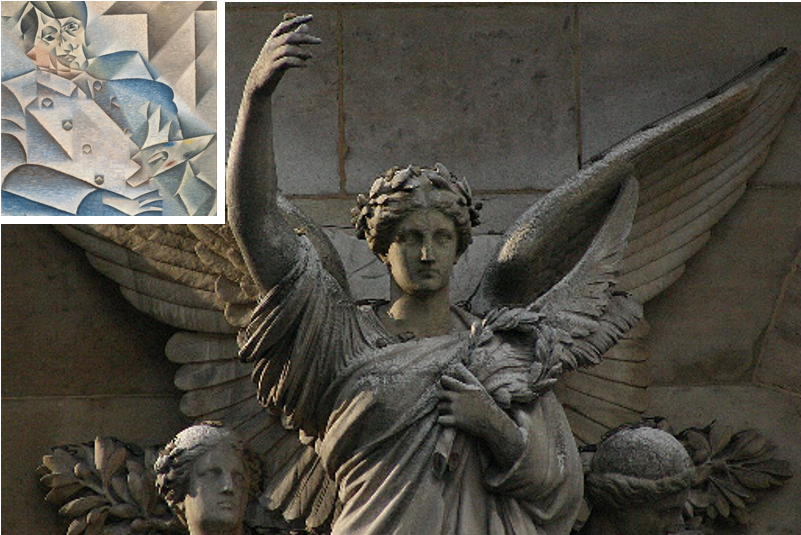}&\includegraphics[width=\x\textwidth]{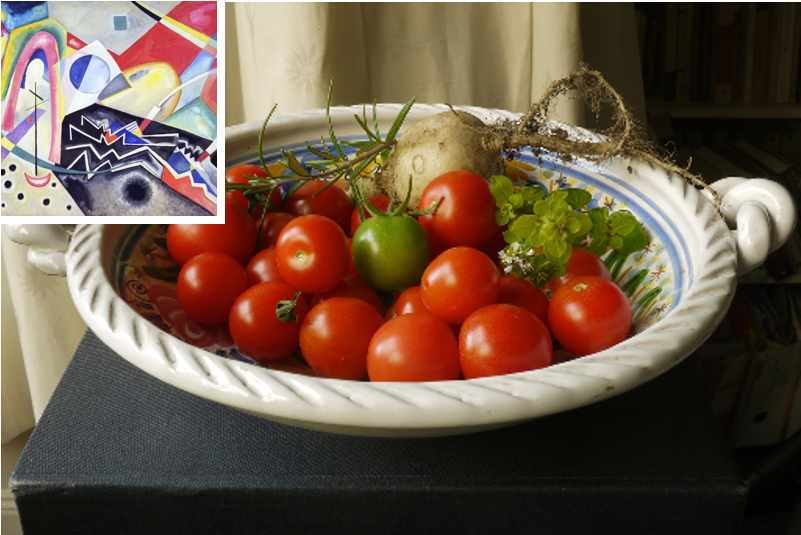} & \includegraphics[width=\x\textwidth]{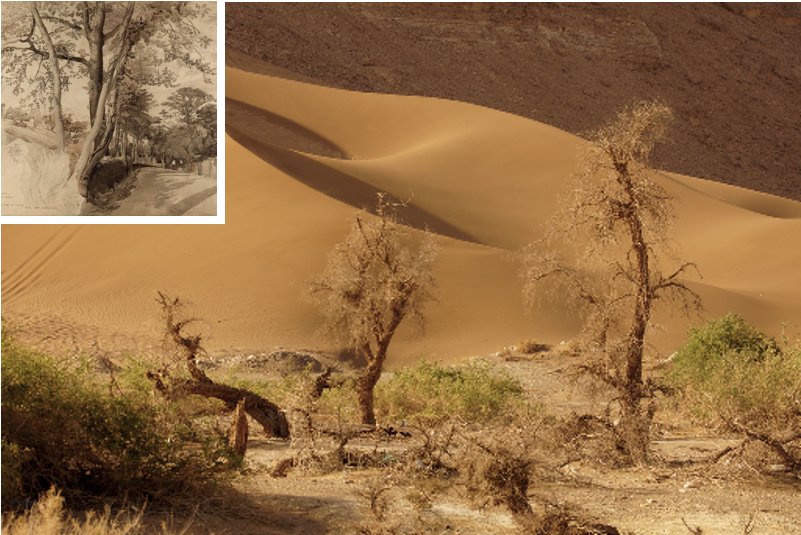} &\includegraphics[width=\x\textwidth]{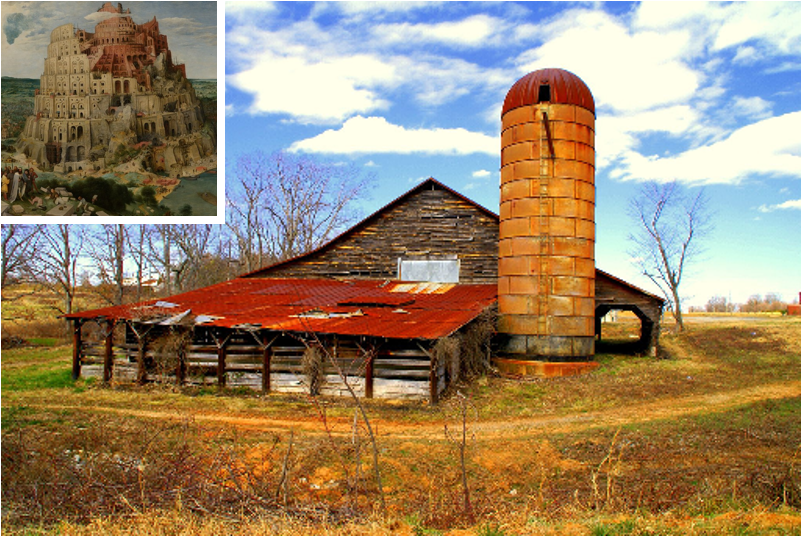} & \includegraphics[width=\x\textwidth]{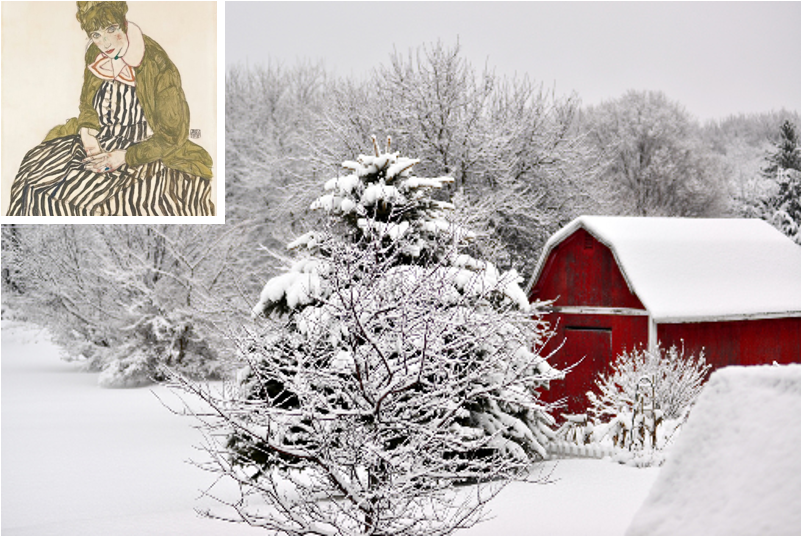} \\

%\toprule
\textbf{\footnotesize{Gatys \etal \cite{gatys2016image}:}} & \includegraphics[width=\x\textwidth]{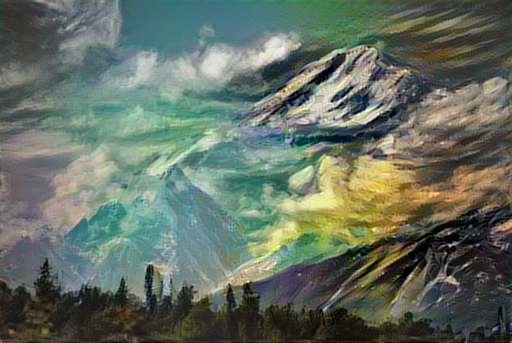} & \includegraphics[width=\x\textwidth]{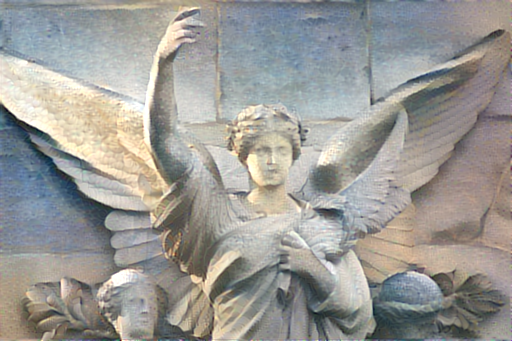}& \includegraphics[width=\x\textwidth]{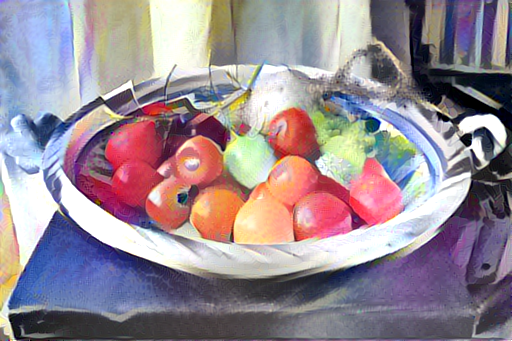} &\includegraphics[width=\x\textwidth]{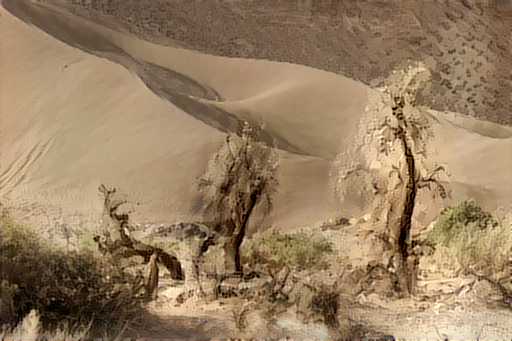} &\includegraphics[width=\x\textwidth]{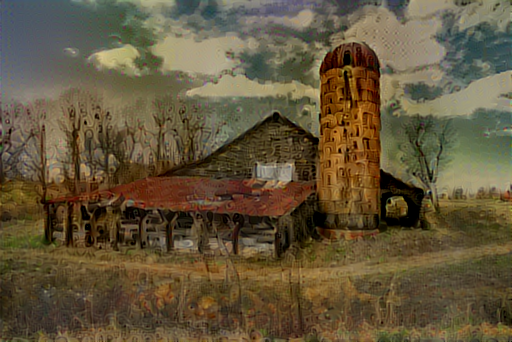} &\includegraphics[width=\x\textwidth]{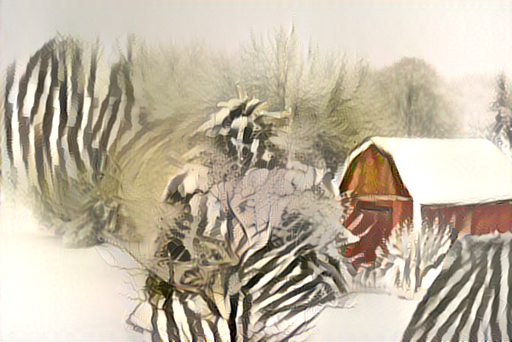} \\
%\toprule

\textbf{\footnotesize{Johnson \etal \cite{Johnson2016perceptual}:}} & \includegraphics[width=\x\textwidth]{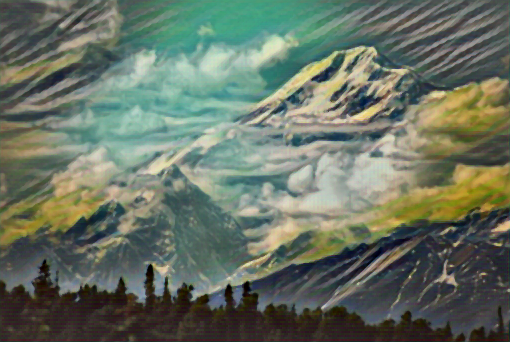} & \includegraphics[width=\x\textwidth]{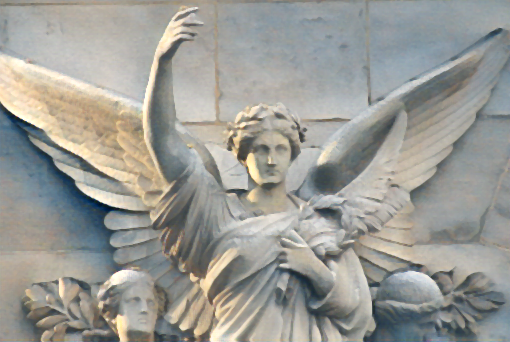} & \includegraphics[width=\x\textwidth]{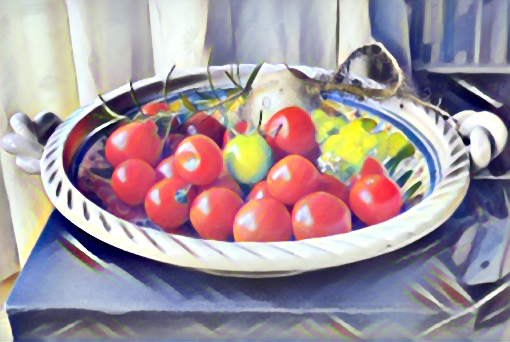}& \includegraphics[width=\x\textwidth]{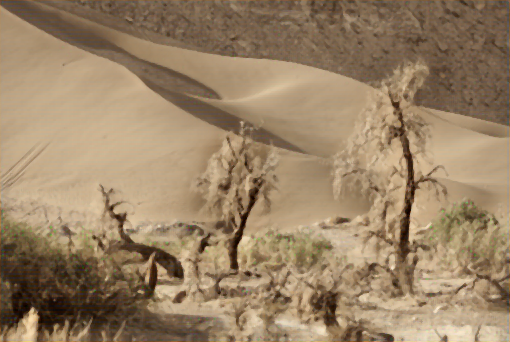}& \includegraphics[width=\x\textwidth]{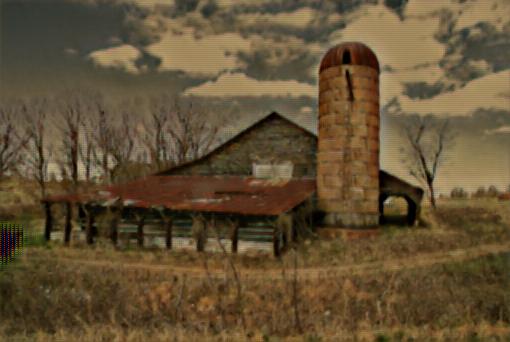}&  \includegraphics[width=\x\textwidth]{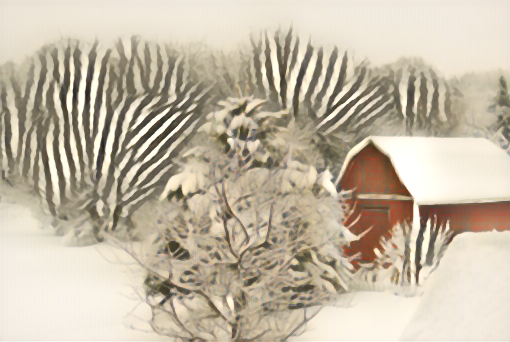}\\

\textbf{\footnotesize{Ulyanov \etal \cite{ulyanov2016texture}:}} & \includegraphics[width=\x\textwidth]{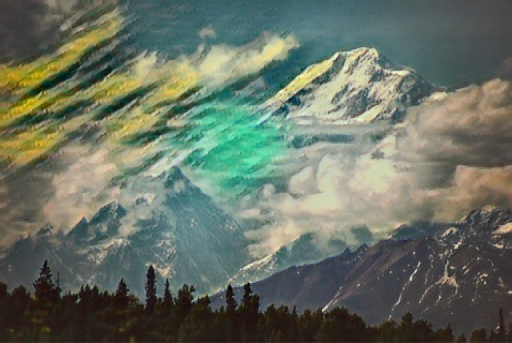} & \includegraphics[width=\x\textwidth]{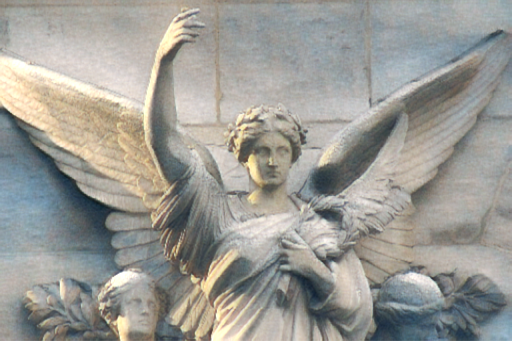} & \includegraphics[width=\x\textwidth]{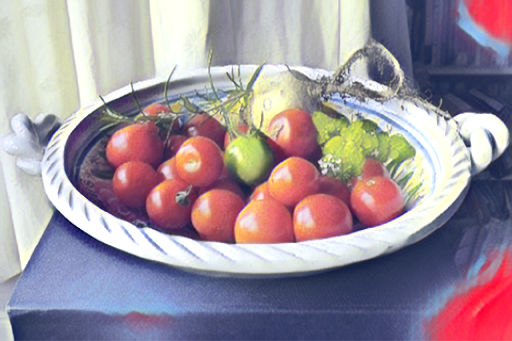}& \includegraphics[width=\x\textwidth]{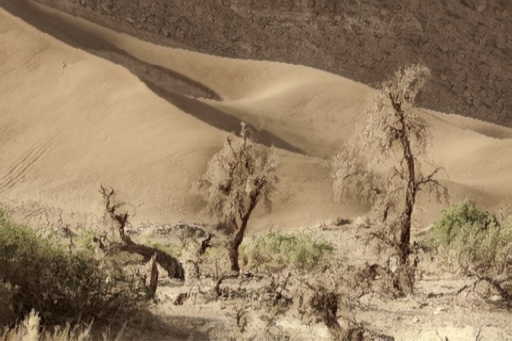}& \includegraphics[width=\x\textwidth]{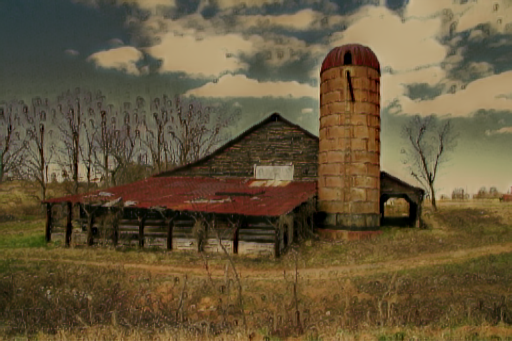}& \includegraphics[width=\x\textwidth]{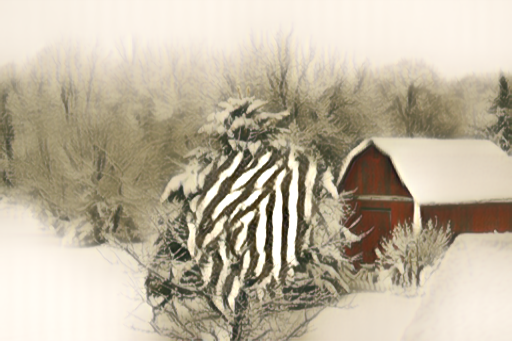}\\

\textbf{\footnotesize{Li and Wand \cite{li2016precomputed}:}} & \includegraphics[width=\x\textwidth]{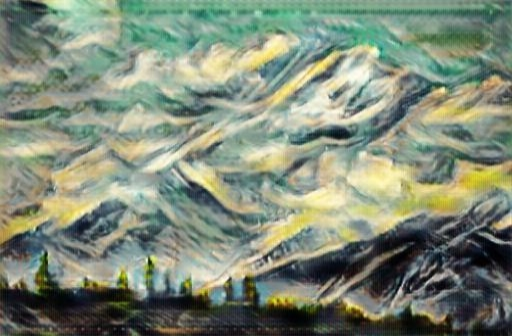} & \includegraphics[width=\x\textwidth]{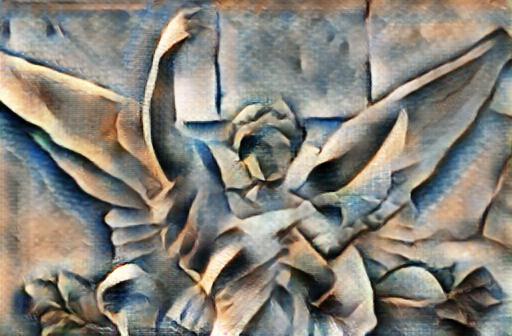} & \includegraphics[width=\x\textwidth]{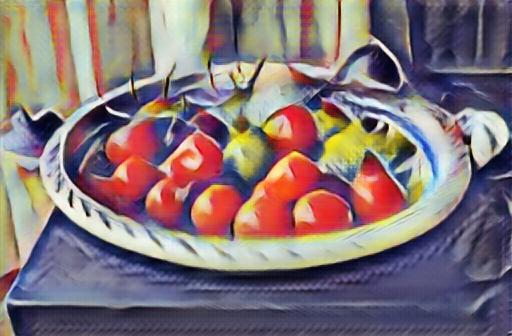}& \includegraphics[width=\x\textwidth]{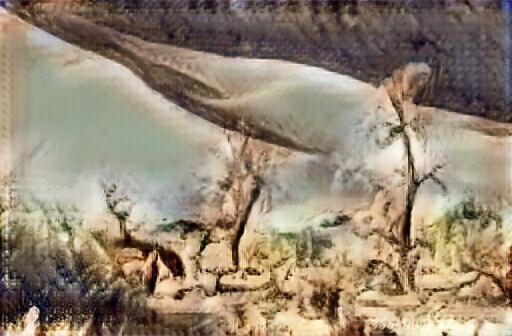}& \includegraphics[width=\x\textwidth]{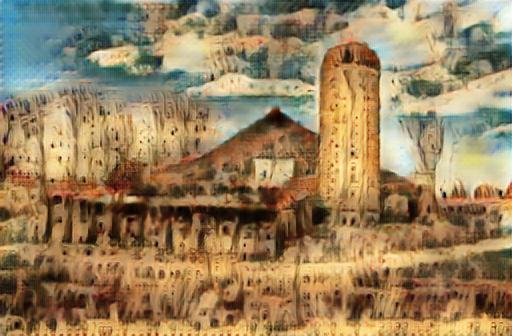}& \includegraphics[width=\x\textwidth]{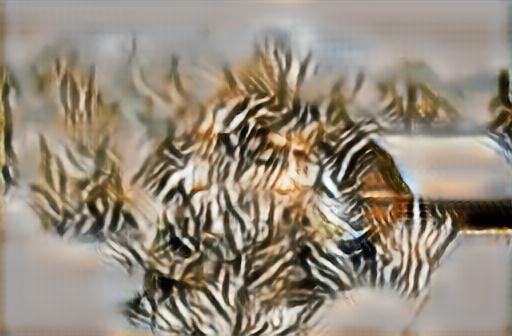}\\

\end{tabular}
}

\caption{Some example results of \textbf{IOB-NST} and \textbf{PSPM-MOB-NST} for qualitative evaluation. The content images are from the benchmark dataset proposed by Mould and Rosin \cite{mould2016benchmark,mould2017developing}. The style images are in the public domain. Detailed information of our style images can be found in Table~\ref{table:styleimage}.}
\label{fig:qualitativeresult1}
\end{figure*}

%\vspace{-10cm}

\begin{figure*}[!tbp]
\setlength\tabcolsep{1.5 pt}
{\renewcommand{\arraystretch}{0.9}
\begin{tabular}{m{\y} >{\centering}m{\z} >{\centering}m{\z} >{\centering}m{\z} >{\centering}m{\z} >{\centering}m{\z} >{\centering\arraybackslash}m{2.5cm}}
\centering

& \textbf{\footnotesize{Group \uppercase\expandafter{\romannumeral1}}} & \textbf{\footnotesize{Group \uppercase\expandafter{\romannumeral2}}} & \textbf{\footnotesize{Group \uppercase\expandafter{\romannumeral3}}} & \textbf{\footnotesize{Group \uppercase\expandafter{\romannumeral4}}} & \textbf{\footnotesize{Group \uppercase\expandafter{\romannumeral5}}} & \textbf{\footnotesize{Group \uppercase\expandafter{\romannumeral6}}}\\ %\vspace{1cm}

%\textbf{\small{Style:}}  & \includegraphics[width=\x\textwidth]{figs/style/1.pdf} & \includegraphics[width=\x\textwidth]{figs/style/2.pdf}& \includegraphics[width=\x\textwidth]{figs/style/4.pdf} & \includegraphics[width=\x\textwidth]{figs/style/5.pdf} & \includegraphics[width=\x\textwidth]{figs/style/6.pdf} & \includegraphics[width=\x\textwidth]{figs/style/7.pdf} & \includegraphics[width=\x\textwidth]{figs/style/9.pdf} & \includegraphics[width=\x\textwidth]{figs/style/10.pdf}\\

\textbf{\footnotesize{Content:}} & \includegraphics[width=\x\textwidth]{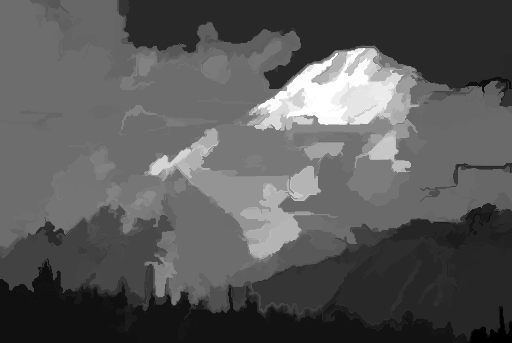} &\includegraphics[width=\x\textwidth]{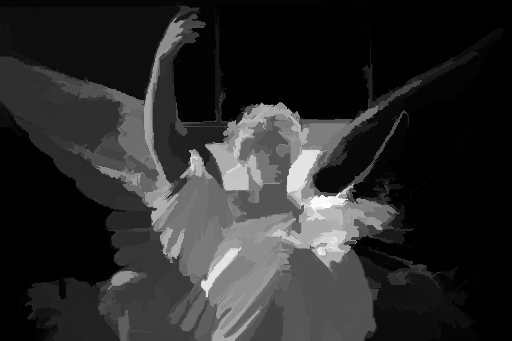}&\includegraphics[width=\x\textwidth]{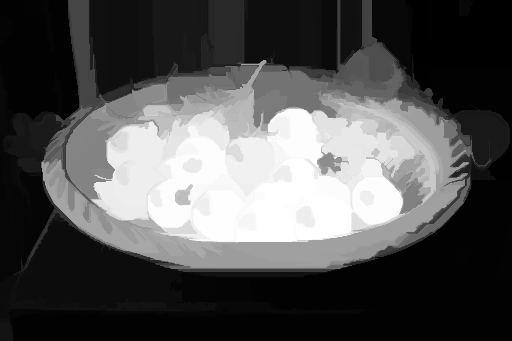} & \includegraphics[width=\x\textwidth]{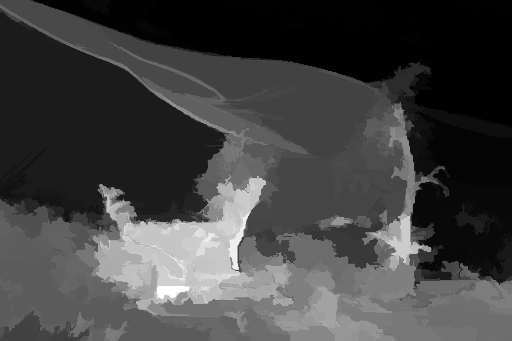} &\includegraphics[width=\x\textwidth]{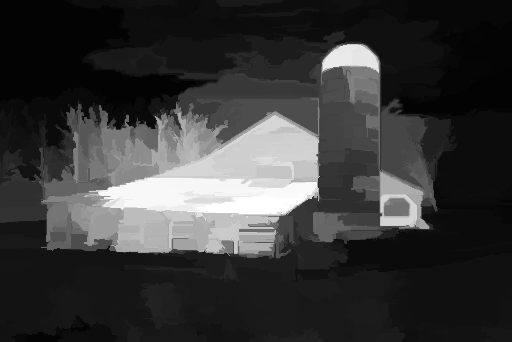} & \includegraphics[width=\x\textwidth]{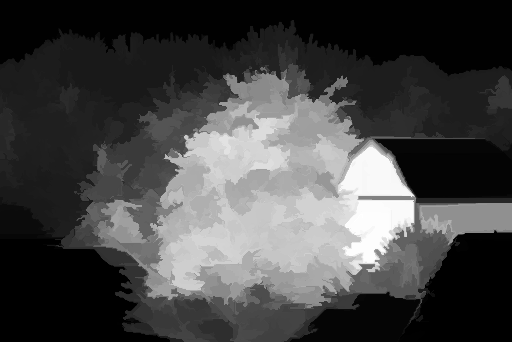} \\

%\toprule
\textbf{\footnotesize{Gatys \etal \cite{gatys2016image}:}} & \includegraphics[width=\x\textwidth]{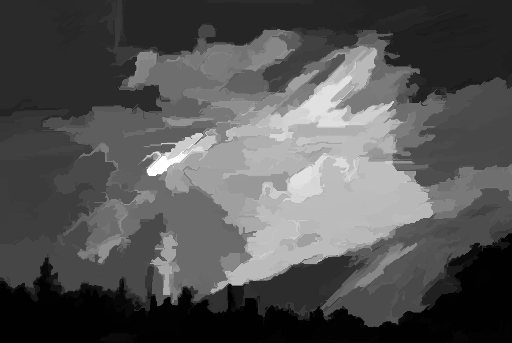} & \includegraphics[width=\x\textwidth]{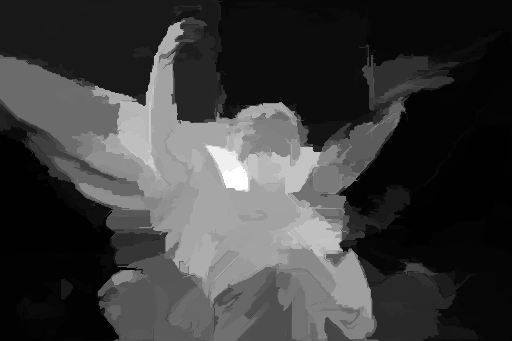}& \includegraphics[width=\x\textwidth]{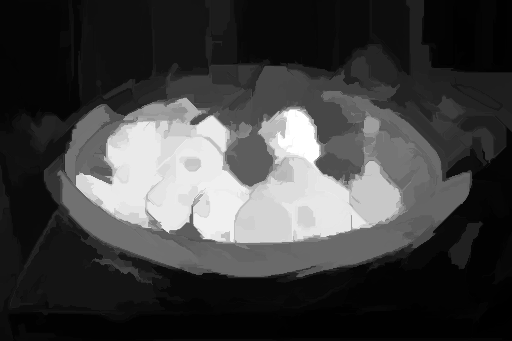} &\includegraphics[width=\x\textwidth]{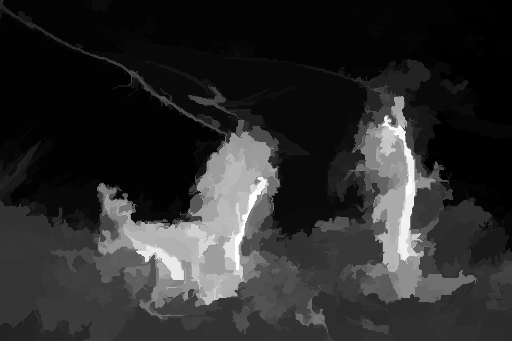} &\includegraphics[width=\x\textwidth]{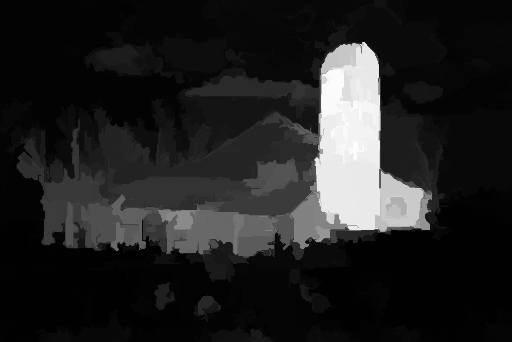} &\includegraphics[width=\x\textwidth]{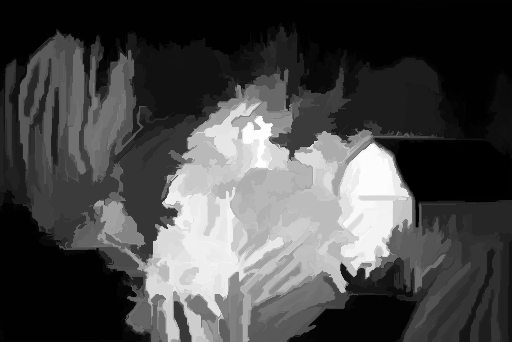} \\
%\toprule

\textbf{\footnotesize{Johnson \etal \cite{Johnson2016perceptual}:}} & \includegraphics[width=\x\textwidth]{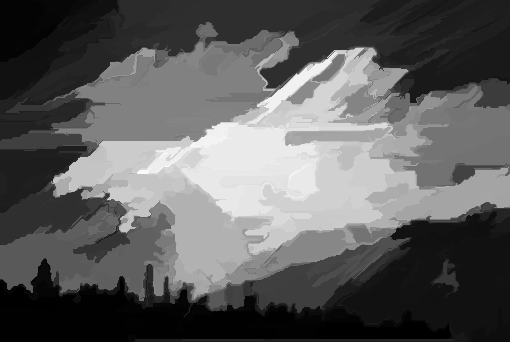} & \includegraphics[width=\x\textwidth]{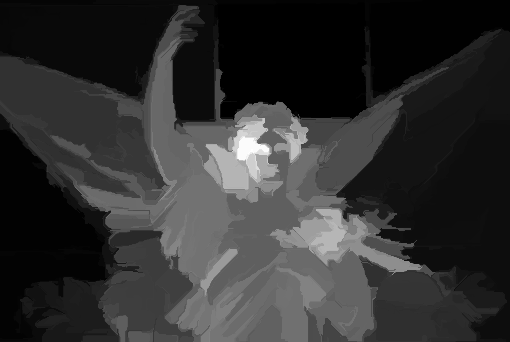} & \includegraphics[width=\x\textwidth]{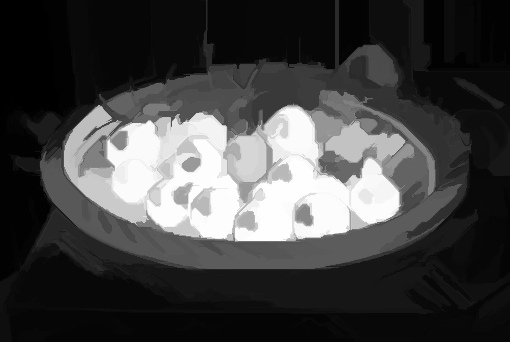}& \includegraphics[width=\x\textwidth]{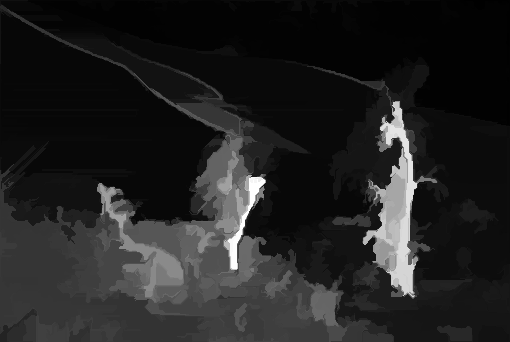}& \includegraphics[width=\x\textwidth]{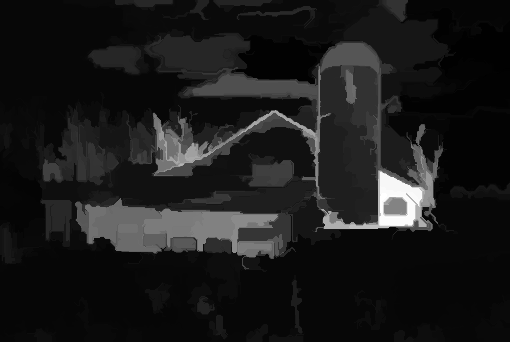}&  \includegraphics[width=\x\textwidth]{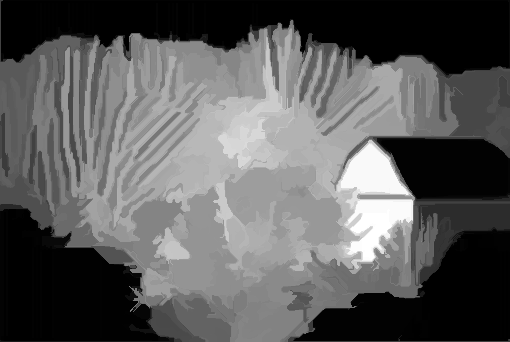}\\

\textbf{\footnotesize{Ulyanov \etal \cite{ulyanov2016texture}:}} & \includegraphics[width=\x\textwidth]{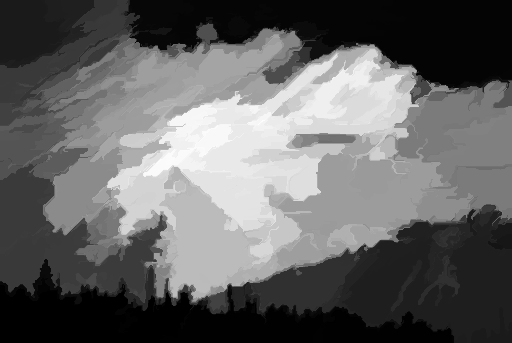} & \includegraphics[width=\x\textwidth]{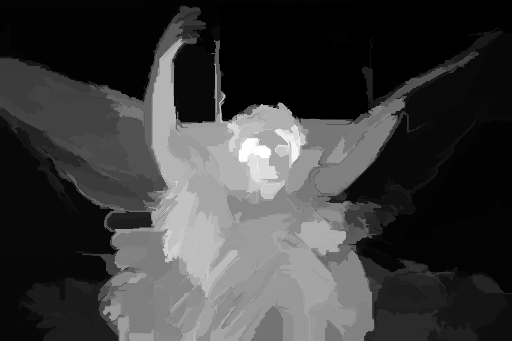} & \includegraphics[width=\x\textwidth]{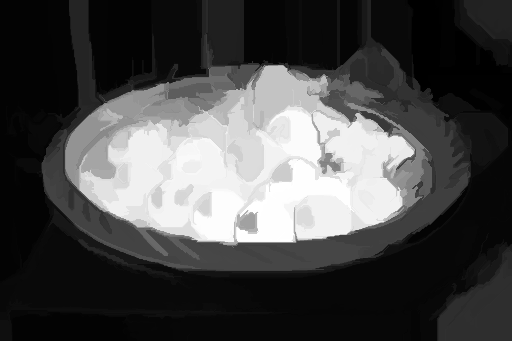}& \includegraphics[width=\x\textwidth]{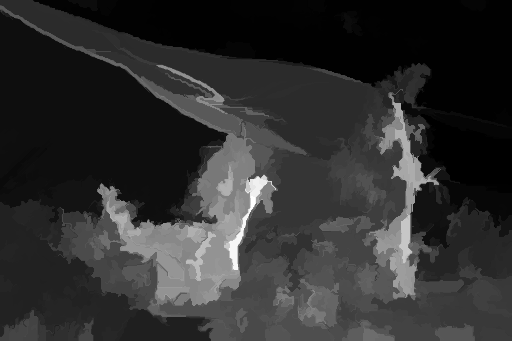}& \includegraphics[width=\x\textwidth]{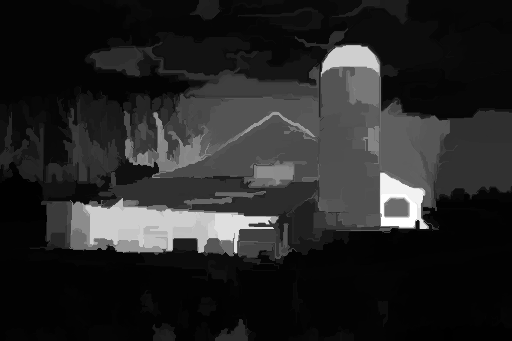}& \includegraphics[width=\x\textwidth]{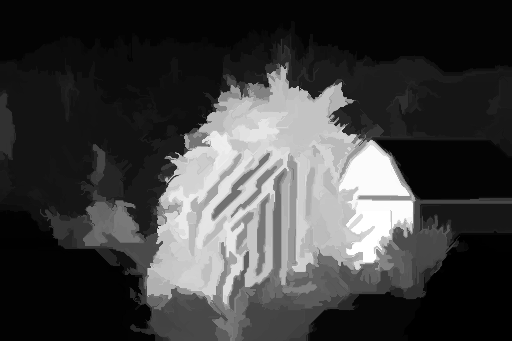}\\

\textbf{\footnotesize{Li and Wand \cite{li2016precomputed}:}} & \includegraphics[width=\x\textwidth]{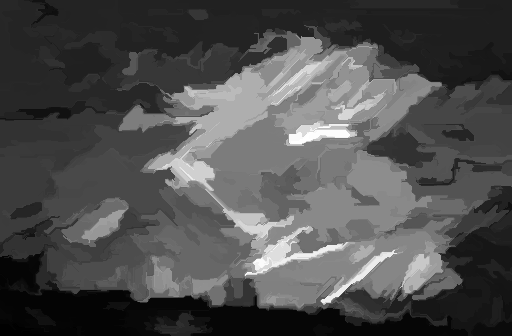} & \includegraphics[width=\x\textwidth]{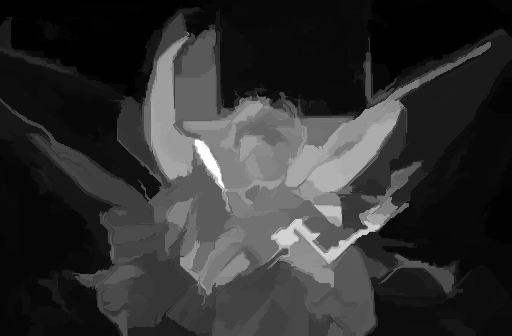} & \includegraphics[width=\x\textwidth]{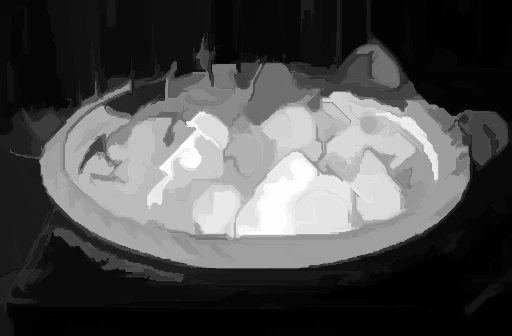}& \includegraphics[width=\x\textwidth]{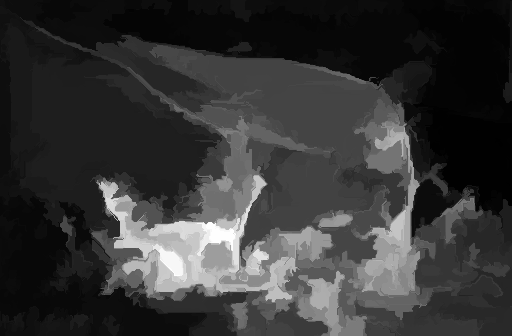}& \includegraphics[width=\x\textwidth]{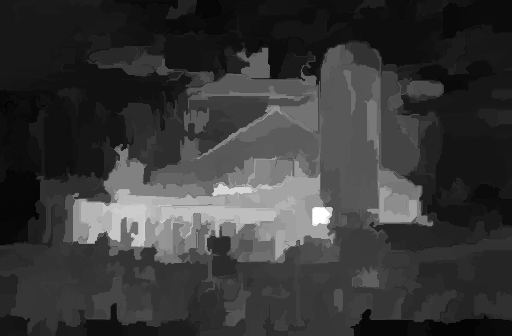}& \includegraphics[width=\x\textwidth]{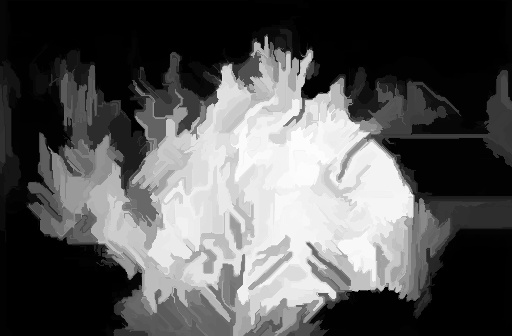}\\

\end{tabular}
}

\caption{Saliency detection results of \textbf{IOB-NST} and \textbf{PSPM-MOB-NST}, corresponding to Figure~\ref{fig:qualitativeresult1}. The results are produced by using the discriminative regional feature integration approach proposed by Wang \etal \cite{wang2017salient}.}
\label{fig:saliency1}
\end{figure*}

\begin{figure*}[!tbp]
\setlength\tabcolsep{1.5 pt}
{\renewcommand{\arraystretch}{0.9}
\begin{tabular}{m{\y} >{\centering}m{\z} >{\centering}m{\z} >{\centering}m{\z} >{\centering}m{\z} >{\centering}m{\z} >{\centering\arraybackslash}m{2.5cm}}
\centering

& \textbf{\footnotesize{Group \uppercase\expandafter{\romannumeral1}}} & \textbf{\footnotesize{Group \uppercase\expandafter{\romannumeral2}}} & \textbf{\footnotesize{Group \uppercase\expandafter{\romannumeral3}}} & \textbf{\footnotesize{Group \uppercase\expandafter{\romannumeral4}}} & \textbf{\footnotesize{Group \uppercase\expandafter{\romannumeral5}}} & \textbf{\footnotesize{Group \uppercase\expandafter{\romannumeral6}}}\\ %\vspace{1cm}

%\textbf{\small{Style:}}  & \includegraphics[width=\x\textwidth]{figs/style/1.pdf} & \includegraphics[width=\x\textwidth]{figs/style/2.pdf}& \includegraphics[width=\x\textwidth]{figs/style/4.pdf} & \includegraphics[width=\x\textwidth]{figs/style/5.pdf} & \includegraphics[width=\x\textwidth]{figs/style/6.pdf} & \includegraphics[width=\x\textwidth]{figs/style/7.pdf} & \includegraphics[width=\x\textwidth]{figs/style/9.pdf} & \includegraphics[width=\x\textwidth]{figs/style/10.pdf}\\

\textbf{\footnotesize{Content \& Style:}} & \includegraphics[width=\x\textwidth]{figs_tvcg/content/1.png} &\includegraphics[width=\x\textwidth]{figs_tvcg/content/2.png}&\includegraphics[width=\x\textwidth]{figs_tvcg/content/3.png} & \includegraphics[width=\x\textwidth]{figs_tvcg/content/4.png} &\includegraphics[width=\x\textwidth]{figs_tvcg/content/5.png} & \includegraphics[width=\x\textwidth]{figs_tvcg/content/6.png} \\

%\textbf{\footnotesize{Li and Wand \cite{li2016precomputed}:}} & \includegraphics[width=\x\textwidth]{figs_tvcg/li_gan/style1/4_MGANs.jpg}& \includegraphics[width=\x\textwidth]{figs_tvcg/li_gan/style2/1_MGANs.jpg}& \includegraphics[width=\x\textwidth]{figs_tvcg/li_gan/style3/5_MGANs.jpg}& \includegraphics[width=\x\textwidth]{figs_tvcg/li_gan/style5/8_MGANs.jpg}& \includegraphics[width=\x\textwidth]{figs_tvcg/li_gan/style9/9_MGANs.jpg}& \includegraphics[width=\x\textwidth]{figs_tvcg/li_gan/style10/15_MGANs.jpg}\\

\multirow{1}{1.5cm}{\textbf{\footnotesize{Dumoulin \etal \cite{dumoulin2016learned}:}}} & \includegraphics[width=\x\textwidth]{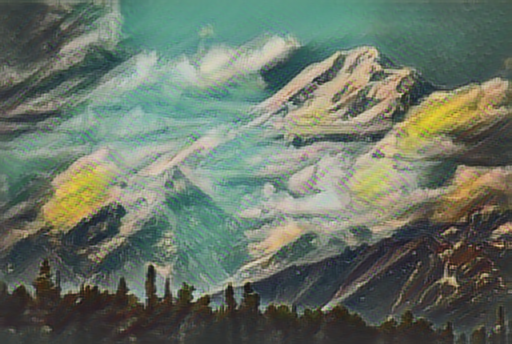} & \includegraphics[width=\x\textwidth]{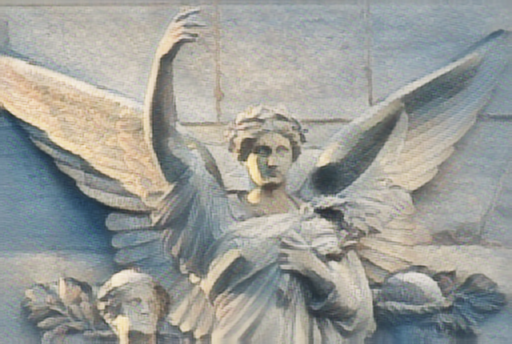} & \includegraphics[width=\x\textwidth]{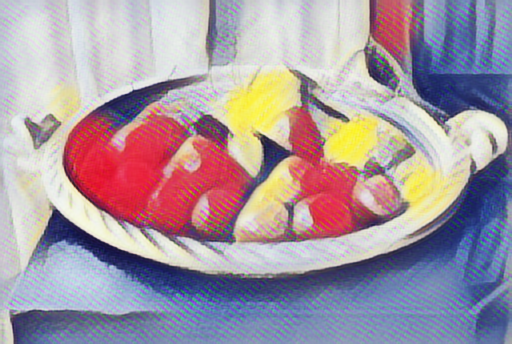}& \includegraphics[width=\x\textwidth]{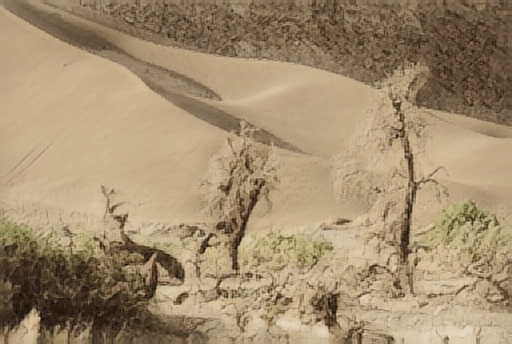}& \includegraphics[width=\x\textwidth]{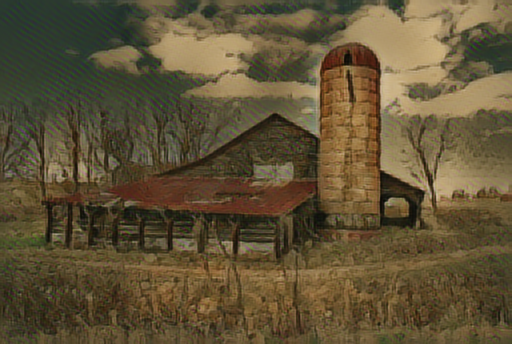}& \includegraphics[width=\x\textwidth]{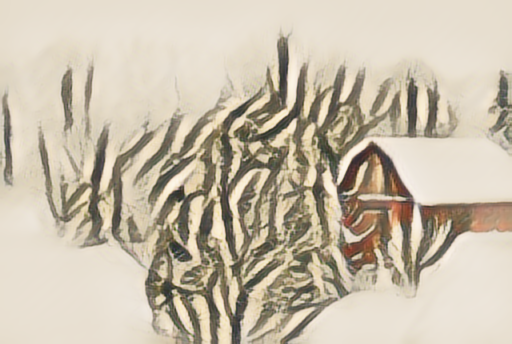}\\

\textbf{\footnotesize{Chen \etal \cite{chen2017stylebank}:}} & \includegraphics[width=\x\textwidth]{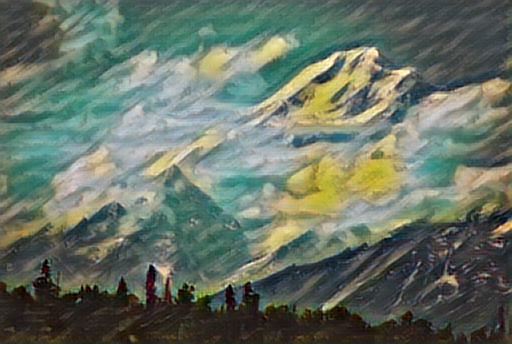} & \includegraphics[width=\x\textwidth]{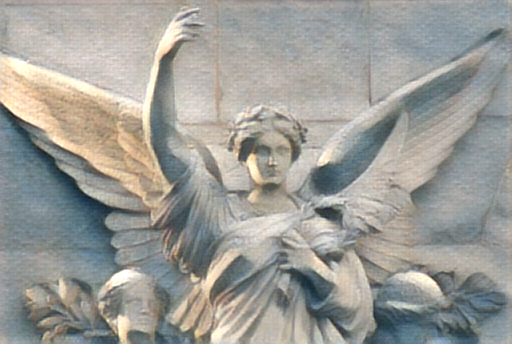} & \includegraphics[width=\x\textwidth]{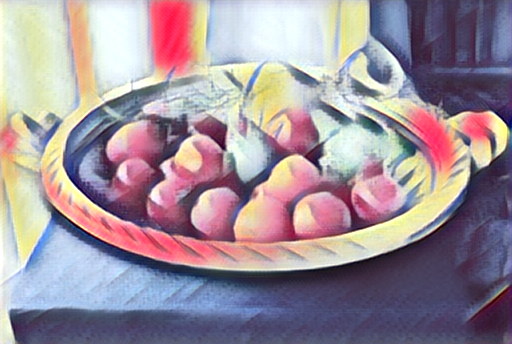}& \includegraphics[width=\x\textwidth]{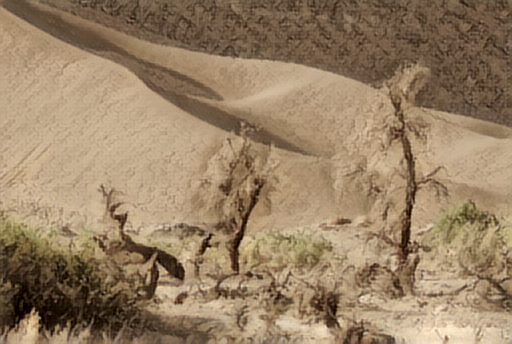}& \includegraphics[width=\x\textwidth]{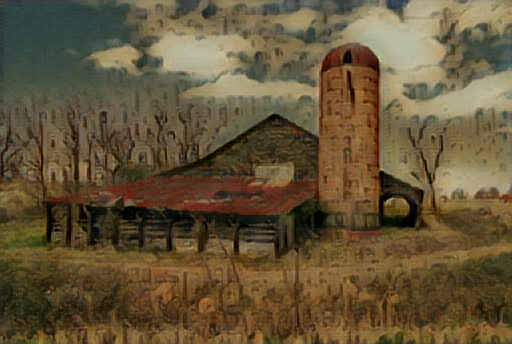}& \includegraphics[width=\x\textwidth]{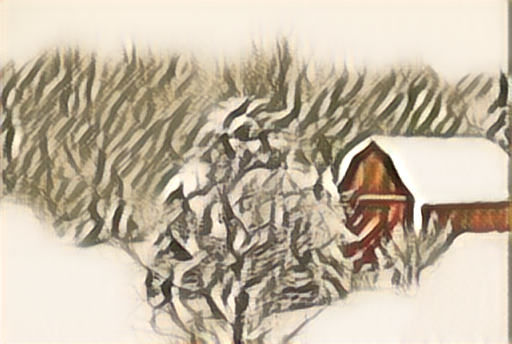}\\

\textbf{\footnotesize{Li \etal \cite{li2017diverse}:}} & \includegraphics[width=\x\textwidth]{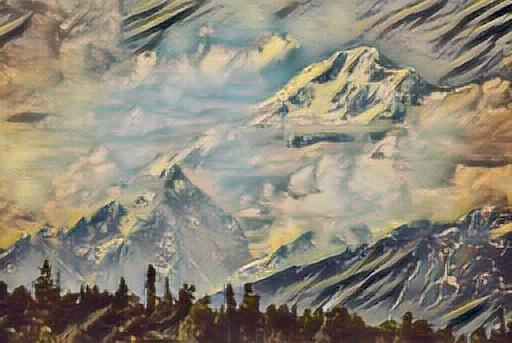} & \includegraphics[width=\x\textwidth]{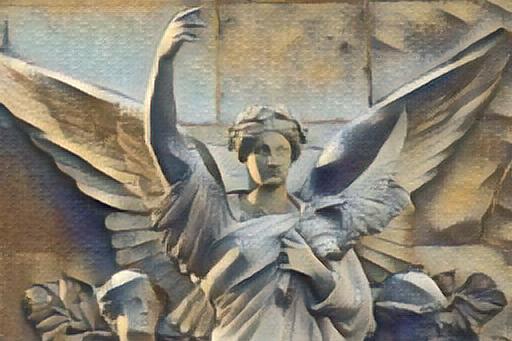} & \includegraphics[width=\x\textwidth]{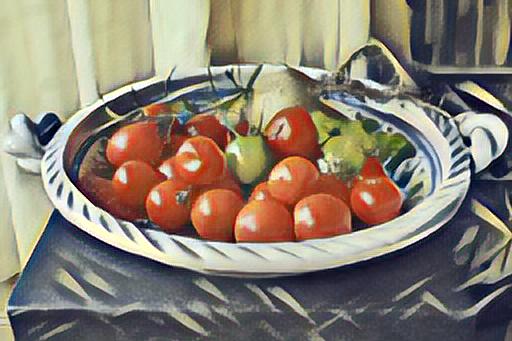}& \includegraphics[width=\x\textwidth]{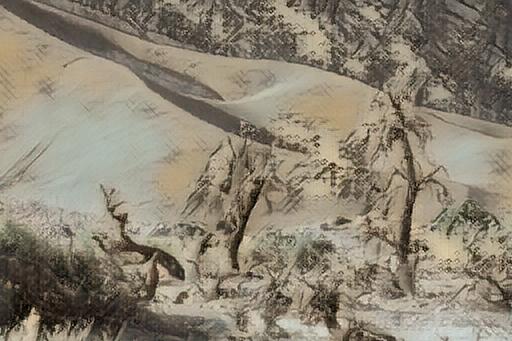}& \includegraphics[width=\x\textwidth]{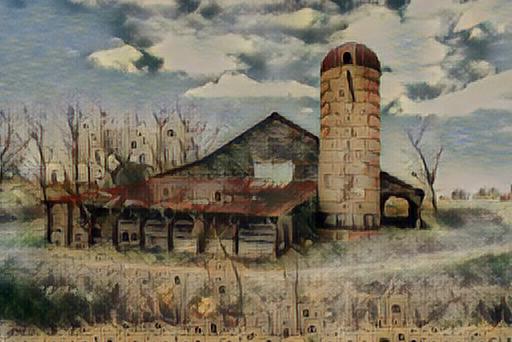}& \includegraphics[width=\x\textwidth]{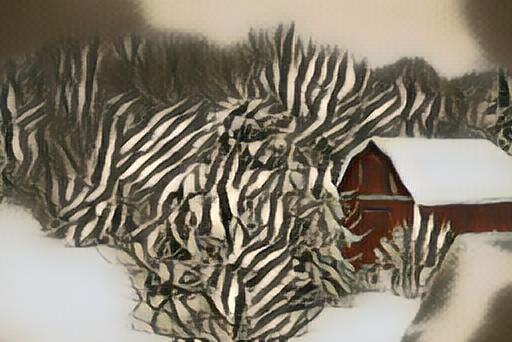}\\

%\textbf{\small{Li \etal \cite{li2017diverse}:}} & \includegraphics[width=\x\textwidth]{figs_tvcg/li_diverse/4_1_1.jpg}& \includegraphics[width=\x\textwidth]{figs_tvcg/li_diverse/1_1_2.jpg}& \includegraphics[width=\x\textwidth]{figs_tvcg/li_diverse/5_1_3.jpg}& \includegraphics[width=\x\textwidth]{figs_tvcg/li_diverse/8_1_5.jpg}& \includegraphics[width=\x\textwidth]{figs_tvcg/li_diverse/9_1_9.jpg}& \includegraphics[width=\x\textwidth]{figs_tvcg/li_diverse/15_1_10.jpg}\\
%

\multirow{1}{2.5cm}{\textbf{\footnotesize{Zhang and Dana \cite{zhang2017multi}:}}} & \includegraphics[width=\x\textwidth]{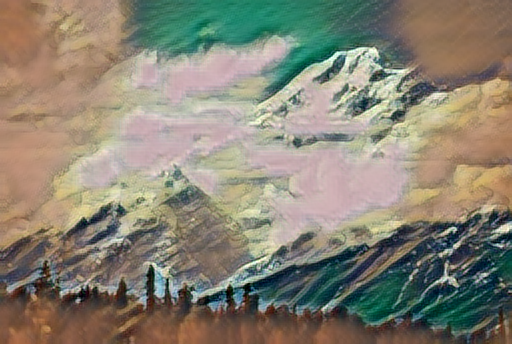} & \includegraphics[width=\x\textwidth]{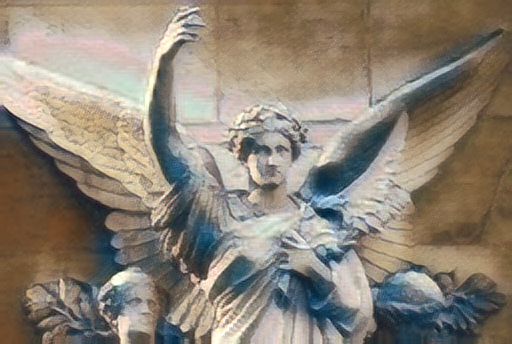} & \includegraphics[width=\x\textwidth]{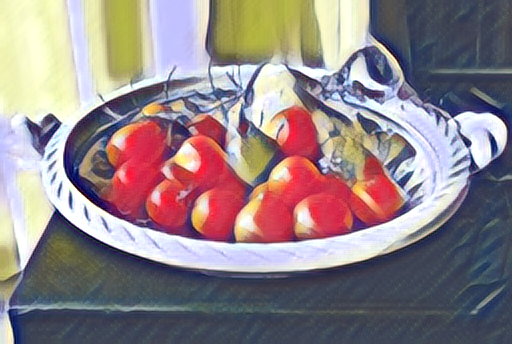}& \includegraphics[width=\x\textwidth]{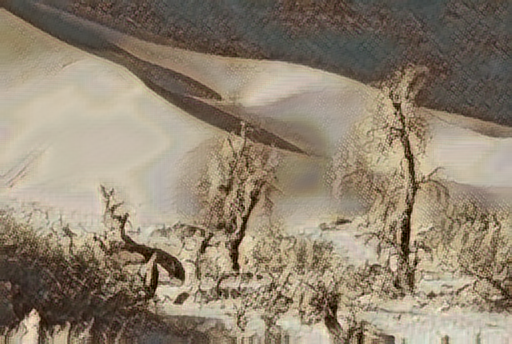}& \includegraphics[width=\x\textwidth]{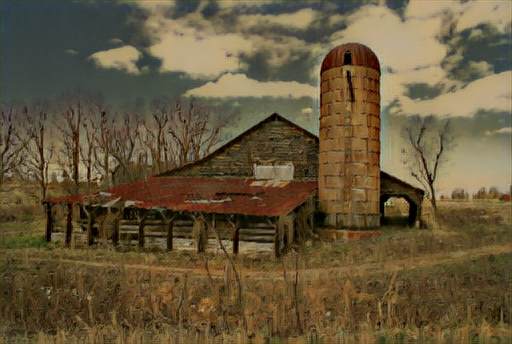}& \includegraphics[width=\x\textwidth]{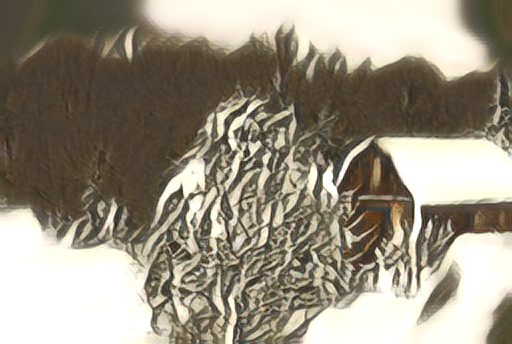}\\

%\textbf{\footnotesize{Chen and Schmidt \cite{chen2016fast}:}} & \includegraphics[width=\x\textwidth]{figs_tvcg/swap/style1/4_stylized.png} & \includegraphics[width=\x\textwidth]{figs_tvcg/swap/style2/1_stylized.png} & \includegraphics[width=\x\textwidth]{figs_tvcg/swap/style3/5_stylized.png}& \includegraphics[width=\x\textwidth]{figs_tvcg/swap/style5/8_stylized.png}& \includegraphics[width=\x\textwidth]{figs_tvcg/swap/style9/9_stylized.png}& \includegraphics[width=\x\textwidth]{figs_tvcg/swap/style10/15_stylized.png}\\

\end{tabular}
}

\caption{Some example results of \textbf{MSPM-MOB-NST} for qualitative evaluation. The content images are from the benchmark dataset proposed by Mould and Rosin \cite{mould2016benchmark,mould2017developing}. The style images are in the public domain. Detailed information of our style images can be found in Table~\ref{table:styleimage}.}
\label{fig:qualitativeresult2}
\end{figure*}
%%%%%%%%%%%
\begin{figure*}[!tbp]
\setlength\tabcolsep{1.5 pt}
{\renewcommand{\arraystretch}{0.9}
\begin{tabular}{m{\y} >{\centering}m{\z} >{\centering}m{\z} >{\centering}m{\z} >{\centering}m{\z} >{\centering}m{\z} >{\centering\arraybackslash}m{2.5cm}}
\centering

& \textbf{\footnotesize{Group \uppercase\expandafter{\romannumeral1}}} & \textbf{\footnotesize{Group \uppercase\expandafter{\romannumeral2}}} & \textbf{\footnotesize{Group \uppercase\expandafter{\romannumeral3}}} & \textbf{\footnotesize{Group \uppercase\expandafter{\romannumeral4}}} & \textbf{\footnotesize{Group \uppercase\expandafter{\romannumeral5}}} & \textbf{\footnotesize{Group \uppercase\expandafter{\romannumeral6}}}\\ %\vspace{1cm}

%\textbf{\small{Style:}}  & \includegraphics[width=\x\textwidth]{figs/style/1.pdf} & \includegraphics[width=\x\textwidth]{figs/style/2.pdf}& \includegraphics[width=\x\textwidth]{figs/style/4.pdf} & \includegraphics[width=\x\textwidth]{figs/style/5.pdf} & \includegraphics[width=\x\textwidth]{figs/style/6.pdf} & \includegraphics[width=\x\textwidth]{figs/style/7.pdf} & \includegraphics[width=\x\textwidth]{figs/style/9.pdf} & \includegraphics[width=\x\textwidth]{figs/style/10.pdf}\\

\textbf{\footnotesize{Content:}} & \includegraphics[width=\x\textwidth]{saliency_tvcg/content/4.png} &\includegraphics[width=\x\textwidth]{saliency_tvcg/content/1.png}&\includegraphics[width=\x\textwidth]{saliency_tvcg/content/5.png} & \includegraphics[width=\x\textwidth]{saliency_tvcg/content/8.png} &\includegraphics[width=\x\textwidth]{saliency_tvcg/content/9.png} & \includegraphics[width=\x\textwidth]{saliency_tvcg/content/15.png} \\

%\textbf{\footnotesize{Li and Wand \cite{li2016precomputed}:}} & \includegraphics[width=\x\textwidth]{saliency_tvcg/li_gan/style1/4_MGANs.jpg}& \includegraphics[width=\x\textwidth]{saliency_tvcg/li_gan/style2/1_MGANs.jpg}& \includegraphics[width=\x\textwidth]{saliency_tvcg/li_gan/style3/5_MGANs.jpg}& \includegraphics[width=\x\textwidth]{saliency_tvcg/li_gan/style5/8_MGANs.jpg}& \includegraphics[width=\x\textwidth]{saliency_tvcg/li_gan/style9/9_MGANs.jpg}& \includegraphics[width=\x\textwidth]{saliency_tvcg/li_gan/style10/15_MGANs.jpg}\\

\multirow{1}{1.5cm}{\textbf{\footnotesize{Dumoulin \etal \cite{dumoulin2016learned}:}}} & \includegraphics[width=\x\textwidth]{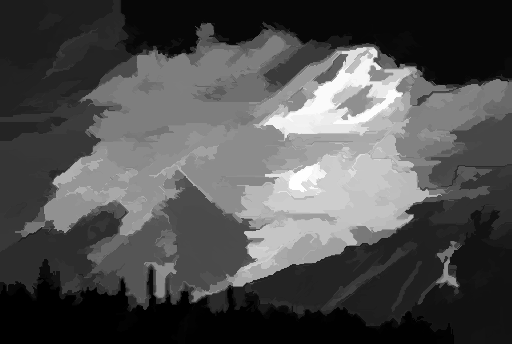} & \includegraphics[width=\x\textwidth]{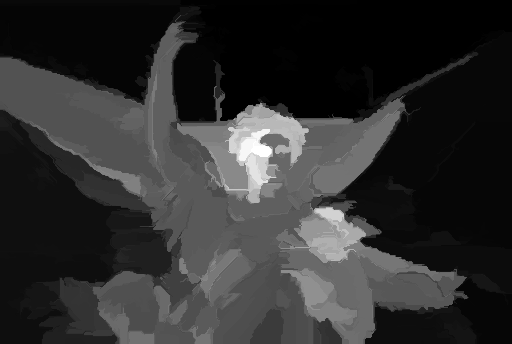} & \includegraphics[width=\x\textwidth]{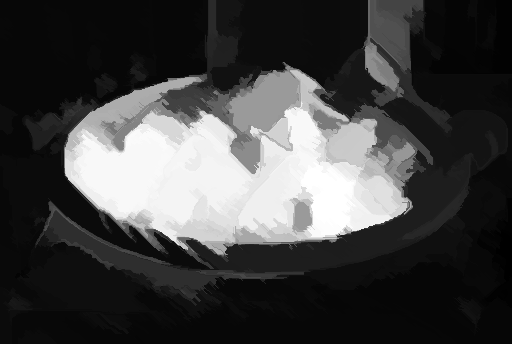}& \includegraphics[width=\x\textwidth]{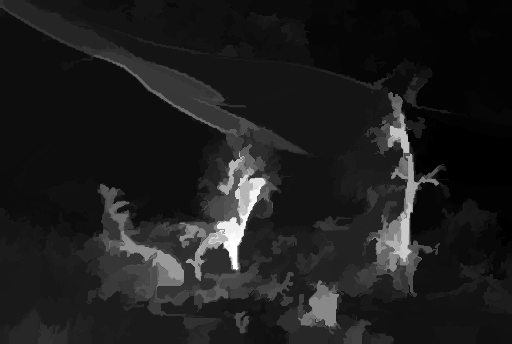}& \includegraphics[width=\x\textwidth]{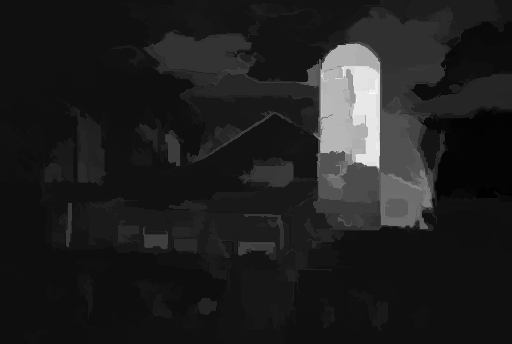}& \includegraphics[width=\x\textwidth]{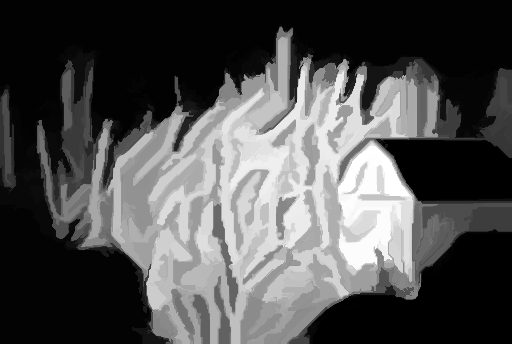}\\

\textbf{\footnotesize{Chen \etal \cite{chen2017stylebank}:}} & \includegraphics[width=\x\textwidth]{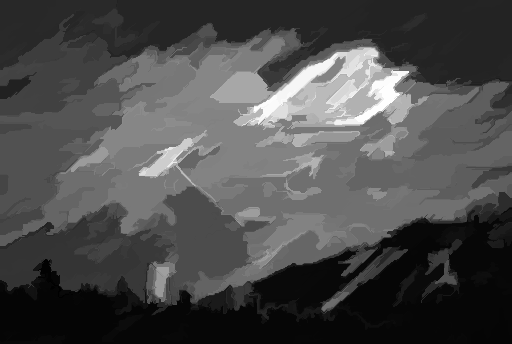} & \includegraphics[width=\x\textwidth]{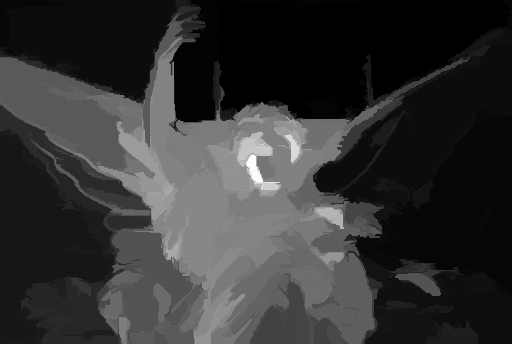} & \includegraphics[width=\x\textwidth]{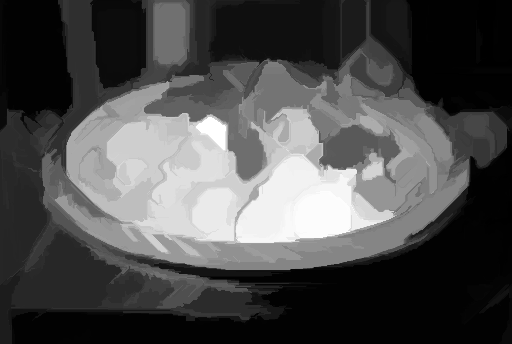}& \includegraphics[width=\x\textwidth]{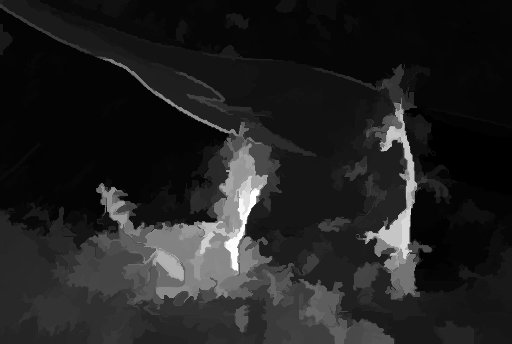}& \includegraphics[width=\x\textwidth]{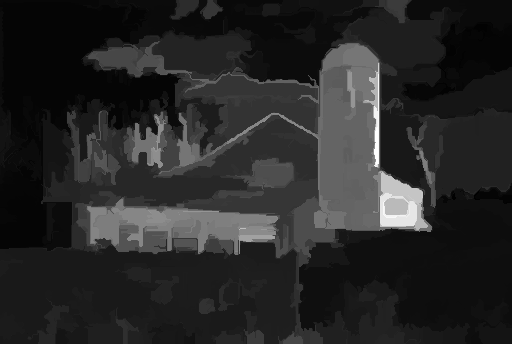}& \includegraphics[width=\x\textwidth]{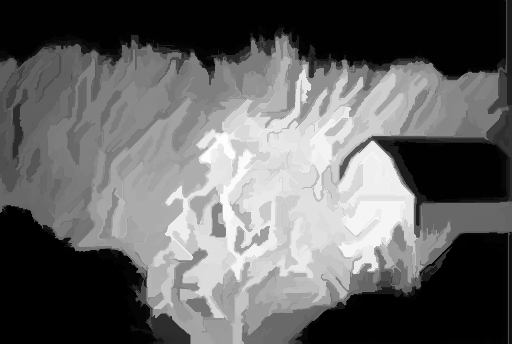}\\

\textbf{\footnotesize{Li \etal \cite{li2017diverse}:}} & \includegraphics[width=\x\textwidth]{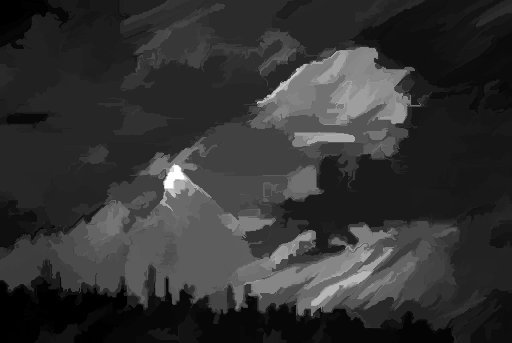} & \includegraphics[width=\x\textwidth]{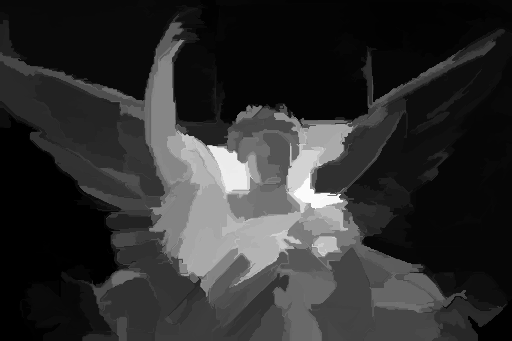} & \includegraphics[width=\x\textwidth]{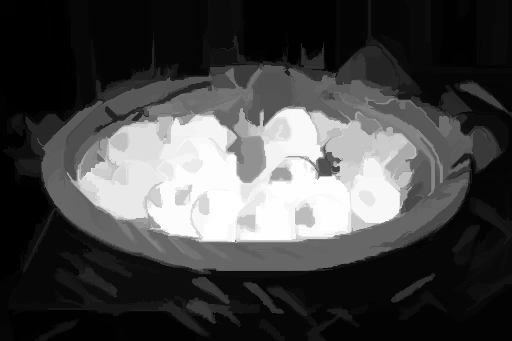}& \includegraphics[width=\x\textwidth]{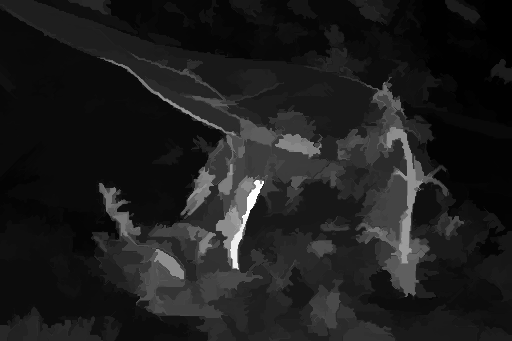}& \includegraphics[width=\x\textwidth]{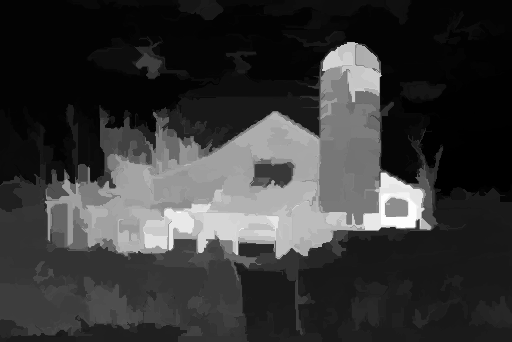}& \includegraphics[width=\x\textwidth]{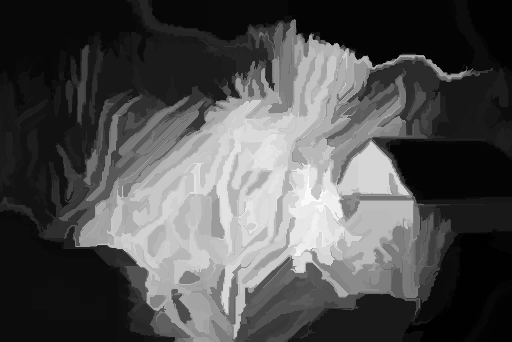}\\

%\textbf{\small{Li \etal \cite{li2017diverse}:}} & \includegraphics[width=\x\textwidth]{saliency_tvcg/li_diverse/4_1_1.jpg}& \includegraphics[width=\x\textwidth]{saliency_tvcg/li_diverse/1_1_2.jpg}& \includegraphics[width=\x\textwidth]{saliency_tvcg/li_diverse/5_1_3.jpg}& \includegraphics[width=\x\textwidth]{saliency_tvcg/li_diverse/8_1_5.jpg}& \includegraphics[width=\x\textwidth]{saliency_tvcg/li_diverse/9_1_9.jpg}& \includegraphics[width=\x\textwidth]{saliency_tvcg/li_diverse/15_1_10.jpg}\\
%

\multirow{1}{2.5cm}{\textbf{\footnotesize{Zhang and Dana \cite{zhang2017multi}:}}} & \includegraphics[width=\x\textwidth]{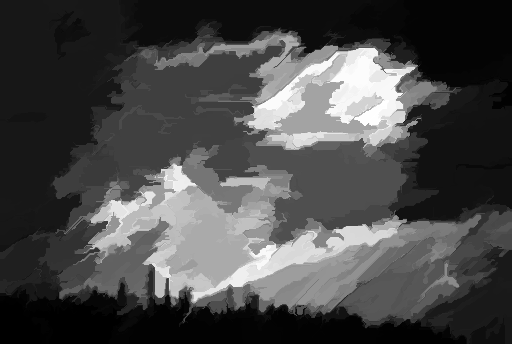} & \includegraphics[width=\x\textwidth]{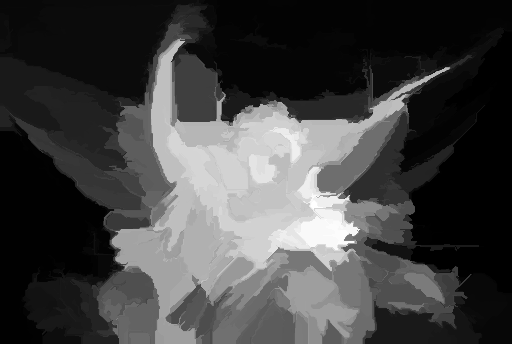} & \includegraphics[width=\x\textwidth]{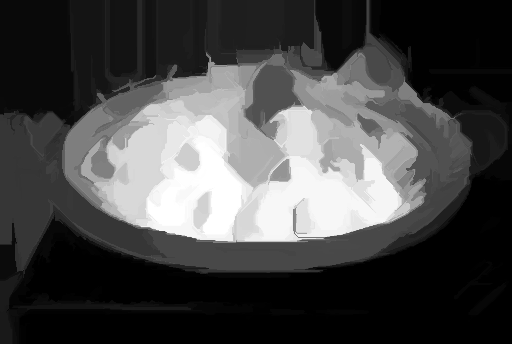}& \includegraphics[width=\x\textwidth]{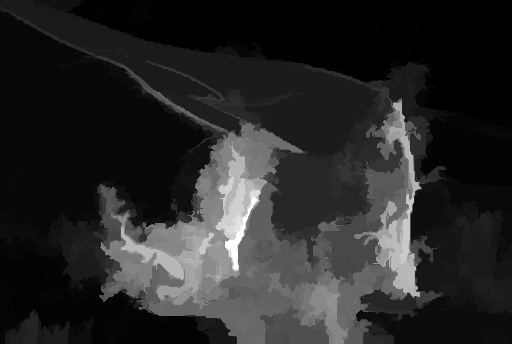}& \includegraphics[width=\x\textwidth]{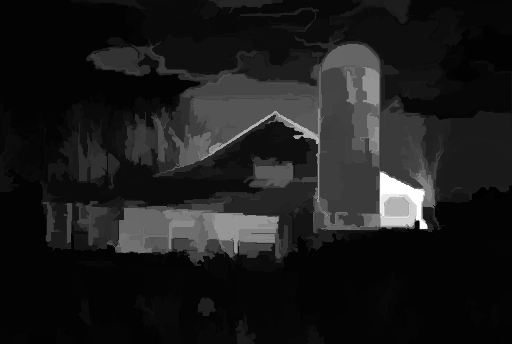}& \includegraphics[width=\x\textwidth]{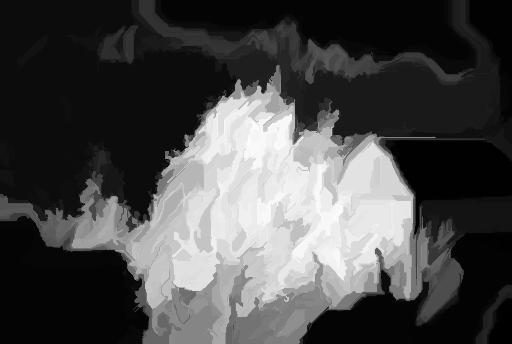}\\

%\textbf{\footnotesize{Chen and Schmidt \cite{chen2016fast}:}} & \includegraphics[width=\x\textwidth]{saliency_tvcg/swap/style1/4_stylized.png} & \includegraphics[width=\x\textwidth]{saliency_tvcg/swap/style2/1_stylized.png} & \includegraphics[width=\x\textwidth]{saliency_tvcg/swap/style3/5_stylized.png}& \includegraphics[width=\x\textwidth]{saliency_tvcg/swap/style5/8_stylized.png}& \includegraphics[width=\x\textwidth]{saliency_tvcg/swap/style9/9_stylized.png}& \includegraphics[width=\x\textwidth]{saliency_tvcg/swap/style10/15_stylized.png}\\

\end{tabular}
}

%\caption{Some example results of \textbf{MSPM-MOB-NST} for qualitative evaluation. The content images are from the benchmark dataset proposed by Mould and Rosin \cite{mould2016benchmark,mould2017developing}. The style images are in the public domain. Detailed information of our style images can be found in Table~\ref{table:styleimage}.}
\caption{Saliency detection results of \textbf{MSPM-MOB-NST}, corresponding to Figure~\ref{fig:qualitativeresult2}. The results are produced by using the discriminative regional feature integration approach proposed by Wang \etal \cite{wang2017salient}.}
\label{fig:saliency2}
\end{figure*}

%%%%%%%%%%%
\begin{figure*}[!tbp]
\setlength\tabcolsep{1.5 pt}
{\renewcommand{\arraystretch}{0.9}
\begin{tabular}{m{\y} >{\centering}m{\z} >{\centering}m{\z} >{\centering}m{\z} >{\centering}m{\z} >{\centering}m{\z} >{\centering\arraybackslash}m{2.5cm}}
\centering

& \textbf{\footnotesize{Group \uppercase\expandafter{\romannumeral1}}} & \textbf{\footnotesize{Group \uppercase\expandafter{\romannumeral2}}} & \textbf{\footnotesize{Group \uppercase\expandafter{\romannumeral3}}} & \textbf{\footnotesize{Group \uppercase\expandafter{\romannumeral4}}} & \textbf{\footnotesize{Group \uppercase\expandafter{\romannumeral5}}} & \textbf{\footnotesize{Group \uppercase\expandafter{\romannumeral6}}}\\ %\vspace{1cm}

%\textbf{\small{Style:}}  & \includegraphics[width=\x\textwidth]{figs/style/1.pdf} & \includegraphics[width=\x\textwidth]{figs/style/2.pdf}& \includegraphics[width=\x\textwidth]{figs/style/4.pdf} & \includegraphics[width=\x\textwidth]{figs/style/5.pdf} & \includegraphics[width=\x\textwidth]{figs/style/6.pdf} & \includegraphics[width=\x\textwidth]{figs/style/7.pdf} & \includegraphics[width=\x\textwidth]{figs/style/9.pdf} & \includegraphics[width=\x\textwidth]{figs/style/10.pdf}\\

\textbf{\footnotesize{Content \& Style:}} & \includegraphics[width=\x\textwidth]{figs_tvcg/content/1.png} &\includegraphics[width=\x\textwidth]{figs_tvcg/content/2.png}&\includegraphics[width=\x\textwidth]{figs_tvcg/content/3.png} & \includegraphics[width=\x\textwidth]{figs_tvcg/content/4.png} &\includegraphics[width=\x\textwidth]{figs_tvcg/content/5.png} & \includegraphics[width=\x\textwidth]{figs_tvcg/content/6.png} \\

%\toprule
\textbf{\footnotesize{Chen and Schmidt \cite{chen2016fast}:}} & \includegraphics[width=\x\textwidth]{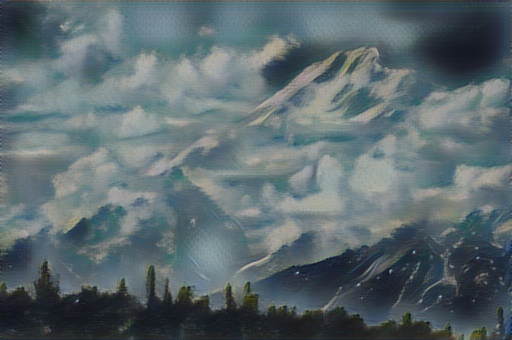} & \includegraphics[width=\x\textwidth]{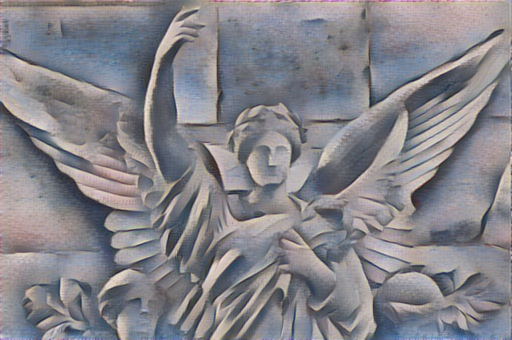} & \includegraphics[width=\x\textwidth]{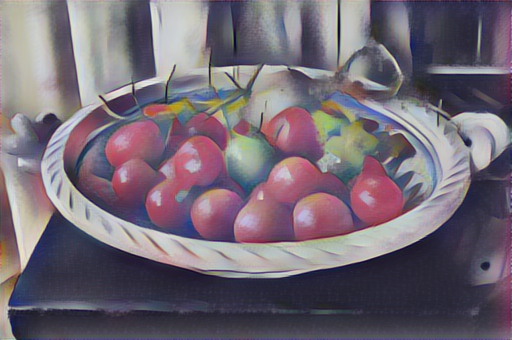}& \includegraphics[width=\x\textwidth]{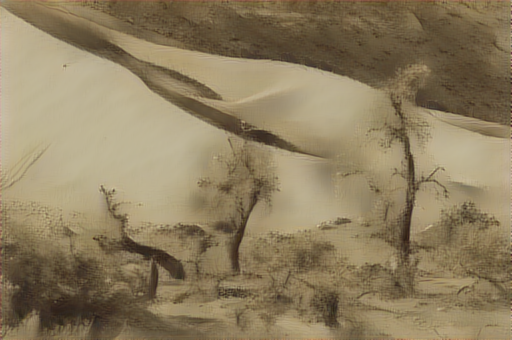}& \includegraphics[width=\x\textwidth]{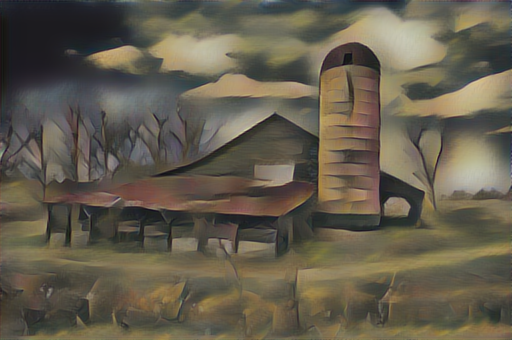}& \includegraphics[width=\x\textwidth]{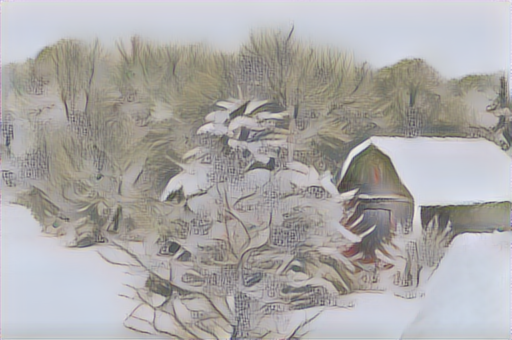}\\

\textbf{\footnotesize{Ghiasi \etal \cite{ghiasi2017exploring}:}} & \includegraphics[width=\x\textwidth]{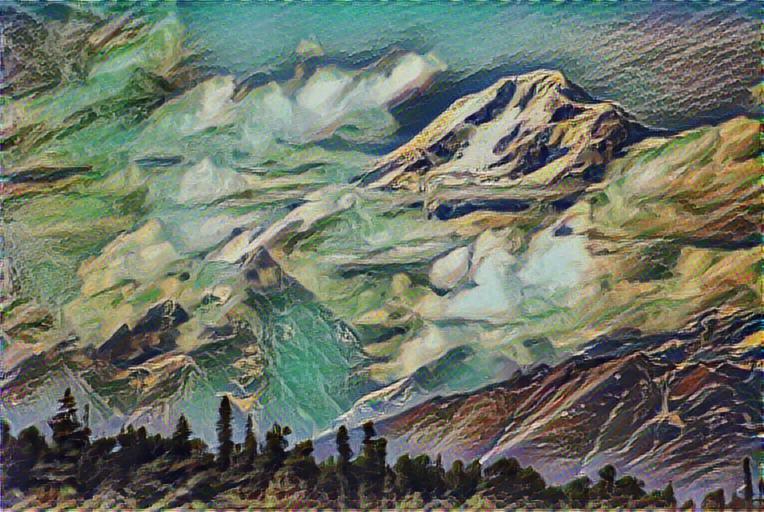} & \includegraphics[width=\x\textwidth]{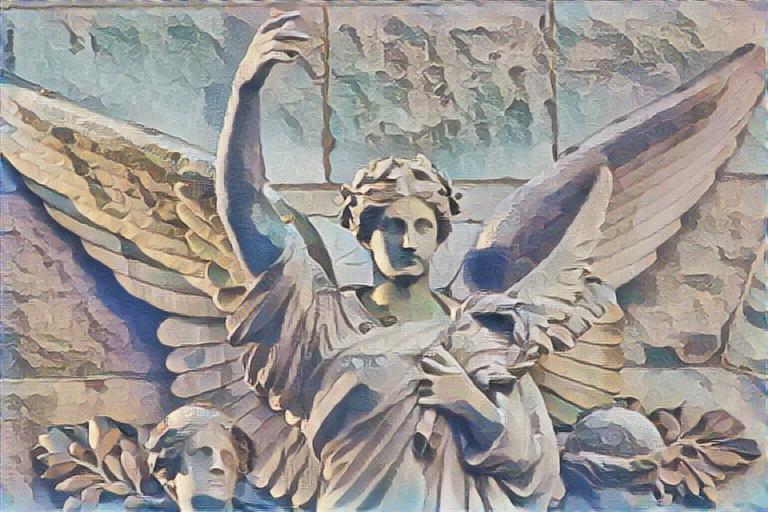} & \includegraphics[width=\x\textwidth]{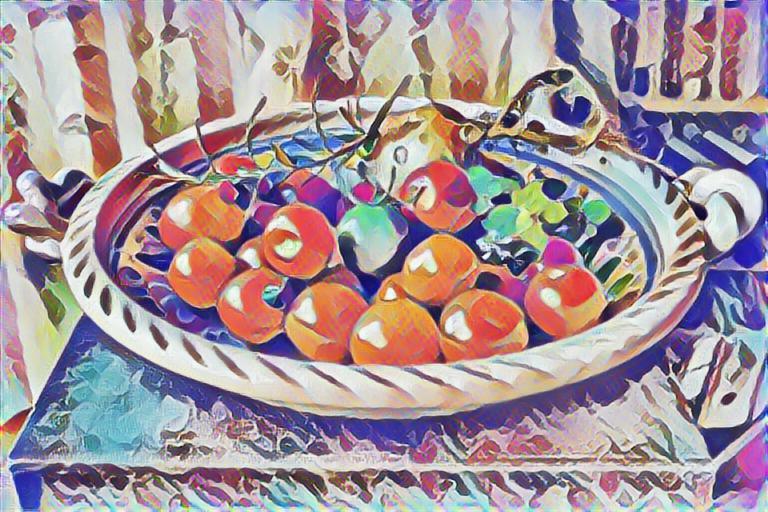}& \includegraphics[width=\x\textwidth]{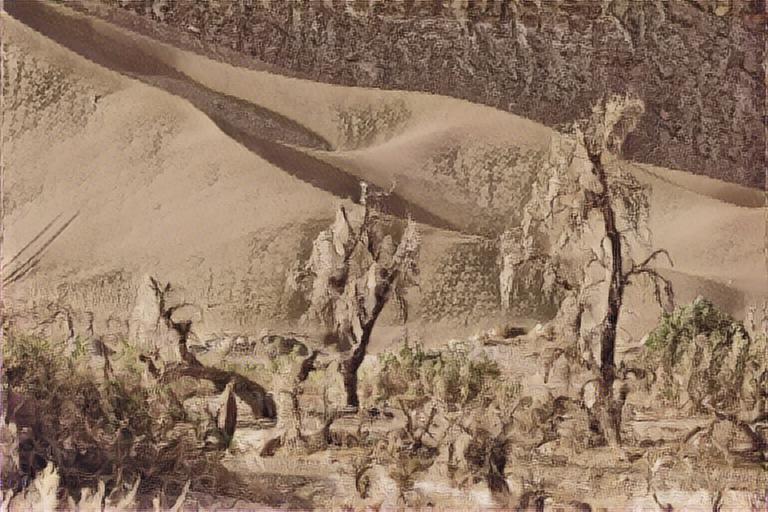}& \includegraphics[width=\x\textwidth]{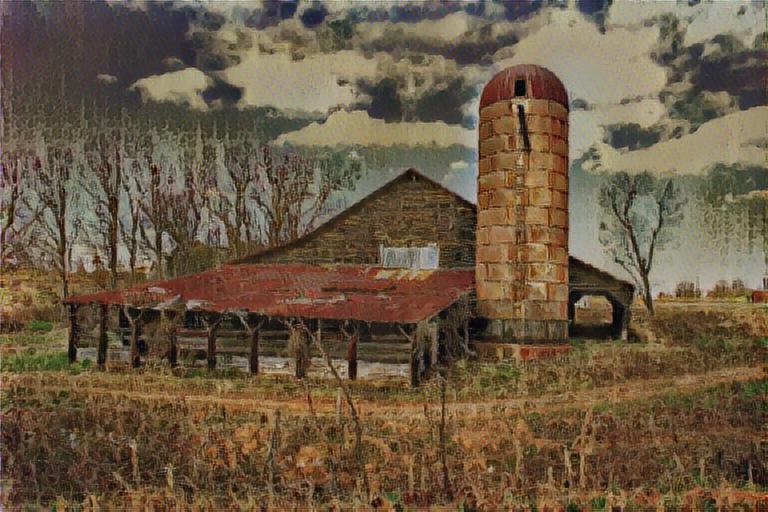}& \includegraphics[width=\x\textwidth]{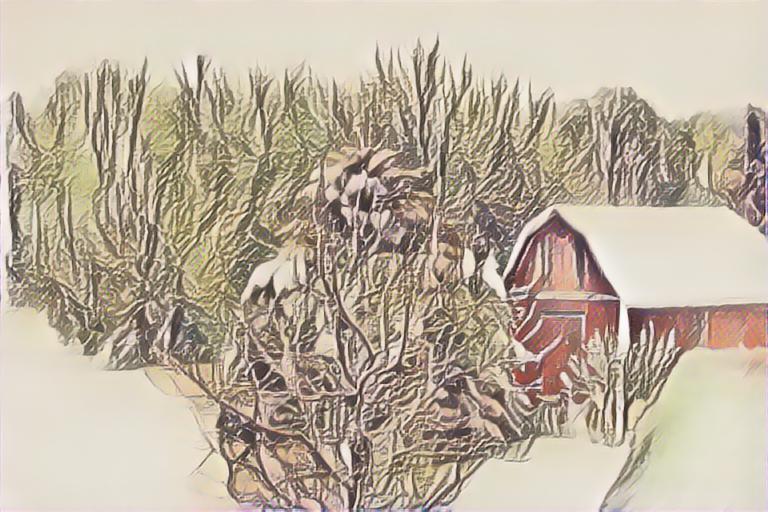}\\

%\textbf{\footnotesize{Ghiasi \etal \cite{ghiasi2017exploring}:}} & \includegraphics[width=\x\textwidth]{figs_tvcg/bmvc/4_stylized_1_0.jpg} & \includegraphics[width=\x\textwidth]{figs_tvcg/bmvc/1_stylized_2_0.jpg} & \includegraphics[width=\x\textwidth]{figs_tvcg/bmvc/5_stylized_3_0.jpg}& \includegraphics[width=\x\textwidth]{figs_tvcg/bmvc/8_stylized_5_0.jpg}& \includegraphics[width=\x\textwidth]{figs_tvcg/bmvc/9_stylized_9_0.jpg}& \includegraphics[width=\x\textwidth]{figs_tvcg/bmvc/10_stylized_10_0.jpg}\\

\multirow{1}{2.5cm}{\textbf{\footnotesize{Huang and Belongie \cite{huang2017arbitrary}:}}} & \includegraphics[width=\x\textwidth]{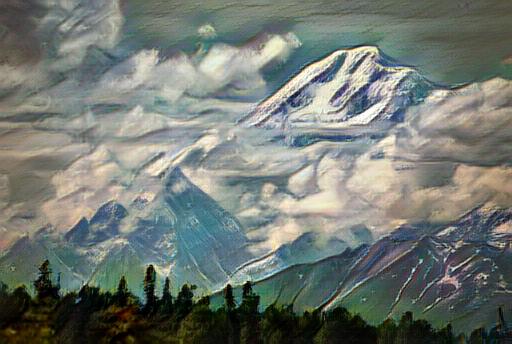} & \includegraphics[width=\x\textwidth]{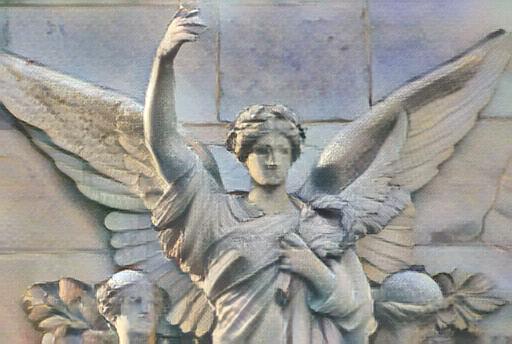} & \includegraphics[width=\x\textwidth]{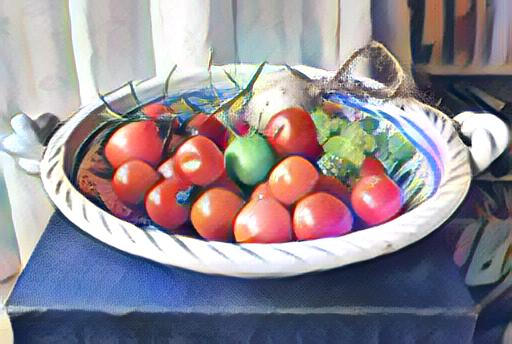}& \includegraphics[width=\x\textwidth]{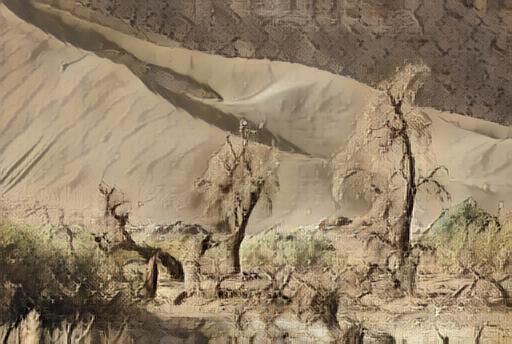}& \includegraphics[width=\x\textwidth]{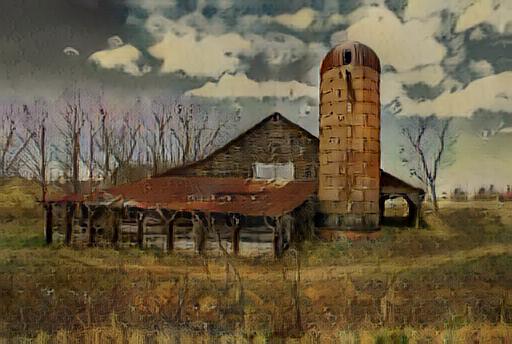}& \includegraphics[width=\x\textwidth]{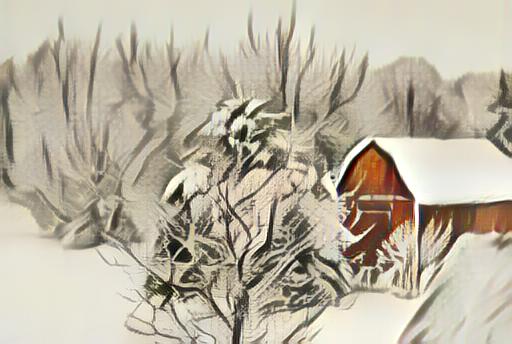}\\

%\multirow{1}{\y}{\textbf{\small{Huang and Belongie \cite{huang2017arbitrary}:}}} & \includegraphics[width=\x\textwidth]{figs_tvcg/xun/4_stylized_1.jpg}& \includegraphics[width=\x\textwidth]{figs_tvcg/xun/1_stylized_2.jpg}& \includegraphics[width=\x\textwidth]{figs_tvcg/xun/5_stylized_3.jpg}&  \includegraphics[width=\x\textwidth]{figs_tvcg/xun/8_stylized_5.jpg}& \includegraphics[width=\x\textwidth]{figs_tvcg/xun/9_stylized_9.jpg}& \includegraphics[width=\x\textwidth]{figs_tvcg/xun/10_stylized_10.jpg}\\

\textbf{\footnotesize{Li \etal \cite{li2017universal}:}} & \includegraphics[width=\x\textwidth]{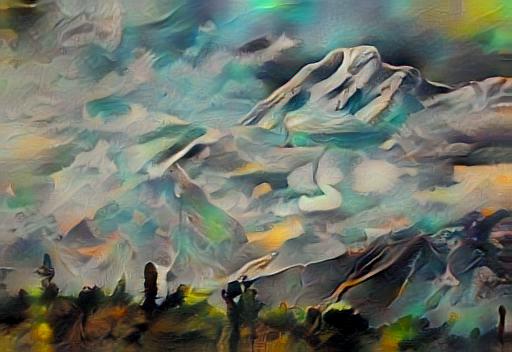} & \includegraphics[width=\x\textwidth]{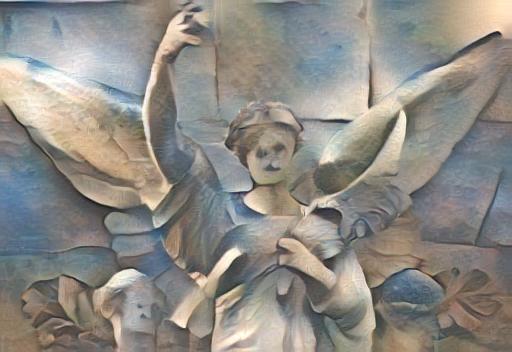} & \includegraphics[width=\x\textwidth]{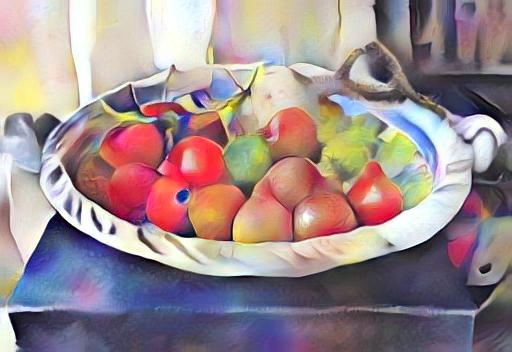}& \includegraphics[width=\x\textwidth]{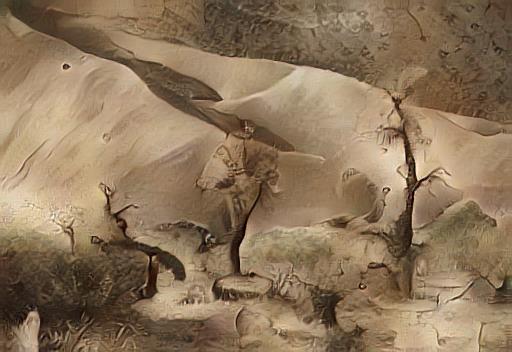}& \includegraphics[width=\x\textwidth]{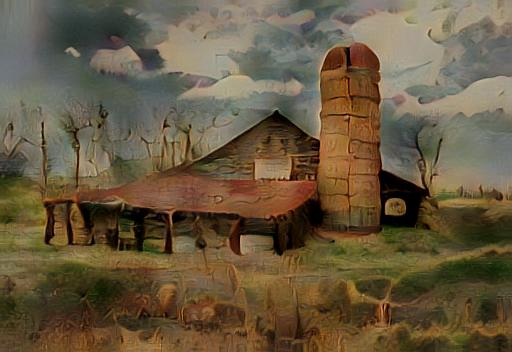}& \includegraphics[width=\x\textwidth]{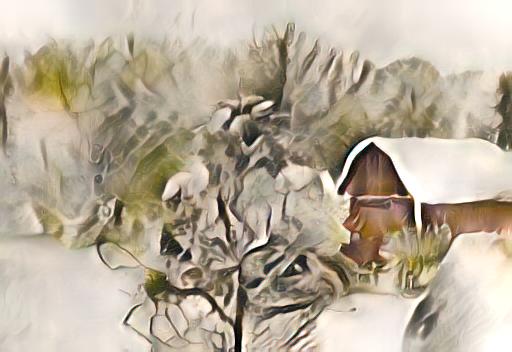}\\

\end{tabular}
}

\caption{Some example results of \textbf{ASPM-MOB-NST} for qualitative evaluation. The content images are from the benchmark dataset proposed by Mould and Rosin \cite{mould2016benchmark,mould2017developing}. The style images are in the public domain. Detailed information of our style images can be found in Table~\ref{table:styleimage}.}
\label{fig:qualitativeresult3}
\end{figure*}
%%%%%%%%%%%%%%%
%%%%%%%%%%%
\begin{figure*}[!tbp]
\setlength\tabcolsep{1.5 pt}
{\renewcommand{\arraystretch}{0.9}
\begin{tabular}{m{\y} >{\centering}m{\z} >{\centering}m{\z} >{\centering}m{\z} >{\centering}m{\z} >{\centering}m{\z} >{\centering\arraybackslash}m{2.5cm}}
\centering

& \textbf{\footnotesize{Group \uppercase\expandafter{\romannumeral1}}} & \textbf{\footnotesize{Group \uppercase\expandafter{\romannumeral2}}} & \textbf{\footnotesize{Group \uppercase\expandafter{\romannumeral3}}} & \textbf{\footnotesize{Group \uppercase\expandafter{\romannumeral4}}} & \textbf{\footnotesize{Group \uppercase\expandafter{\romannumeral5}}} & \textbf{\footnotesize{Group \uppercase\expandafter{\romannumeral6}}}\\ %\vspace{1cm}

%\textbf{\small{Style:}}  & \includegraphics[width=\x\textwidth]{figs/style/1.pdf} & \includegraphics[width=\x\textwidth]{figs/style/2.pdf}& \includegraphics[width=\x\textwidth]{figs/style/4.pdf} & \includegraphics[width=\x\textwidth]{figs/style/5.pdf} & \includegraphics[width=\x\textwidth]{figs/style/6.pdf} & \includegraphics[width=\x\textwidth]{figs/style/7.pdf} & \includegraphics[width=\x\textwidth]{figs/style/9.pdf} & \includegraphics[width=\x\textwidth]{figs/style/10.pdf}\\

\textbf{\footnotesize{Content:}} & \includegraphics[width=\x\textwidth]{saliency_tvcg/content/4.png} &\includegraphics[width=\x\textwidth]{saliency_tvcg/content/1.png}&\includegraphics[width=\x\textwidth]{saliency_tvcg/content/5.png} & \includegraphics[width=\x\textwidth]{saliency_tvcg/content/8.png} &\includegraphics[width=\x\textwidth]{saliency_tvcg/content/9.png} & \includegraphics[width=\x\textwidth]{saliency_tvcg/content/15.png} \\

%\toprule
\textbf{\footnotesize{Chen and Schmidt \cite{chen2016fast}:}} & \includegraphics[width=\x\textwidth]{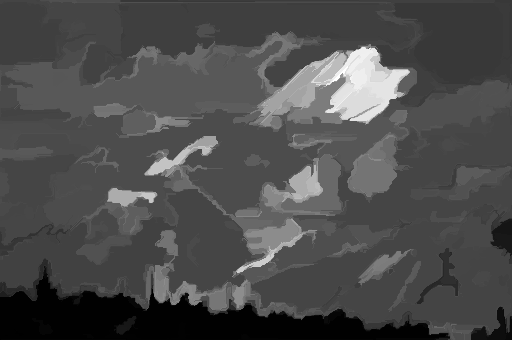} & \includegraphics[width=\x\textwidth]{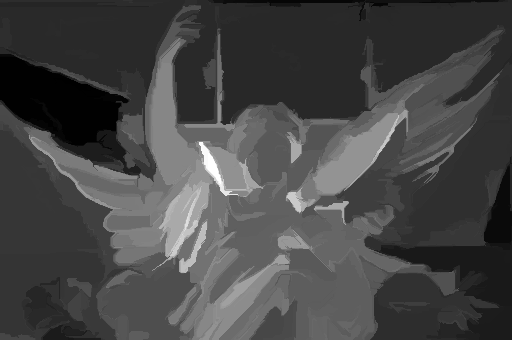} & \includegraphics[width=\x\textwidth]{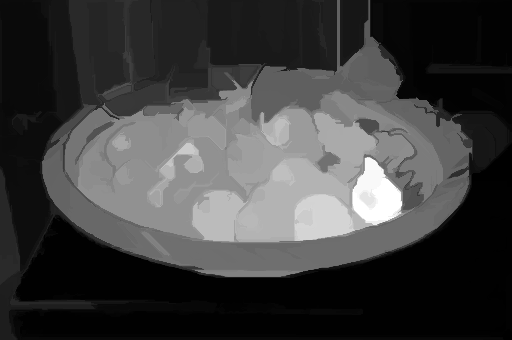}& \includegraphics[width=\x\textwidth]{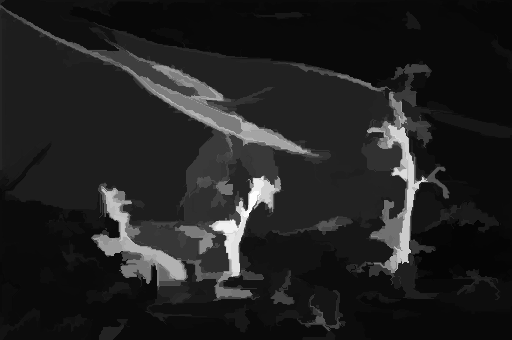}& \includegraphics[width=\x\textwidth]{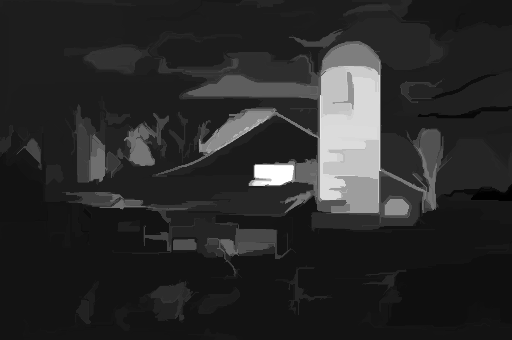}& \includegraphics[width=\x\textwidth]{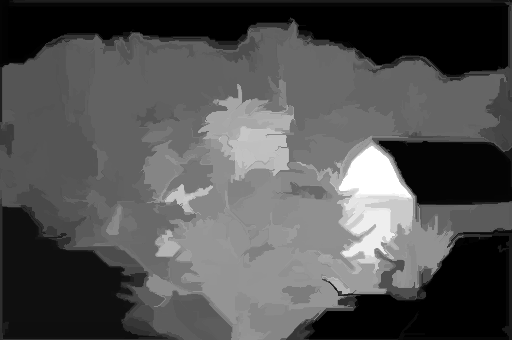}\\

\textbf{\footnotesize{Ghiasi \etal \cite{ghiasi2017exploring}:}} & \includegraphics[width=\x\textwidth]{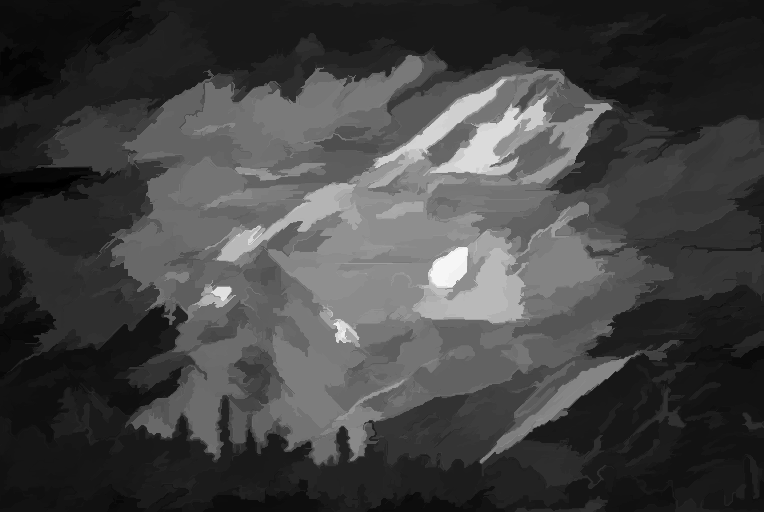} & \includegraphics[width=\x\textwidth]{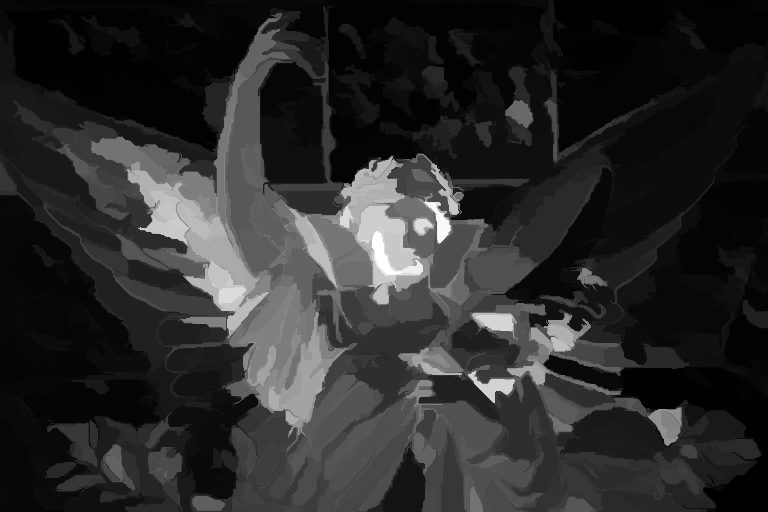} & \includegraphics[width=\x\textwidth]{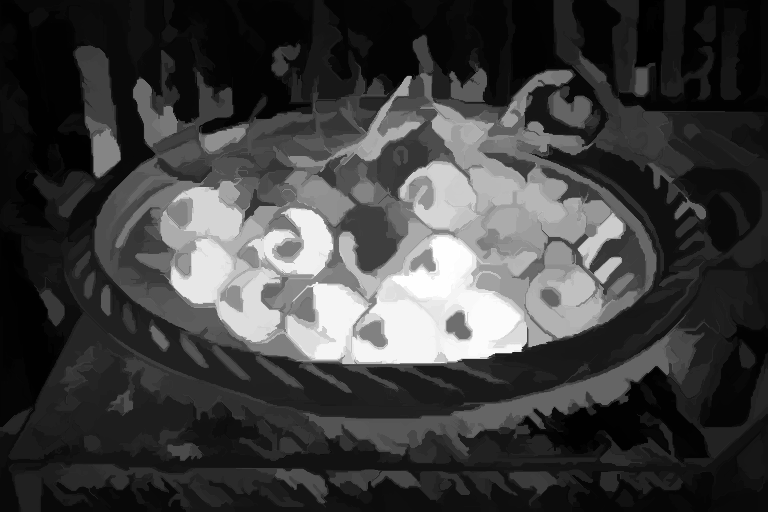}& \includegraphics[width=\x\textwidth]{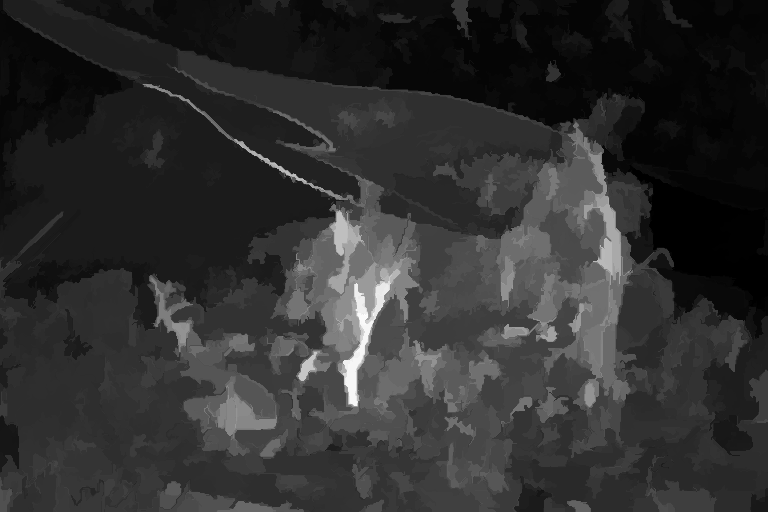}& \includegraphics[width=\x\textwidth]{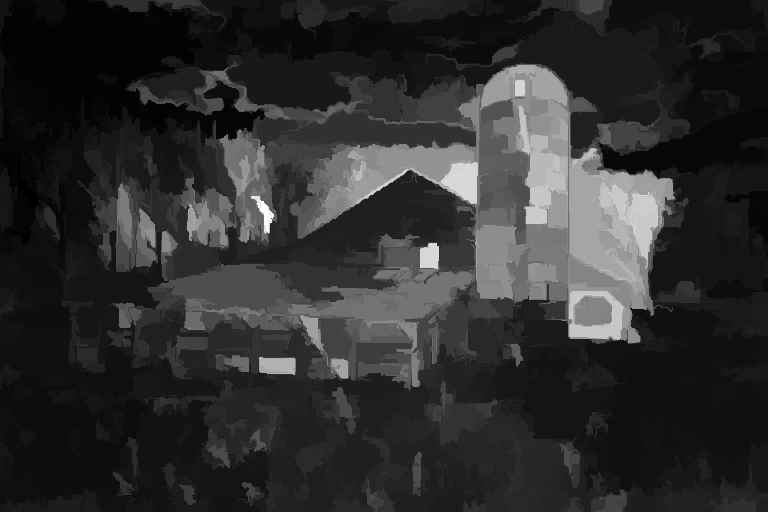}& \includegraphics[width=\x\textwidth]{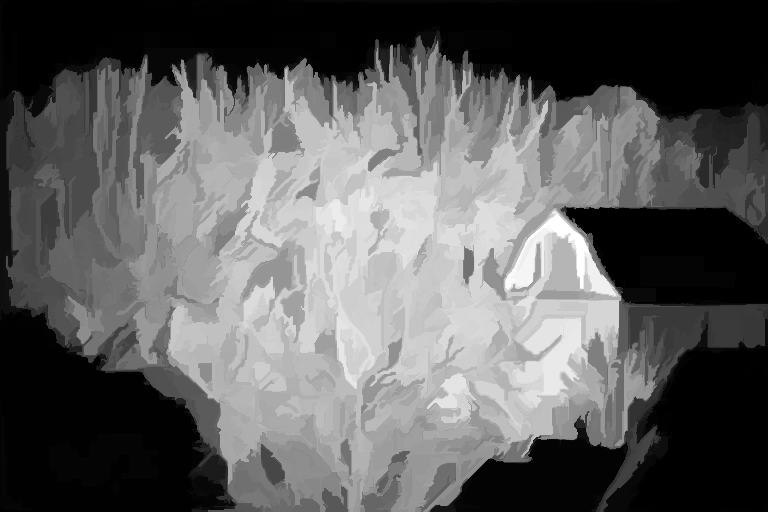}\\

%\textbf{\footnotesize{Ghiasi \etal \cite{ghiasi2017exploring}:}} & \includegraphics[width=\x\textwidth]{saliency_tvcg/bmvc/4_stylized_1_0.jpg} & \includegraphics[width=\x\textwidth]{saliency_tvcg/bmvc/1_stylized_2_0.jpg} & \includegraphics[width=\x\textwidth]{saliency_tvcg/bmvc/5_stylized_3_0.jpg}& \includegraphics[width=\x\textwidth]{saliency_tvcg/bmvc/8_stylized_5_0.jpg}& \includegraphics[width=\x\textwidth]{saliency_tvcg/bmvc/9_stylized_9_0.jpg}& \includegraphics[width=\x\textwidth]{saliency_tvcg/bmvc/10_stylized_10_0.jpg}\\

\multirow{1}{2.5cm}{\textbf{\footnotesize{Huang and Belongie \cite{huang2017arbitrary}:}}} & \includegraphics[width=\x\textwidth]{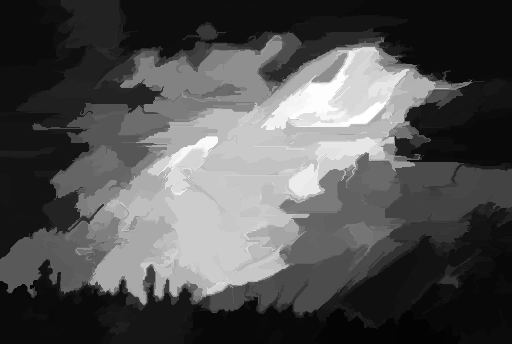} & \includegraphics[width=\x\textwidth]{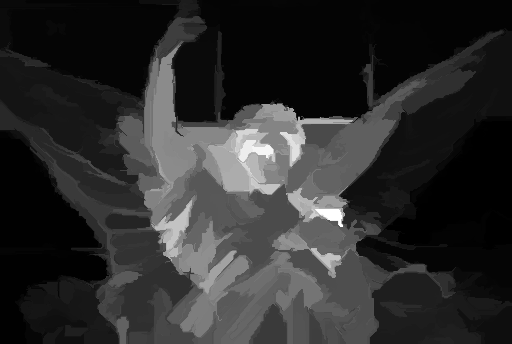} & \includegraphics[width=\x\textwidth]{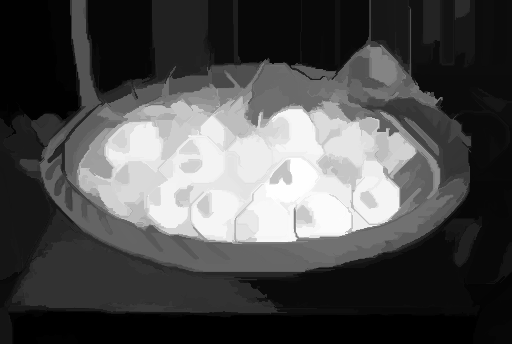}& \includegraphics[width=\x\textwidth]{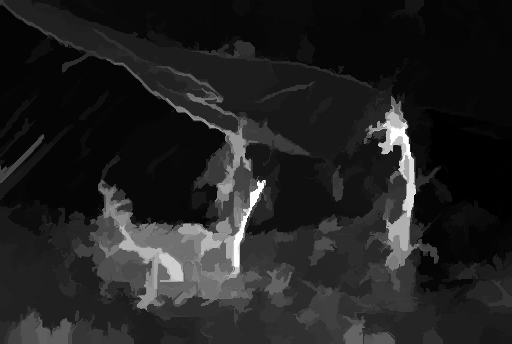}& \includegraphics[width=\x\textwidth]{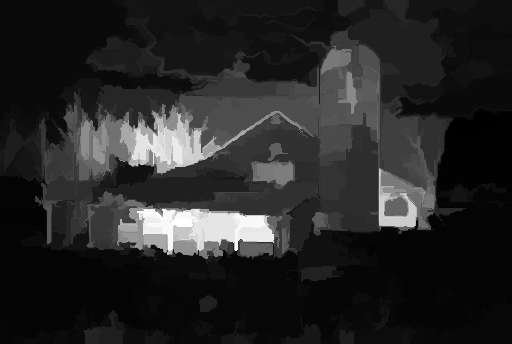}& \includegraphics[width=\x\textwidth]{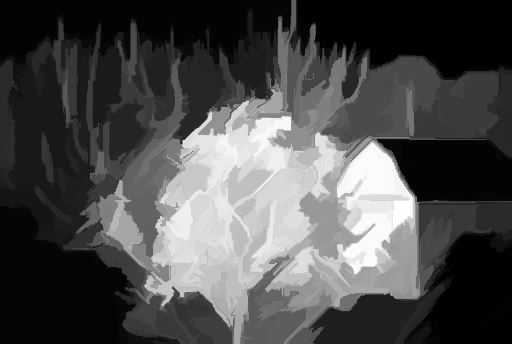}\\

%\multirow{1}{\y}{\textbf{\small{Huang and Belongie \cite{huang2017arbitrary}:}}} & \includegraphics[width=\x\textwidth]{saliency_tvcg/xun/4_stylized_1.jpg}& \includegraphics[width=\x\textwidth]{saliency_tvcg/xun/1_stylized_2.jpg}& \includegraphics[width=\x\textwidth]{saliency_tvcg/xun/5_stylized_3.jpg}&  \includegraphics[width=\x\textwidth]{saliency_tvcg/xun/8_stylized_5.jpg}& \includegraphics[width=\x\textwidth]{saliency_tvcg/xun/9_stylized_9.jpg}& \includegraphics[width=\x\textwidth]{saliency_tvcg/xun/10_stylized_10.jpg}\\

\textbf{\footnotesize{Li \etal \cite{li2017universal}:}} & \includegraphics[width=\x\textwidth]{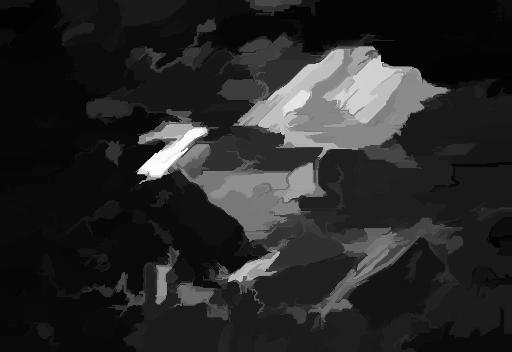} & \includegraphics[width=\x\textwidth]{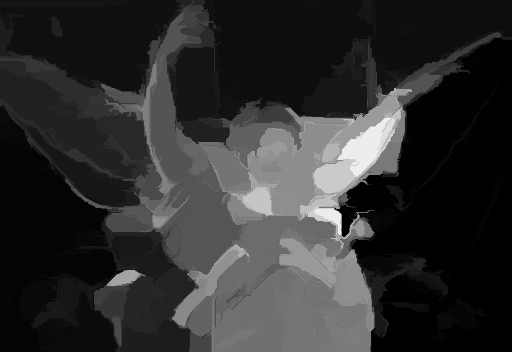} & \includegraphics[width=\x\textwidth]{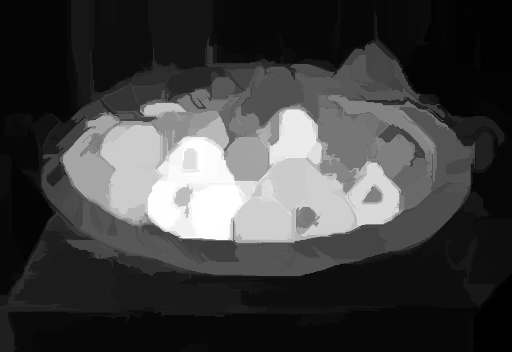}& \includegraphics[width=\x\textwidth]{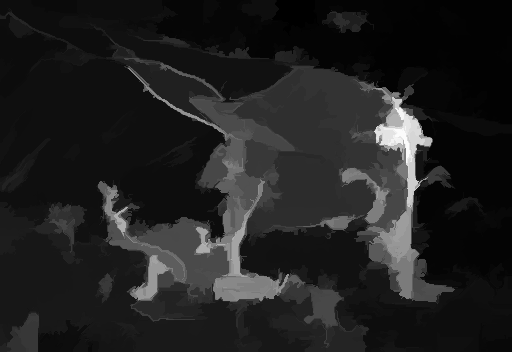}& \includegraphics[width=\x\textwidth]{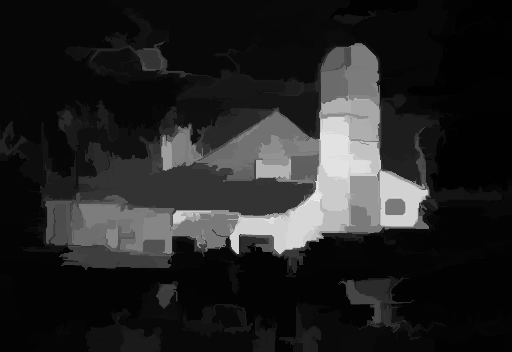}& \includegraphics[width=\x\textwidth]{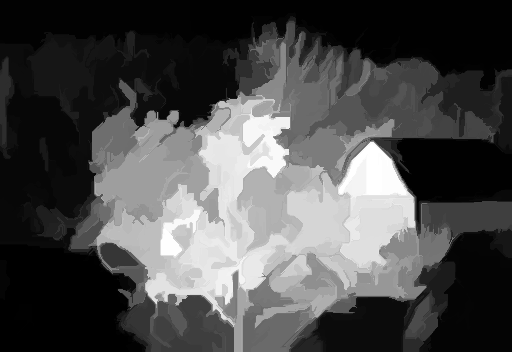}\\

\end{tabular}
}
\caption{Saliency detection results of \textbf{ASPM-MOB-NST}, corresponding to Figure~\ref{fig:qualitativeresult3}. The results are produced by using the discriminative regional feature integration approach proposed by Wang \etal \cite{wang2017salient}.}
%\caption{Some example results of \textbf{ASPM-MOB-NST} for qualitative evaluation. The content images are from the benchmark dataset proposed by Mould and Rosin \cite{mould2016benchmark,mould2017developing}. The style images are in the public domain. Detailed information of our style images can be found in Table~\ref{table:styleimage}.}
\label{fig:saliency3}
\end{figure*}

\subsection{Experimental Setup}

\textbf{Evaluation datasets.} Totally, there are ten style images and twenty content images used in our experiment.

For style images, we select artworks of diversified styles, as shown in Figure~\ref{fig:styleimage}. For example, there are impressionism, cubism, abstract, contemporary, futurism, surrealist, and expressionism art. Regarding the mediums, some of these artworks are painted on canvas, while others are painted on cardboard or wool, cotton, polyester, \etc. In addition, we also try to cover a range of image characteristics (such as details, contrast, complexity and color distributions), inspired by the works in \cite{mould2016benchmark,mould2017developing,rosin2017benchmarking}. More detailed information of our style images are given in Table~\ref{table:styleimage}.

For content images, there are already carefully selected and well-described benchmark datasets for evaluating stylisation by Mould and Rosin \cite{mould2016benchmark,mould2017developing,rosin2017benchmarking}. Their proposed NPR benchmark called \emph{NPRgeneral} consists of the images that cover a wide range of characteristics (\eg, contrast, texture, edges and meaningful structures) and satisfy lots of criteria. Therefore, we directly use the selected twenty images in their proposed \emph{NPRgeneral} benchmark as our content images.

For the algorithms based on offline model optimisation, MS-COCO dataset \cite{lin2014microsoft} is used to perform the training. All the content images are not used in training.

%we also try to select a wide variety of photos, which include animal photography, still life photography, landscape photography and portrait photography.

%In the area of NPR, Mould and Rosin
%
%detail, contrast, complexity, color distribution etc
%
%
%dataset\cite{mould2016benchmark,mould2017developing,rosin2017benchmarking}

\textbf{Principles.} To maximise the fairness of the comparisons, we also obey the following principles during our experiment:

%\begin{itemize}
 % \item
\textbf{1)} In order to cover every detail in each algorithm, we try to use the provided implementation from their published literatures. To maximise the fairness of comparison especially for speed comparison, for \cite{gatys2015neural}, we use a popular torch-based open source code \cite{code_Johnson_slow}, which is also admitted by the authors. In our experiment, except for \cite{dumoulin2016learned,berger2016incorporating} which are based on TensorFlow, all the other codes are implemented based on Torch 7.

%For \cite{gatys2015neural}, we use a popular torch-based open source code \cite{code_Johnson_slow} which is also admitted by the authors. Except for \cite{dumoulin2016learned,berger2016incorporating} which are based on TensorFlow, all the other codes are based on Torch 7, which maximises the fairness especially for speed comparison.
  %\item

\textbf{2)} Since the visual effect is influenced by the content and style weight, it is difficult to compare results with different degrees of stylisation. Simply giving the same content and style weight is not an optimal solution due to the different ways to calculate losses in each algorithm (\eg, different choices of content and style layers, different loss functions). Therefore, in our experiment, we try our best to balance the content and style weight among different algorithms.
 % \item

\textbf{3)} We try to use the default parameters (\eg, choice of layers, learning rate, \etc) suggested by the authors except for the aforementioned content and style weight. Although the results for some algorithms may be further improved by more careful hyperparameter tuning, we select the authors' default parameters since we hold the point that the \emph{sensitivity for hyperparameters} is also an important implicit criterion for comparison. For example, we cannot say an algorithm is effective if it needs heavy work to tune its parameters for each style.
%\end{itemize}

There are also some other implementation details to be noted. For \cite{Johnson2016perceptual} and \cite{ulyanov2016texture}, we use the instance normalisation strategy proposed in \cite{ulyanov2017improved}, which is not covered in the published papers. Also, we do not consider the diversity loss term (proposed in \cite{ulyanov2017improved,li2017diverse}) for all algorithms, \ie, one pair of content and style images corresponds to one stylised result in our experiment. For Chen and Schmidt's algorithm \cite{chen2016fast}, we use the feed-forward reconstruction to reconstruct the stylised results.

% of stylisation algorithms, we do not consider the diversity loss term for all algorithms, which is proposed in \cite{ulyanov2017improved} and \cite{li2017diverse}.

\subsection{Qualitative Evaluation}

%\textbf{Stylisation perceptual studies.} The ultimate test of Neural Style Transfer is how appealing the stylised results are to a human observer. Therefore, for qualitative evaluation of Neural Style Transfer, we propose a stylisation perceptual study which is to make human observers to judge and rank the results produced by different algorithms. About 40 observers with different occupations and ages participated in our experiment. A series of groups of images were shown to participants during our experiment. Each group consisted of 8 images of size $512 \times 512$ pixels, including one content image, one style image and six stylised images produced by different algorithms (five novel Generative Neural Methods and one Descriptive Neural Method). Participants were asked to rank six stylised images according to their own appreciation. For each group, participants were given unlimited time to appreciate and respond.
%
%\textbf{Evaluation metric of stylisation perceptual studies.} The average stylisation rank score is used as our evaluation metric here. For each group, if one of the stylised results is ranked first by an observer, it will get 6 as a stylisation rank score as there are totally six images to rank. Similarly, it gets 5 if it ranks second, 4 if third, \etc The final stylisation rank score of the stylised result produced by an algorithm is the average score over all observers.
Example stylised results are shown in Figure~\ref{fig:qualitativeresult1}, Figure~\ref{fig:qualitativeresult2} and Figure~\ref{fig:qualitativeresult3}. More results can be found in the supplementary material\footnote{\url{https://www.dropbox.com/s/5xd8iizoigvjcxz/SupplementaryMaterial_neuralStyleReview.pdf?dl=0}}.

%It is difficult to define the aesthetic criterion for an artwork. Therefore, for the same stylised result, different people may have different or even opposite views. Here, we choose to present stylised results of different algorithms and leave the judgement to readers.
%
%Since it is complex to define the notion of style \cite{rosin2012image,xie2007feature} and therefore very subjective to define what criteria are important to make a successful style transfer algorithm \cite{ashikhmin2003fast},

%In Figure~\ref{fig:qualitativeresult}, we build several blocks to separate results of different categories of NST algorithms.

%\begin{enumerate}[1)]
%  \item

\textbf{1) Results of IOB-NST.} Following the content and style images, Figure~\ref{fig:qualitativeresult1} contains the results of Gatys et al.'s IOB-NST algorithm based on online image optimisation \cite{gatys2016image}. The style transfer process is computationally expensive, but in contrast, the results are appealing in visual quality. Therefore, the algorithm of Gatys \etal is usually regarded as the gold-standard method in the community of NST.

%  \item

\textbf{2) Results of PSPM-MOB-NST.} Figure~\ref{fig:qualitativeresult1} shows the results of \emph{Per-Style-Per-Model} MOB-NST algorithms (Section \ref{sect:fastMethod}). Each model only fits one style. It can be noticed that the stylised results of Ulyanov \etal \cite{ulyanov2016texture} and Johnson \etal \cite{Johnson2016perceptual} are somewhat similar. This is not surprising since they share a similar idea and only differ in their detailed network architectures. For the results of Li and Wand \cite{li2016precomputed}, the results are sightly less impressive. Since \cite{li2016precomputed} is based on Generative Adversarial Network (GAN), to some extent, the training process is not that stable. But we believe that GAN-based style transfer is a very promising direction, and there are already some other GAN-based works \cite{azadi2018multi,zhang2017style,zhu2017unpaired} (Section~\ref{sect:improvementextensions}) in the field of NST.

%We believe that further improving the GAN-based algorithm of Li and Wand \cite{li2016precomputed} is a very promising direction in the future.GAN to image style transfer \cite{zhu2017unpaired}

%  \item

\textbf{3) Results of MSPM-MOB-NST.} Figure~\ref{fig:qualitativeresult2} demonstrates the results of \emph{Multiple-Style-Per-Model} MOB-NST algorithms. Multiple styles are incorporated into a single model. The idea of both Dumoulin et al.'s algorithm \cite{dumoulin2016learned} and Chen et al.'s algorithm \cite{chen2017stylebank} is to tie a small number of parameters to each style. Also, both of them build their algorithm upon the architecture of \cite{Johnson2016perceptual}. Therefore, it is not surprising that their results are visually similar. Although the results of \cite{dumoulin2016learned,chen2017stylebank} are appealing, their model size will become larger with the increase of the number of learned styles. In contrast, Zhang and Dana's algorithm \cite{zhang2017multi} and Li et al.'s algorithm \cite{li2017diverse} use a single network with the same trainable network weights for multiple styles. The model size issue is tackled, but there seem to be some interferences among different styles, which slightly influences the stylisation quality.
%(Group \uppercase\expandafter{\romannumeral2} and \uppercase\expandafter{\romannumeral7})
%For the algorithm of Dumoulin \etal \cite{dumoulin2016learned}, it seems that for some styles (Group \uppercase\expandafter{\romannumeral3} and \uppercase\expandafter{\romannumeral6}), their algorithm is capable of retaining more style elements than others. Chen \etal's algorithm \cite{chen2017stylebank}
%
%For Chen \etal's algorithm \cite{chen2017stylebank}, the results are very similar to the gold-standard Gatys \etal's algorithm.

%Next, we observe the results of Zhang and Dana's algorithm \cite{zhang2017multi} and Li \etal's algorithm \cite{li2017diverse} jointly. One observation is that their results are somewhat visually similar. This is not surprising since the ideas of both \cite{zhang2017multi} and \cite{li2017diverse} are to share the same network architecture and the same trainable network weights for multiple styles and there may be some interferences between different styles. In contrast, the ideas of \cite{dumoulin2016learned} and \cite{chen2017stylebank} are to use the same architecture but different trainable network weights of some network components for multiple styles. In this way, different styles may be better separated during training, though there is a little increased extra storage overhead for different trainable network weights with the increase in style numbers.

%  \item

\textbf{4) Results of ASPM-MOB-NST.} Figure~\ref{fig:qualitativeresult3} presents the last category of MOB-NST algorithms, namely \emph{Arbitrary-Style-Per-Model} MOB-NST algorithms. Their idea is one-model-for-all. Globally, the results of ASPM are slightly less impressive than other types of algorithms. This is acceptable in that a three-way trade-off between speed, flexibility and quality is common in research. Chen and Schmidt's patch-based algorithm \cite{chen2016fast} seems to not combine enough style elements into the content image. Their algorithm is based on similar patch swap. When lots of content patches are swapped with style patches that do not contain enough style elements, the target style will not be reflected well. Ghiasi et al.'s algorithm \cite{ghiasi2017exploring} is data-driven and their stylisation quality is very dependent on the varieties of training styles. For the algorithm of Huang and Belongie \cite{huang2017arbitrary}, they propose to match global summary feature statistics and successfully improve the visual quality compared with \cite{chen2016fast}. However, their algorithm seems not good at handling complex style patterns, and their stylisation quality is still related to the varieties of training styles. The algorithm of Li \etal \cite{li2017universal} replaces the training process with a series of transformations. But \cite{li2017universal} is not effective at producing sharp details and fine strokes.

\textbf{Saliency Comparison.} NST is an art creation process. As indicated in \cite{rosin2012image,xie2007feature,ashikhmin2003fast}, the definition of \emph{style} is subjective and also very complex, which involves personal preferences, texture compositions as well as the used tools and medium. As a result, it is difficult to define the aesthetic criterion for a stylised artwork. For the same stylised result, different people may have different or even opposite views. Nevertheless, our goal is to compare the results of different NST techniques (shown in Figure~\ref{fig:qualitativeresult1}, Figure~\ref{fig:qualitativeresult2} and Figure~\ref{fig:qualitativeresult3}) as objectively as possible. Here, we consider comparing saliency maps, as proposed in \cite{liu2017depthaware}. The corresponding results are shown in Figure~\ref{fig:saliency1}, Figure~\ref{fig:saliency2} and Figure~\ref{fig:saliency3}. Saliency maps can demonstrate visually dominant locations in images. Intuitively, a successful style transfer could weaken or enhance the saliency maps in content images, but should not change the integrity and coherence. From Figure~\ref{fig:saliency1} (saliency detection results of IOB-NST and PSPM-MOB-NST), it can be noticed that the stylised results of \cite{gatys2016image,ulyanov2016texture,Johnson2016perceptual} preserve the structures of content images well; however, for \cite{li2016precomputed}, it might be harder for an observer to recognise the objects after stylisation. Using similar analytical method, from Figure~\ref{fig:saliency2} (saliency detection results of MSPM-MOB-NST), \cite{dumoulin2016learned} and \cite{chen2017stylebank} preserve similar saliency of the original content images since they both tie a small number of parameters to each style. \cite{zhang2017multi} and \cite{li2017diverse} are also similar regarding the ability to retain the integrity of the original saliency maps, because they both use a single network for all styles. As shown in Figure~\ref{fig:saliency3}, for the saliency detection results of ASPM-MOB-NST, \cite{ghiasi2017exploring} and \cite{huang2017arbitrary} perform better than \cite{chen2016fast} and \cite{li2017universal}; however, both \cite{ghiasi2017exploring} and \cite{huang2017arbitrary} are data-driven methods and their quality depends on the diversity of training styles. In general, it seems that the results of MSPM-MOB-NST preserve better saliency coherence than ASPM-MOB-NST, but a little inferior to IOB-NST and PSPM-MOB-NST.

\begin{table*}
\renewcommand\arraystretch{1.3}
\caption{Average speed comparison of NST algorithms for images of size $256 \times 256$ pixels, $512 \times 512$ pixels and $1024 \times 1024$ pixels (on an NVIDIA Quadro M6000)}
%\begin{minipage}{\columnwidth}
\begin{center}
\begin{tabular}{  l | c | c | c | c  }
    %\toprule[1pt]
    %\hline
    \textbf{Methods} & \multicolumn{3}{c|}{\textbf{Time(s)}} & \textbf{Styles/Model}  \\  & \textbf{256 $\times$ 256} & \textbf{512 $\times$ 512} & \textbf{1024 $\times$ 1024}& \\
    \hline
    Gatys \etal \cite{gatys2015neural}  & 14.32 & 51.19 & 200.3 & $\infty$  \\
    Johnson \etal \cite{Johnson2016perceptual}  & 0.014 & 0.045 & 0.166 & 1  \\
    Ulyanov \etal \cite{ulyanov2016texture}  & 0.022 & 0.047 & 0.145 & 1  \\
    Li and Wand \cite{li2016precomputed} & 0.015 & 0.055 & 0.229 & 1  \\
    Zhang and Dana \cite{zhang2017multi} & 0.019 (\textbf{0.039}) & 0.059 (\textbf{0.133}) & 0.230 (\textbf{0.533}) & $k (k \in Z^{+})$ \\
    Li \etal \cite{li2017diverse} & 0.017 & 0.064 & 0.254 & $k (k \in Z^{+})$ \\
    Chen and Schmidt \cite{chen2016fast} & 0.123 (\textbf{0.130}) & 1.495 (\textbf{1.520}) & $-$ & $\infty$ \\
    Huang and Belongie \cite{huang2017arbitrary} & 0.026 (\textbf{0.037}) & 0.095 (\textbf{0.137}) & 0.382 (\textbf{0.552}) & $\infty$  \\
    Li \etal \cite{li2017universal} & 0.620 & 1.139 & 2.947 & $\infty$  \\

    %\midrule[1pt]
     \end{tabular}
\end{center}
%\bigskip\centering
\footnotesize
%\begin{flushleft}
\smallskip
\emph{Note:} The fifth column shows the number of styles that a single model can produce. Time both excludes (out of parenthesis) and includes (in parenthesis) the style encoding process is shown, since \cite{zhang2017multi}, \cite{chen2016fast} and \cite{huang2017arbitrary} support storing encoded style statistics in advance to further speed up the stylisation process for the same style but different content images. Time of \cite{chen2016fast} for producing $1024 \times 1024$ images is not shown due to the memory limitation. The speed of \cite{dumoulin2016learned,ghiasi2017exploring} are similar to \cite{Johnson2016perceptual} since they share similar architecture. We do not redundantly list them in this table. \hfill
%\end{flushleft}
\label{table:speed}
%\end{minipage}
\end{table*}

\begin{table*}
\caption{A summary of the advantages and disadvantages of the mentioned algorithms in our experiment.}
\linespread{1.1}
%\begin{center}
%\tabcolsep0.01in
\begin{small}
%\newpage
%\begin{tabular}{p{3cm} p{2cm} p{3cm} p{3cm} p{3cm} p{3cm}}
\begin{center}
\begin{tabular}{l|l|c|c|c|p{7.5cm}l}
%\toprule[1.2pt]
\textbf{Types} & \textbf{Methods} & \multicolumn{4}{c}{\textbf{Pros \& Cons}} \\ \cline{3-7}  & & \footnotesize{\textbf{E}} & \footnotesize{\textbf{AS}} & \footnotesize{\textbf{LF}} & \multicolumn{1}{c}{\footnotesize{\textbf{VQ}}}  \\
%\midrule[1pt]
\hline

IOB-NST
& Gatys \etal \cite{gatys2016image} & $\times$ & $\surd$ & $\surd$ & \multirow{1}{9cm}{\scriptsize Good and usually regarded as a gold standard.}\\

%\midrule[0.7pt]
\hline

\multirow{3}{*}{\shortstack{\textbf{PSPM}-\\MOB-NST}}
& Ulyanov \etal \cite{Johnson2016perceptual} & $\surd$ & $\times$ & $\times$ & \multirow{3}{8cm}{\scriptsize The results of \cite{Johnson2016perceptual,ulyanov2017improved} are close to \cite{gatys2016image}. \cite{li2016precomputed} is generally less appealing than \cite{Johnson2016perceptual,ulyanov2017improved}.} &\\
&Johnson \etal \cite{ulyanov2017improved} & $\surd$& $\times$ & $\times$ &\\
&Li and Wand \cite{li2016precomputed} &$\surd$ & $\times$ & $\times$ & \\

%\midrule[0.7pt]
\hline

\multirow{4}{*}{\shortstack{\textbf{MSPM}-\\MOB-NST}}
& Dumoulin \etal \cite{dumoulin2016learned} & $\surd$ & $\times$ &$\times$ & \multirow{4}{8cm}{\scriptsize The results of \cite{dumoulin2016learned} and \cite{chen2017stylebank} are close to \cite{gatys2016image}, but the model size generally becomes larger with the increase of the number of learned styles. \cite{li2017diverse,zhang2017multi} have a fixed model size but there seem to be some interferences among different styles. }&\\
& Chen \etal \cite{chen2017stylebank} & $\surd$ &$\times$ & $\times$ & &\\ %\cline{2-4}
& Li \etal \cite{li2017diverse} & $\surd$ &$\times$ & $\times$ & &\\
& Zhang and Dana \cite{zhang2017multi} & $\surd$ &$\times$ & $\times$ & &\\

%\midrule[0.7pt]
\hline

\multirow{4}{*}{\shortstack{\textbf{ASPM}-\\MOB-NST}}
& Chen and Schmidt \cite{chen2016fast} & $\surd$ & $\surd$& $\times$ & \multirow{4}{8cm}{\scriptsize In general, the results of ASPM are less impressive than other types of NST algorithms. \cite{chen2016fast} does not combine enough style elements. \cite{ghiasi2017exploring,huang2017arbitrary} are generally not effective at producing complex style patterns. \cite{li2017universal} is not good at producing sharp details and fine strokes.}\\
& Ghiasi \etal \cite{ghiasi2017exploring} & $\surd$ & $\surd$& $\times$ & \\
& Huang and Belongie \cite{huang2017arbitrary} & $\surd$ &$\surd$ & $\times$ & \\
& Li \etal \cite{li2017universal} & $\surd$ &$\surd$ & $\surd$ & \\
%\bottomrule[1.2pt]
\hline
\end{tabular}
\end{center}
\end{small}
%\end{center}
\footnotesize
%\begin{flushleft}

\smallskip
\smallskip
\smallskip
\emph{Note:} \textbf{E}, \textbf{AS}, \textbf{LF}, and \textbf{VQ} represent \textbf{\emph{Efficient}}, \textbf{\emph{Arbitrary Style}}, \textbf{\emph{Learning-Free}}, and \textbf{\emph{Visual Quality}}, respectively. IOB-NST denotes the category \emph{Image-Optimisation-Based Neural Style Transfer} and MOB-NST represents \emph{Model-Optimisation-Based Neural Style Transfer}.
\label{table:summary}
\end{table*}

\subsection{Quantitative Evaluation}
%\stdcomment{
%1. Stylisation Turing test (quality) 2. Time (speed) 3. Memory-consuming (computational cost)
%}
%
%\stdcomment{\textbf{LASSO Regression? -- Sharon Lin- Modeling How People Extract colour Themes from Images- Standford 2013}}

Regarding the quantitative evaluation, we mainly focus on five evaluation metrics, which are: generating time for a single content image of different sizes; training time for a single model; average loss for content images to measure how well the loss function is minimised; loss variation during training to measure how fast the model converges; style scalability to measure how large the learned style set can be.

\textbf{1) Stylisation speed.} The issue of efficiency is the focus of MOB-NST algorithms. In this subsection, we compare different algorithms quantitatively in terms of the stylisation speed. Table~\ref{table:speed} demonstrates the average time to stylise one image with three resolutions using different algorithms. In our experiment, the style images have the same size as the content images. The fifth column in Table~\ref{table:speed} represents the number of styles one model of each algorithm can produce. $k (k \in Z^{+})$ denotes that a single model can produce multiple styles, which corresponds to MSPM algorithms. $\infty$ means a single model works for any style, which corresponds to ASPM algorithms. The numbers reported in Table~\ref{table:speed} are obtained by averaging the generating time of 100 images. Note that we do not include the speed of \cite{dumoulin2016learned,ghiasi2017exploring} in Table~\ref{table:speed} as their algorithm is to scale and shift parameters based on the algorithm of Johnson \etal \cite{Johnson2016perceptual}. The time required to stylise one image using \cite{dumoulin2016learned,berger2016incorporating} is very close to \cite{Johnson2016perceptual} under the same setting. For Chen et al.'s algorithm in \cite{chen2017stylebank}, since their algorithm is protected by patent and they do not make public the detailed architecture design, here we just attach the speed information provided by the authors for reference: On a Pascal Titan X GPU, $256 \times 256$: $0.007$s; $512 \times 512$: $0.024$s; $1024 \times 1024$: $0.089$s. For Chen and Schmidt's algorithm \cite{chen2016fast}, the time for processing a $1024 \times 1024$ image is not reported due to the limit of video memory. Swapping patches for two $1024 \times 1024$ images needs more than 24 GB video memory and thus, the stylisation process is not practical. We can observe that except for \cite{chen2016fast,li2017universal}, all the other MOB-NST algorithms are capable of stylising even high-resolution content images in real-time. ASPM algorithms are generally slower than PSPM and MSPM, which demonstrates the aforementioned three-way trade-off again.

\textbf{2) Training time.} Another concern is the training time for one single model. The training time of different algorithms is hard to compare as sometimes the model trained with just a few iterations is capable of producing enough visually appealing results. So we just outline our training time of different algorithms (under the same setting) as a reference for follow-up studies. On a NVIDIA Quadro M6000, the training time for a single model is about $3.5$ hours for the algorithm of Johnson \etal \cite{Johnson2016perceptual}, $3$ hours for the algorithm of Ulyanov \etal \cite{ulyanov2016texture}, $2$ hours for the algorithm of Li and Wand \cite{li2016precomputed}, $4$ hours for Zhang and Dana \cite{zhang2017multi}, and $8$ hours for Li \etal \cite{li2017diverse}. Chen and Schmidt's algorithm \cite{chen2016fast} and Huang and Belongie's algorithm \cite{huang2017arbitrary} take much longer (\eg, a couple of days), which is acceptable since a pre-trained model can work for any style. The training time of \cite{ghiasi2017exploring} depends on how large the training style set is. For MSPM algorithms, the training time can be further reduced through incremental learning over a pre-trained model. For example, the algorithm of Chen \etal only needs $8$ minutes to incrementally learn a new style, as reported in \cite{chen2017stylebank}.

%6.5 hours for Chen \etal \cite{chen2017stylebank},
%%%%%%%%%%%%%%%%%%

%%%%%
\begin{figure*}
  \centering
  \subfigure[Total Loss Curve]{
    %\label{fig:subfig:a} %% label for first subfigure
    \includegraphics[width=0.32\textwidth]{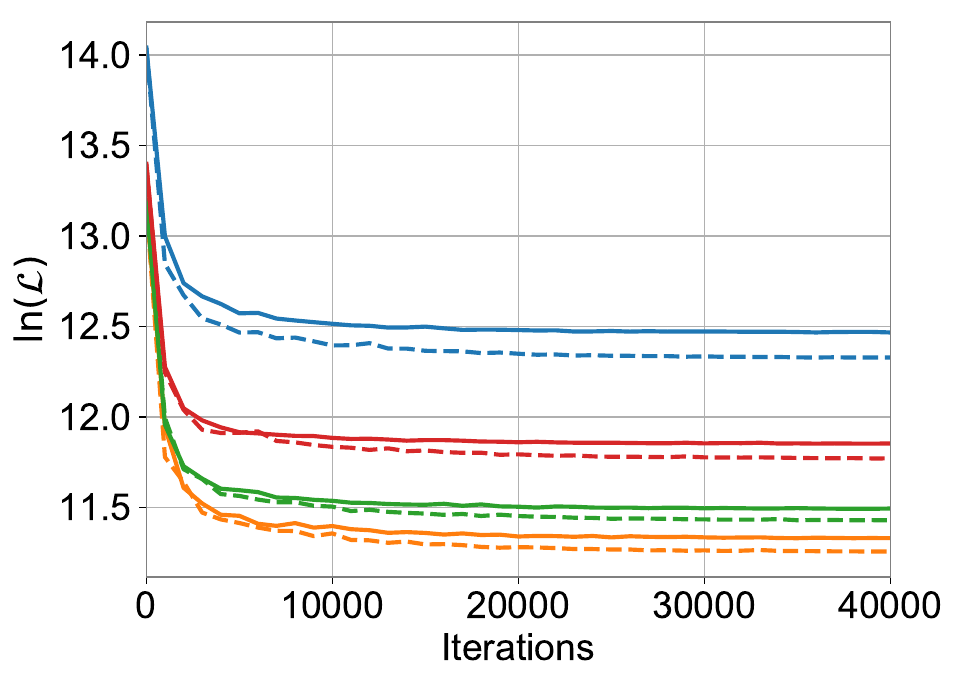}}
  %\hspace{0.01in}
  \subfigure[Style Loss Curve]{
    %\label{fig:subfig:a} %% label for first subfigure
    \includegraphics[width=0.32\textwidth]{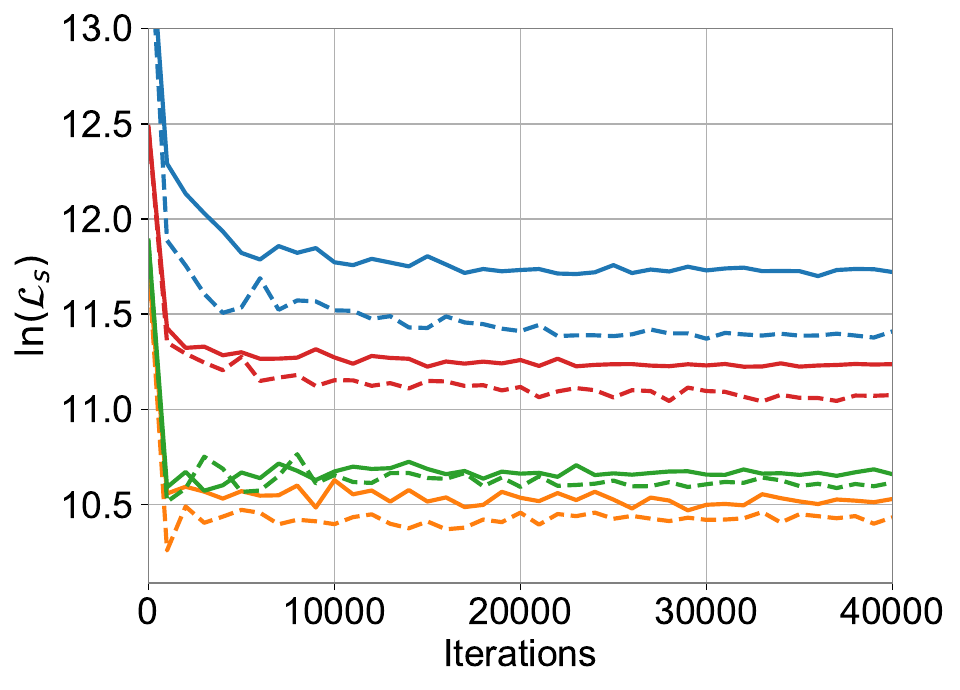}}
  %\hspace{0.01in}
  \subfigure[Content Loss Curve]{
    %\label{fig:subfig:b} %% label for second subfigure
    \includegraphics[width=0.32\textwidth]{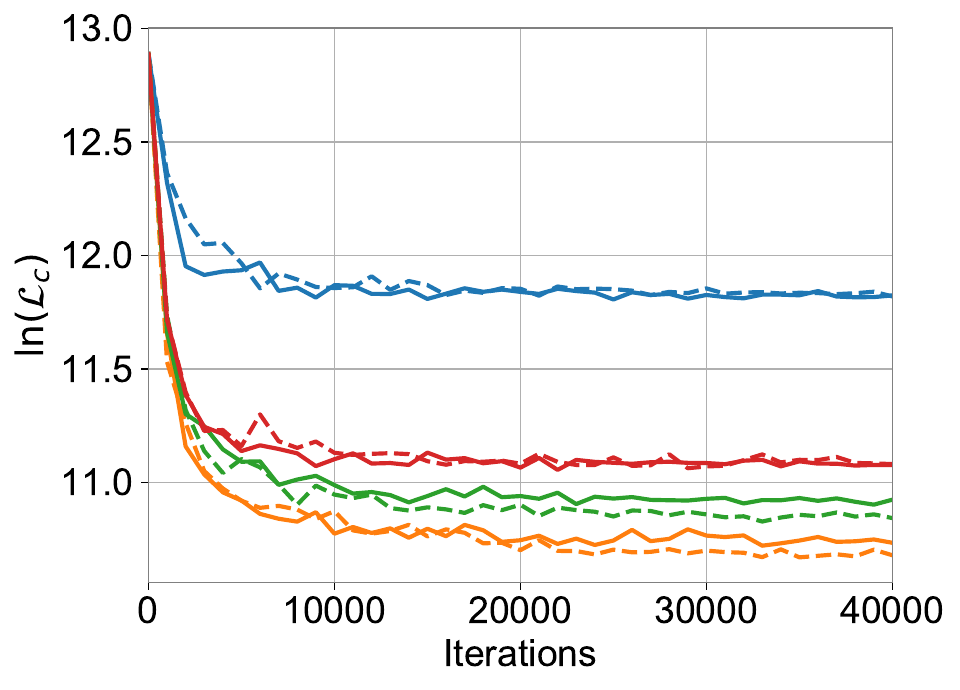}}
  \caption{Training curves of total loss, style loss and content loss of different algorithms. Solid curves represent the loss variation of the algorithm of Ulyanov \etal \cite{ulyanov2016texture}, while the dashed curves represent the algorithm of Johnson \etal \cite{Johnson2016perceptual}. Different colours correspond to different randomly selected styles from our style set.}
  \label{fig:losscurve} %% label for entire figure The content and style loss have the same weight when training.
\end{figure*}
%%%%%

%%%%%
\begin{figure*}
  \centering
  \subfigure[Total Loss]{
    %\label{fig:subfig:a} %% label for first subfigure
    \includegraphics[width=0.32\textwidth]{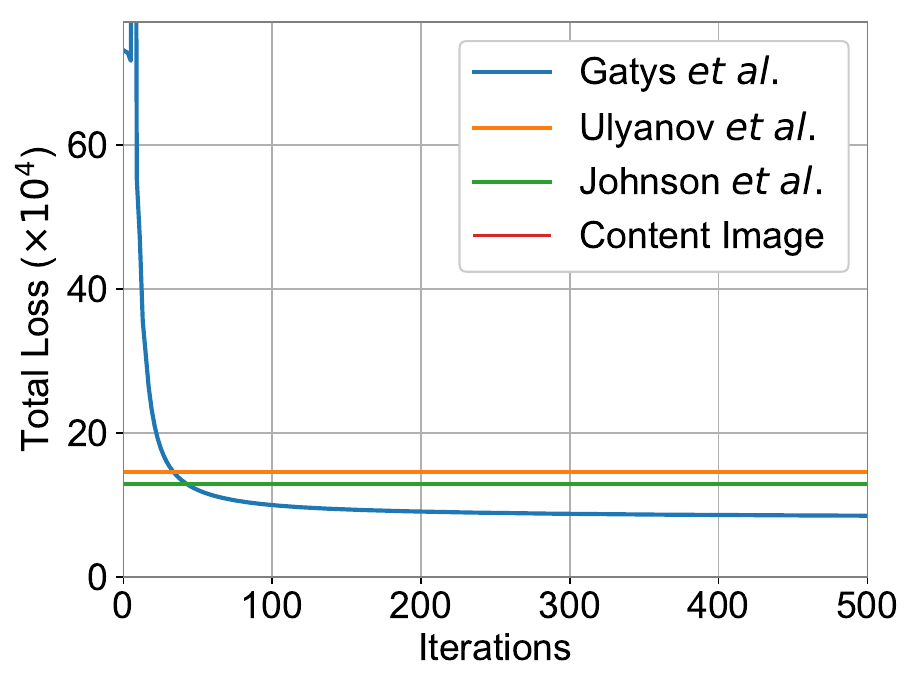}}
  %\hspace{0.01in}
  \subfigure[Style Loss]{
    %\label{fig:subfig:a} %% label for first subfigure
    \includegraphics[width=0.32\textwidth]{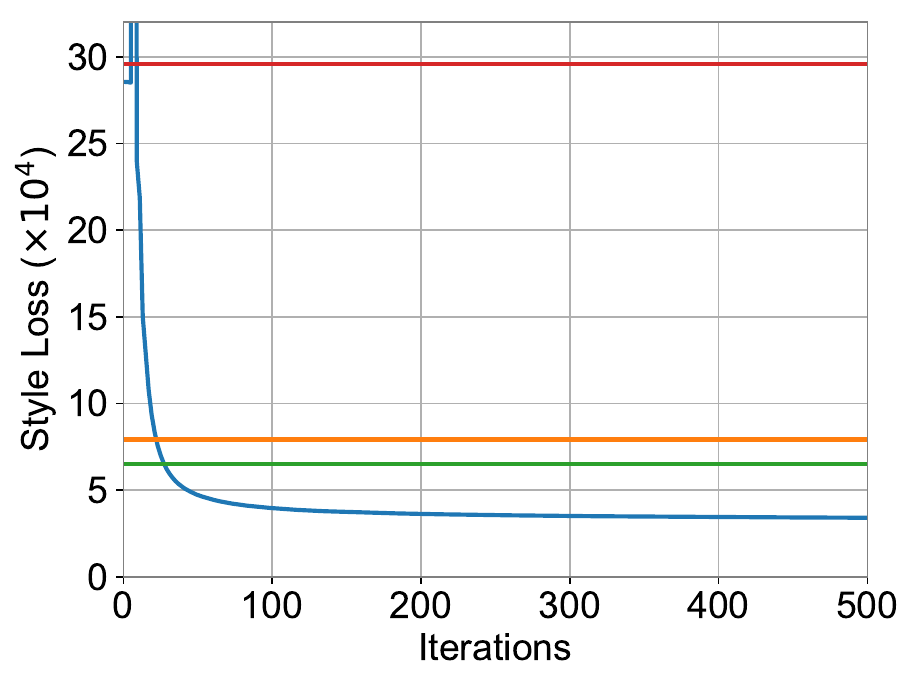}}
  %\hspace{0.01in}
  \subfigure[Content Loss]{
    %\label{fig:subfig:b} %% label for second subfigure
    \includegraphics[width=0.32\textwidth]{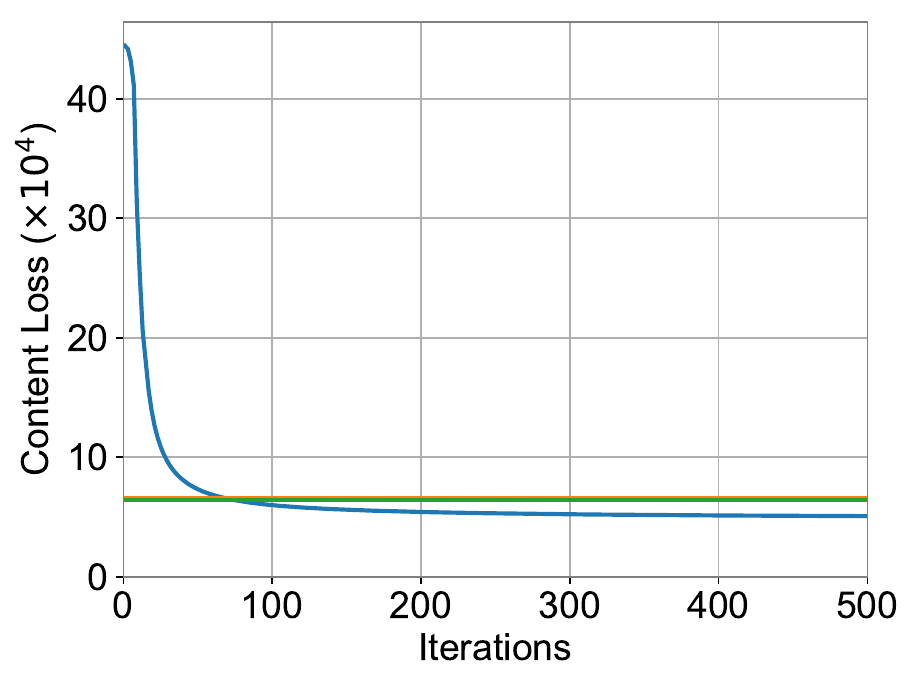}}
  \caption{Average total loss, style loss and content loss of different algorithms \cite{gatys2016image,Johnson2016perceptual,ulyanov2016texture}. The reported numbers are averaged over our set of style and content images.}
  \label{fig:finalloss} %% label for entire figure
\end{figure*}
%%%%%
%%%%%%%%%%%%%%%%%%
\textbf{3) Loss comparison.} One way to evaluate some MOB-NST algorithms which share the same loss function is to compare their loss variation during training, \ie, the training curve comparison. It helps researchers to justify the choice of architecture design by measuring how fast the model converges and how well the same loss function can be minimised. Here we compare training curves of two popular MOB-NST algorithms \cite{Johnson2016perceptual,ulyanov2016texture} in Figure~\ref{fig:losscurve}, since most of the follow-up works are based on their architecture designs. We remove the total variation term and keep the same objective for both two algorithms. Other settings (\eg, loss network, chosen layers) are also kept the same. For the style images, we randomly select four styles from our style set and represent them in different colours in Figure~\ref{fig:losscurve}. It can be observed that the two algorithms are similar in terms of the convergence speed. Also, both algorithms minimise the content loss well during training, and they mainly differ in the speed of learning the style objective. The algorithm in \cite{Johnson2016perceptual} minimises the style loss better.

Another related criterion is to compare the final loss values of different algorithms over a set of test images. This metric demonstrates how well the same loss function can be minimised by using different algorithms. For a fair comparison, the loss function and other settings are also required to be kept the same. We show the results of one IOB-NST algorithm \cite{gatys2016image} and two MOB-NST algorithms \cite{Johnson2016perceptual,ulyanov2016texture} in Figure~\ref{fig:finalloss}. The result is consistent with the aforementioned trade-off between speed and quality. Although MOB-NST algorithms are capable of stylising images in real-time, they are not good as IOB-NST algorithms in terms of minimising the same loss function.

%This measurement can also help compare different categories of algorithms that have the same loss function to minimise.

\textbf{4) Style scalability.} Scalability is a very important criterion for MSPM algorithms. However, it is very hard to measure since the maximum capabilities of a single model is highly related to the set of particular styles. If most styles have somewhat similar patterns, a single model can produce thousands of styles or even more, since these similar styles share somewhat similar distribution of style feature statistics. In contrast, if the style patterns vary a lot among different style images, the capability of a single model will be much smaller. But it is hard to measure how much these styles differ from each other in style patterns. Therefore, to provide the reader a reference, here we just summarise the authors' attempt for style scalability: the number is $32$ for \cite{dumoulin2016learned}, $1000$ for both \cite{chen2017stylebank} and \cite{li2017diverse}, and $100$ for \cite{zhang2017multi}.

A summary of the advantages and disadvantages of the mentioned algorithms in this experiment section can be found in Table~\ref{table:summary}.

%%%%%%%%%%%%%%%%%%%%%%%%%%%%
%%%%%%%%%%%%%%%%%%%%%%%%%%%%
%%%%%%%%%%%%%%%%%%%%%%%%%%%%

\section{Applications}
\label{sect:applications}

%\myScomment{
%\begin{itemize}
%  \item Becattini 2016- Imaging Novecento. A Mobile App for Automatic Recognition of Artworks and Transfer of Artistic Styles \cite{becattini2016imaging}
%  \item Bhautik J Joshi 2017- Bringing Impressionism to Life with Neural Style Transfer in Come Swim \cite{joshi2017bringing}
%\end{itemize}
%}
%\myLcomment{caffe2go}

Due to the visually plausible stylised results, the research of NST has led to many successful industrial applications and begun to deliver commercial benefits. In this section, we summarise these applications and present some potential usages.
%There are also some application papers aiming to investigate how to apply NST technique in different applications \cite{becattini2016imaging, joshi2017bringing}.
%showcase some applications

%\stdcomment{
%Discuss some applications of Neural Style Transfer. Does current algorithm meet the special needs for these applications?
%}
%
%\stdcomment{
%\begin{itemize}
%%  \item Social communication app (discuss android, ios, and webpage app)
%  \item HD Wallpapers for computer desktop, or for a wall painting?
%  \item Textile printing
%  \item Tool for designer and painter
%\end{itemize}
%}

\subsection{Social Communication}

One reason why NST catches eyes in both academia and industry is its popularity in some social networking sites, \eg, Facebook and Twitter. A recently emerged mobile application named \emph{Prisma} \cite{prisma} is one of the first industrial applications that provide the NST algorithm as a service. Due to its high stylisation quality, \emph{Prisma} achieved great success and is becoming popular around the world. Some other applications providing the same service appeared one after another and began to deliver commercial benefits, \eg, a web application \emph{Ostagram} \cite{ostagram} requires users to pay for a faster stylisation speed. Under the help of these industrial applications \cite{deepforger,deepart,neuralstyler}, people can create their own art paintings and share their artwork with others on Twitter and Facebook, which is a new form of social communication. There are also some related application papers: \cite{STDKD17} introduces an iOS app \emph{Pictory} which combines style transfer techniques with image filtering; \cite{PSKD17} further presents the technical implementation details of \emph{Pictory}; \cite{DSSTD17} demonstrates the design of another GPU-based mobile app \emph{ProsumerFX}.

The application of NST in social communication reinforces the connections between people and also has positive effects on both academia and industry. For academia, when people share their own masterpiece, their comments can help the researchers to further improve the algorithm. Moreover, the application of NST in social communication also drives the advances of other new techniques. For instance, inspired by the real-time requirements of NST for videos, Facebook AI Research (FAIR) first developed a new mobile-embedded deep learning system \emph{Caffe2Go} and then \emph{Caffe2} (now merged with PyTorch), which can run deep neural networks on mobile phones \cite{caffe2go}. For industry, the application brings commercial benefits and promotes the economic development.

%Germany friend's paper
%
%mobile design \cite{DSSTD17}
%
%demo: picture \cite{PSKD17}
%
%picture \cite{SID17}
%
%neural style trasnfer: a shift for .. \cite{SID17}

%There are also some other similar applications,

%\eg, \emph{Deep Forger} \cite{deepforger}, \emph{DeepArt} \cite{deepart} and \emph{NeuralStyler} \cite{neuralstyler}. Also

\subsection{User-assisted Creation Tools}

Another use of NST is to make it act as user-assisted creation tools. Although there are no popular applications that applied the NST technique in creation tools, we believe that it will be a promising potential usage in the future.

As a creation tool for painters and designers, NST can make it more convenient for a painter to create an artwork of a particular style, especially when creating computer-made artworks. Moreover, with NST algorithms, it is trivial to produce stylised fashion elements for fashion designers and stylised CAD drawings for architects in a variety of styles, which will be costly when creating them by hand.

\subsection{Production Tools for Entertainment Applications}

Some entertainment applications such as movies, animations and games are probably the most application forms of NST. For example, creating an animation usually requires $8$ to $24$ painted frames per second. The production costs will be largely reduced if NST can be applied to automatically stylise a live-action video into an animation style. Similarly, NST can significantly save time and costs when applied to the creation of some movies and computer games.

There are already some application papers aiming at introducing how to apply NST for production, \eg, Joshi \etal explore the use of NST in redrawing some scenes in a movie named \emph{Come Swim} \cite{joshi2017bringing}, which indicates the promising potential applications of NST in this field. In \cite{fivser2016stylit}, Fi{\v{s}}er \etal study an illumination-guided style transfer algorithm for stylisation of 3D renderings. They demonstrate how to exploit their algorithm for rendering previews on various geometries, autocomplete shading, and transferring style without a reference 3D model.

%to preview a rendering style on geometri

%Fi{\v{s}}er \etal
%illumination guided as shown by Fišer et al. [2016] for stylized 3D models

%%%%%%%%%%%%%%%%%%%%%%%%%%%%
%%%%%%%%%%%%%%%%%%%%%%%%%%%%

\section{Future Challenges}
\label{sect:challenges}

The advances in the field of NST are inspiring and some algorithms have already found use in industrial applications. Although current algorithms are capable of good performance, there are still several challenges and open issues. In this section, we summarise key challenges within this field of NST and discuss possible strategies on how to deal with them in future works. Since NST is very related to NPR, some critical problems in NPR (summarised in \cite{rosin2012image,salesin2002non,gooch2010viewing,decarlo2010visual,hertzmann2010non,kyprianidis2013state}) also remain future challenges for the research of NST. Therefore, we first review some of the major challenges existing in both NPR and NST and then discuss the research questions specialised for the field of NST.

%we first review some of the challenges existing
%cannot control the brush size in some pointsdiscuss their corresponding possible solutions.

%\stdcomment{
%Mention what does current methods cannot do.
%}

%\begin{itemize}
%  \item control orientation for oil painting
%  \item Need the user to justify whether this style and this content can generate good result and which ratio is good. Need some evaluation criterion to select automatically
%  \item for fast, training model is not stable
%  \item quality issue
%\end{itemize}

%\cite{Johnson2016perceptual,ulyanov2016texture,li2016precomputed,ulyanov2016instance}
%  \item faster and flexible \cite{dumoulin2016learned,chen2016fast}

%\subsection{Challenges}

%%%%

\subsection{Evaluation Methodology}

%%%%%%%%%%
\begin{figure}[!tbp]

\begin{tabular}{c c c}
%\begin{tabular}{m{\y} >{\centering}m{\z} >{\centering}m{\z} >{\centering}m{\z} >{\centering}m{\z} >{\centering}m{\z} >{\centering}m{\z} >{\centering}m{\z} >{\centering\arraybackslash}m{\z}}
\includegraphics[width=0.122\textwidth]{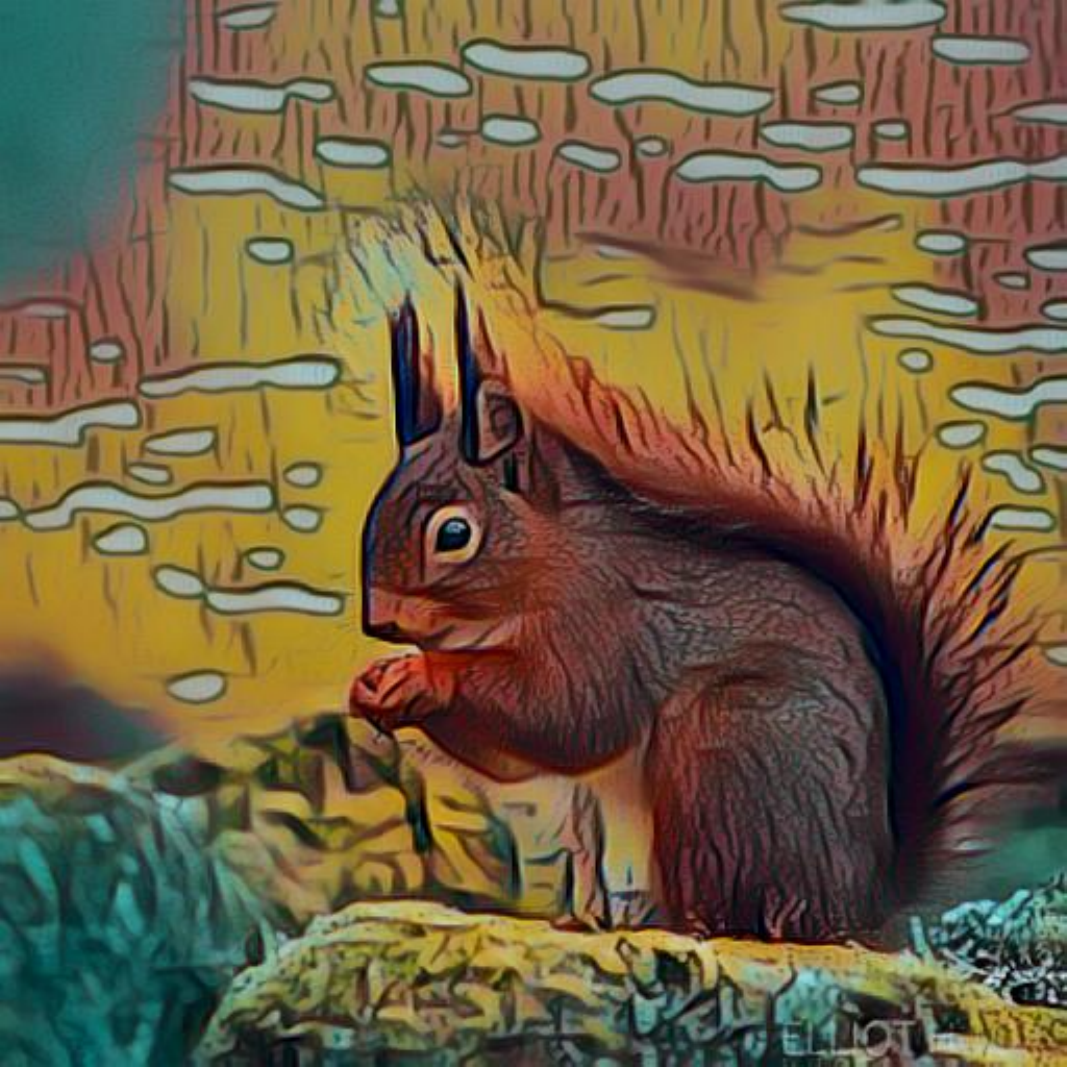}& \includegraphics[width=0.122\textwidth]{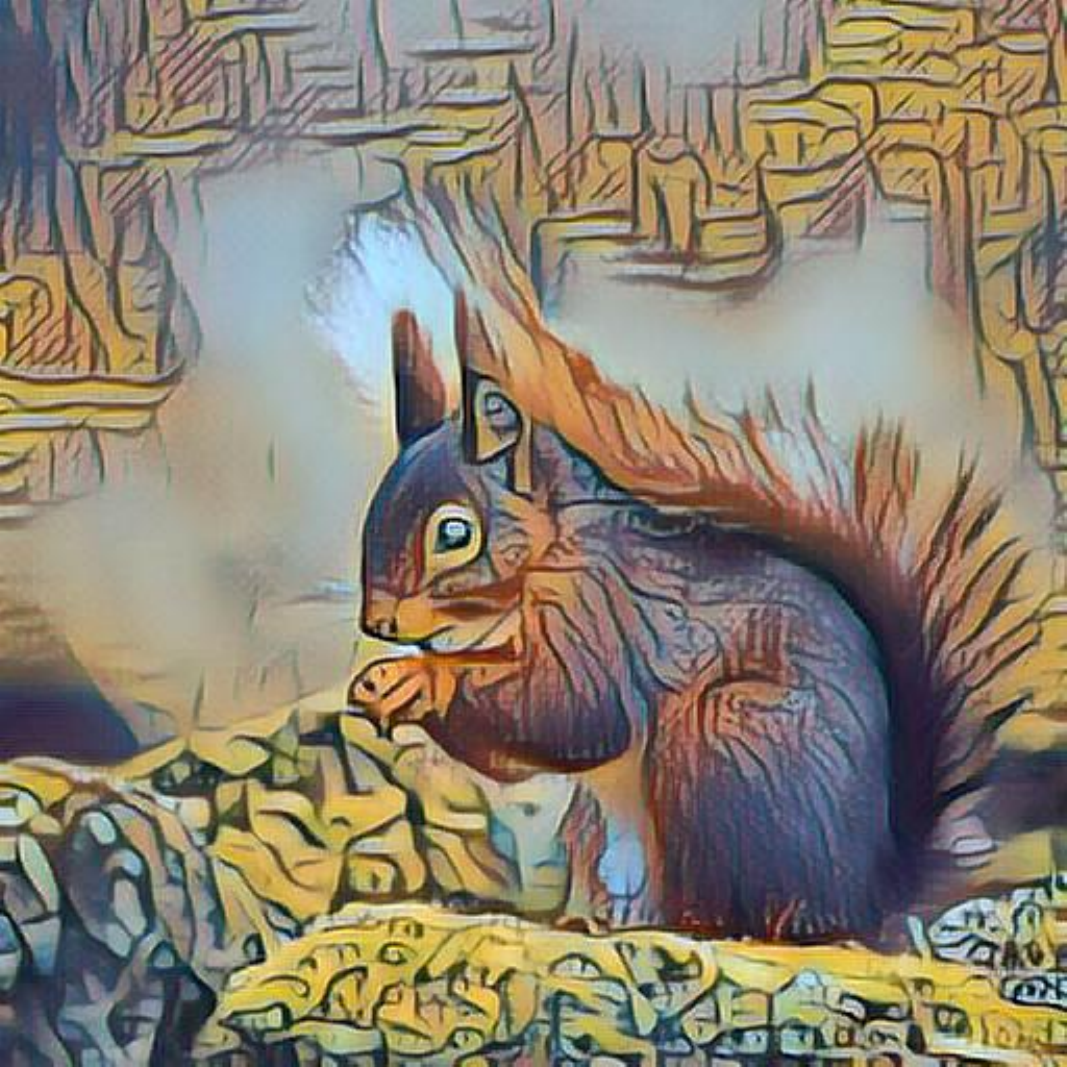}& \includegraphics[width=0.122\textwidth]{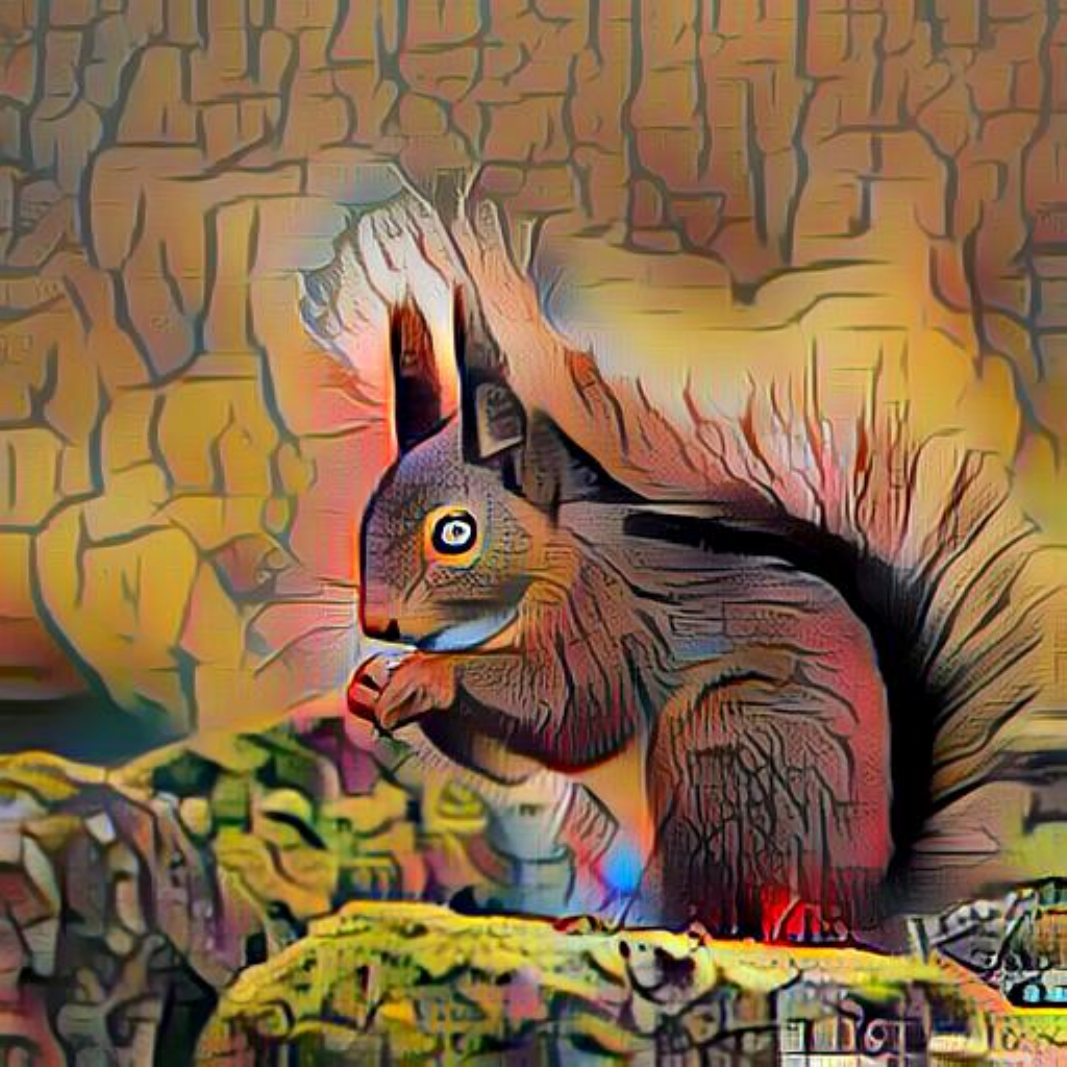}

\\

\includegraphics[width=0.13\textwidth]{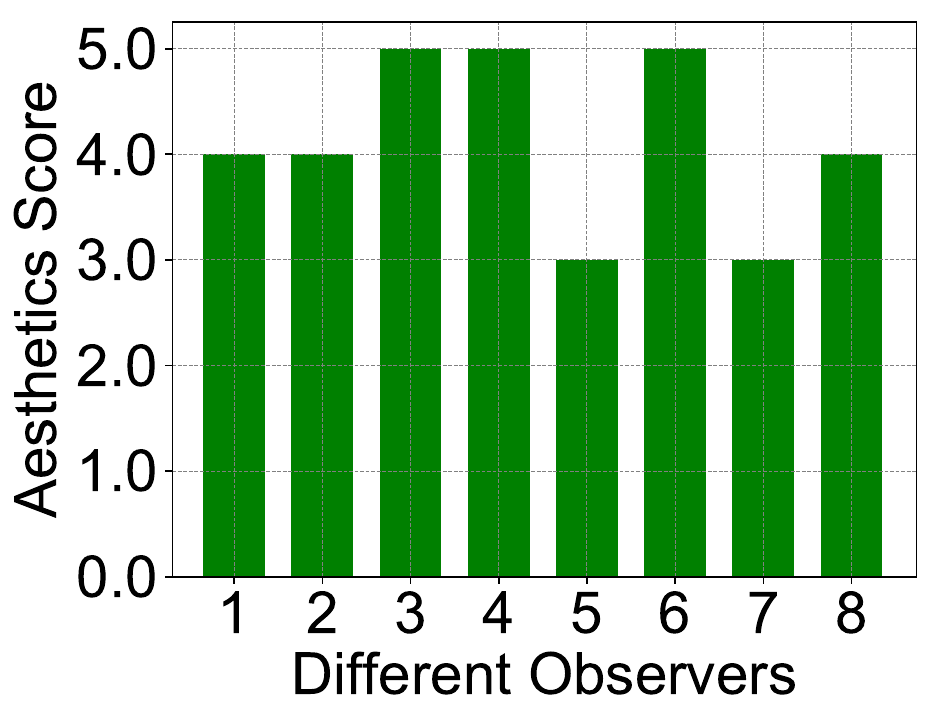}& \includegraphics[width=0.13\textwidth]{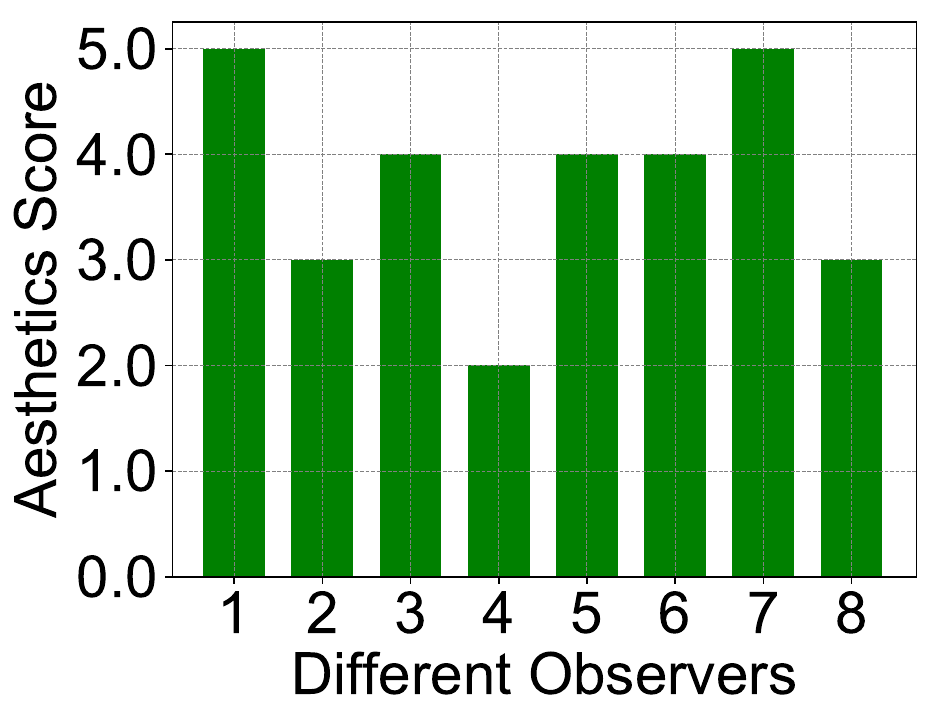}&
\includegraphics[width=0.13\textwidth]{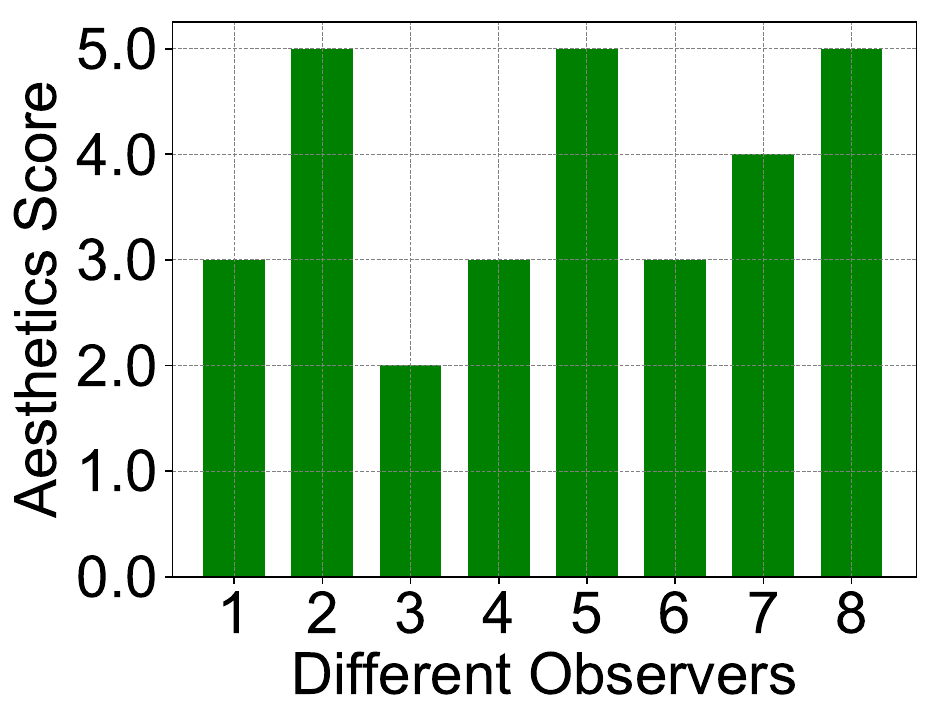}

\end{tabular}

\caption{Example of aesthetic preference scores for the outputs of different algorithms given the same style and content.}

\label{fig:aesthetic} %% label for entire figure
\end{figure}
%%%%%%%%%

Aesthetic evaluation is a critical issue in both NPR and NST. In the field of NPR, the necessity of aesthetic evaluation is explained by many researchers \cite{rosin2012image,salesin2002non,gooch2010viewing,decarlo2010visual,hertzmann2010non,kyprianidis2013state}, \eg, in \cite{rosin2012image}, Rosin and Collomosse use two chapters to explore this issue. This problem is increasingly critical as the fields of NPR and NST mature. As pointed out in \cite{rosin2012image}, researchers need some reliable criteria to assess the benefits of their proposed approach over the prior art and also a way to evaluate the suitability of one particular approach to one particular scenario. However, most NPR and NST papers evaluate their proposed approach with side-by-side subjective visual comparisons, or through measurements derived from various user studies \cite{mould2014authorial,isenberg2006non,li2017universal}. For example, to evaluate the proposed universal style transfer algorithm, Li \etal \cite{li2017universal} conduct a user study which is to ask participants to vote for their favourite stylised results. We argue that it is not an optimal solution since the results vary a lot with different observers. Inspired by \cite{ren2017personalized}, we conduct a simple experiment for user studies with the stylised results of different NST algorithms. In our experiment, each stylised image is rated by 8 different raters (4 males and 4 females) with the same occupation and age. As depicted in Figure~\ref{fig:aesthetic}, given the same stylised result, different observers with the same occupation and age still have quite different ratings. Nevertheless, there is currently no gold standard evaluation method for assessing NPR and NST algorithms. This challenge of aesthetic evaluation will continue to be an open question in both NPR and NST communities, the solution of which might require the collaboration with professional artists and the efforts in the identification of underlying aesthetic principles.

%Salesin \emph{Artistic Turing Test}
%
%Hertzmann
%
%Gooch

%There is no gold standard method for evaluation.

%by , \cite{gooch2010viewing} by Gooch \etal, \cite{decarlo2010visual} by DeCarlo and Stone, \cite{hertzmann2010non} by Hertzmann, \cite{kyprianidis2013state}

In the field of NST, there is another important issue related to aesthetic evaluation. Currently, there is no standard benchmark image set for evaluating NST algorithms. Different authors typically use their own images for evaluation. In our experiment, we use the carefully selected NPR benchmark image set named \emph{NPRgeneral} \cite{mould2016benchmark,mould2017developing} as our content images to compare different techniques, which is backed by the comprehensive study in \cite{mould2016benchmark,mould2017developing}; however, we have to admit that the selection of our style images is far from being a standard NST benchmark style set. Different from NPR, NST algorithms do not have explicit restrictions on the types of style images. Therefore, to compare the style scalability of different NST methods, it is critical to seek a benchmark style set which collectively exhibits a broad range of possible properties, accompanied by a detailed description of adopted principles, numerical measurements of image characteristics as well as a discussion of limitations like the works in \cite{mould2016benchmark,mould2017developing,rosin2017benchmarking}. Based on the above discussion, seeking an NST benchmark image set is quite a separate and important research direction, which provides not only a way for researchers to demonstrate the improvement of their proposed approach over the prior art, but also a tool to measure the suitability of one particular NST algorithm to one particular requirement. In addition, as the emergence of several NST extensions (Section~\ref{sect:improvementextensions}), it remains another open problem to study the specialised benchmark data set and also the corresponding evaluation criteria for assessing those extended works (\eg, video style transfer, audio style transfer, stereoscopic style transfer, character style transfer and fashion style transfer).

% so as  Establishing an NST benchmark style set should also
%
%Establishing a benchmark style set for NST requires a very careful selection of style images, accompanied by a detailed description of adopted principles, numerical measurements of image characteristics as well as a discussion of limitations like the works in \cite{mould2016benchmark,mould2017developing,rosin2017benchmarking}.

%In addition, for video and ste style transfer,

%Mould and Rosin
%dataset
%
%
%benckmark dataset video comparison
%
%far from a ..

%\subsection{Image Abstraction}
%
%Photorealistic rendering for sketches
%
%NST works well for artistic styles
%but not for   Low Poly rendering Pix Art Rendering
%
%\cite{gerstner2012pixelated} \cite{zhang2015low,gai2016artistic}

\subsection{Interpretable Neural Style Transfer}

Another challenging problem is the interpretability of NST algorithms. Like many other CNN-based vision tasks, the process of NST is like a black box, which makes it quite uncontrollable. In this part, we focus on three critical issues related to the interpretability of NST, \ie, interpretable and controllable NST via disentangled representations, normalisation methods associated with NST, and adversarial examples in NST.

\textbf{Representation disentangling.} The goal of representation disentangling is to learn dimension-wise interpretable representations, where some changes in one or more specific dimensions correspond to changes precisely in a single factor of variation while being invariant to other factors \cite{bengio2013representation,kimmnih16,feng2018dual,feng2018interpretable}. Such representations are useful to a variety of machine learning tasks, \eg, visual concepts learning \cite{higgins2017scan} and transfer learning \cite{lake2017building}. For example, in style transfer, if one could learn a representation where the factors of variation (\eg, colour, shape, stroke size, stroke orientation and stroke composition) are precisely disentangled, these factors could then be freely controlled during stylisation. For example, one could change the stroke orientations in a stylised image by simply changing the corresponding dimension in the learned disentangled representation. Towards the goal of disentangled representation, current methods fit into two categories, which are supervised approaches and unsupervised ones. The basic idea of supervised disentangling methods is to exploit annotated data to supervise the mapping between inputs and attributes \cite{AAAI1816521,wang2017tag}. Despite their effectiveness, supervised disentangling approaches typically require numbers of training samples. However, in the case of NST, it is quite complicated to model and capture some of those aforementioned factors of variation. For example, it is hard to collect a set of images which have different stroke orientations but exactly the same colour distribution, stroke size and stroke composition. By contrast, unsupervised disentangling methods do not require annotations; however, they usually yield disentangled representations which are dimension-wise uncontrollable and uninterpretable \cite{higgins2016beta}, \ie, we could not control what would be encoded in each specific dimension. Based on the above discussion, to acquire disentangled representations in NST, the first issue to be addressed is how to define, model and capture the complicated factors of variation in NST.

\begin{table}
\renewcommand\arraystretch{1.8}
\caption{Normalisation methods in NST.}
%\begin{minipage}{\columnwidth}
\begin{center}
%\begin{tabular}{  |c | p{3.33cm}| p{3.9cm}|  }
\begin{tabular}{  |c |l|l|}
    %\toprule[1pt]
    %\hline
    \hline
    \textbf{Paper} &  \textbf{Author}&  \textbf{Name}\\
    \hline
    \cite{ulyanov2017improved} & Ulyanov \etal & \emph{Instance Normalisation}\\
    \cite{dumoulin2016learned} & Dumoulin \etal & \emph{Conditional Instance Normalisation}\\
    \cite{huang2017arbitrary} & Huang and Belongie & \emph{Adaptive Instance Normalisation}\\
    %\cite{li2017universal} & Li \etal & \emph{Whitening Instance Normalisation} \\
    \hline
    %\midrule[1pt]
     \end{tabular}
\end{center}
%\bigskip\centering
\footnotesize
%\begin{flushleft}
\smallskip

%\end{flushleft}
\label{table:normalization}
%\end{minipage}
\end{table}
%%%%%%%%%%%%%%

\textbf{Normalisation methods.} The advances in the field of NST are closely related to the emergence of novel normalisation methods, as shown in Table~\ref{table:normalization}. Some of these normalisation methods also have an influence on a larger vision community beyond style transfer (\eg, image recolourisation \cite{cho2017palettenet} and video colour propagation \cite{meyer2018deep}). In this part, we first briefly review these normalisation methods in NST and then discuss the corresponding problem.
The first emerged normalisation method in NST is \emph{instance normalisation} (or \emph{contrast normalisation}) proposed by Ulyanov \etal \cite{ulyanov2017improved}. \emph{Instance normalisation} is equivalent to \emph{batch normalisation} when the batch size is one. It is shown that style transfer network with \emph{instance normalisation} layer converges faster and produces visually better results compared with the network with \emph{batch normalisation} layer. Ulyanov \etal believe that the superior performance of \emph{instance normalisation} results from the fact that \emph{instance normalisation} enables the network to discard contrast information in content images and therefore makes learning simpler. Another explanation proposed by Huang and Belongie \cite{huang2017arbitrary} is that \emph{instance normalisation} performs a kind of \emph{style normalisation} by normalising feature statistics (\ie, the mean and variance). With \emph{instance normalisation}, the style of each individual image could be directly normalised to the target style. As a result, the rest of the network only needs to take care of the content loss, making the objective easier to learn. Based on \emph{instance normalisation}, Dumoulin \etal \cite{dumoulin2016learned} further propose \emph{conditional instance normalisation}, which is to scale and shift parameters in \emph{instance normalisation} layers (shown in Equation~(\ref{eq:CIN})). Following the interpretation proposed by Huang and Belongie, by using different affine parameters, the feature statistics could be normalised to different values. Correspondingly, the style of each individual sample could be normalised to different styles.
Furthermore, in \cite{huang2017arbitrary}, Huang and Belongie propose \emph{adaptive instance normalisation} to adaptively instance normalise content feature by the style feature statistics (shown in Equation~(\ref{eq:AdaIN})). In this way, they believe that the style of an individual image could be normalised to arbitrary styles.
Despite the superior performance achieved by \emph{instance normalisation}, \emph{conditional instance normalisation} and \emph{adaptive instance normalisation}, the reason behind their success still remains unclear. Although Ulyanov \etal \cite{ulyanov2017improved} and Huang and Belongie \cite{huang2017arbitrary} propose their own hypothesis based on pixel space and feature space respectively, there is a lack of theoretical proof for their proposed theories. In addition, their proposed theories are also built on other hypothesises, \eg, Huang and Belongie propose their interpretation based on the observation by Li \etal \cite{li2017demystifying}: channel-wise feature statistics, namely mean and variance, could represent styles. However, it remains uncertain why feature statistics could represent the style, or even whether the feature statistics could represent all styles, which relates back to the interpretability of style representations.

% However, the reason behind the success of
%which relates back to the interpretation of style representations.
%
%Despite the explanation proposed , the reason behind
%
%there is no theoretical proof for

%, normalise an individual sample to different styles by normalising feature statistics to different values.

%Equation~(\ref{eq:CIN})

%Equation~(\ref{eq:AdaIN})
%\emph{instance normalisation} is also called \emph{contrast normalisation} by some

%Huang
%
%介绍一遍
%说明challenge是啥
%
%There is no theoretical proof
%
%There is well-known normalisation methods as shown in Table~\ref{table:normalization}.
%
%
%Like batch normalisation 不一定适合所有场合，

%
%Normalisation
%Disentangling:flexible control stroke orientation control

\begin{figure}[!t]
\setlength\tabcolsep{1.5 pt}
{\renewcommand{\arraystretch}{0.8}
%\begin{tabular}{>{\centering}n{\p} >{\centering}n{\p} >{\centering\arraybackslash}n{\p}}
%\begin{tabular}{>{\centering}m{2cm} >{\centering}m{2cm} >{\centering}m{2cm} >{\centering\arraybackslash}m{2cm}}
\begin{tabular}{cccc}
\centering

\includegraphics[width=0.116\textwidth]{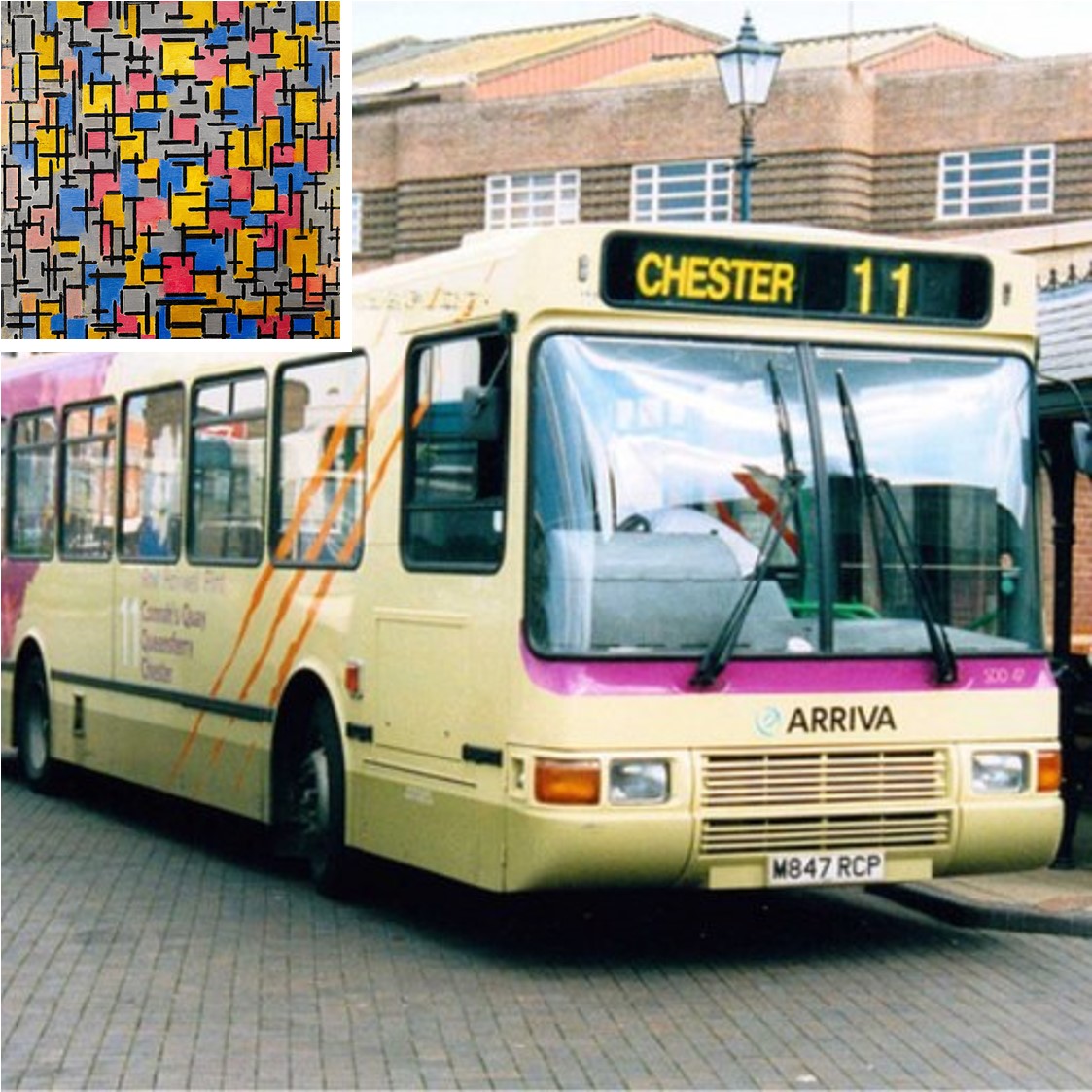}&\includegraphics[width=0.116\textwidth]{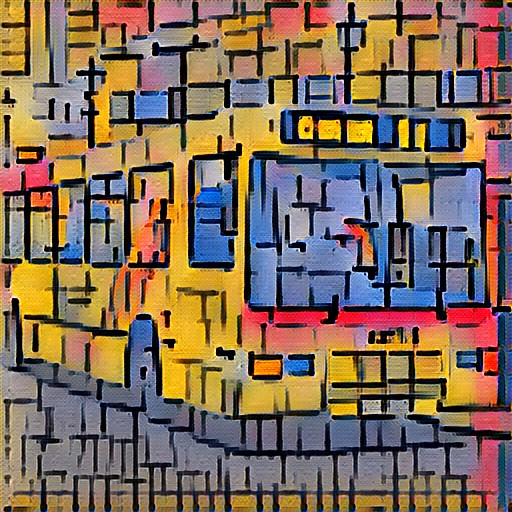}& \includegraphics[width=0.116\textwidth]{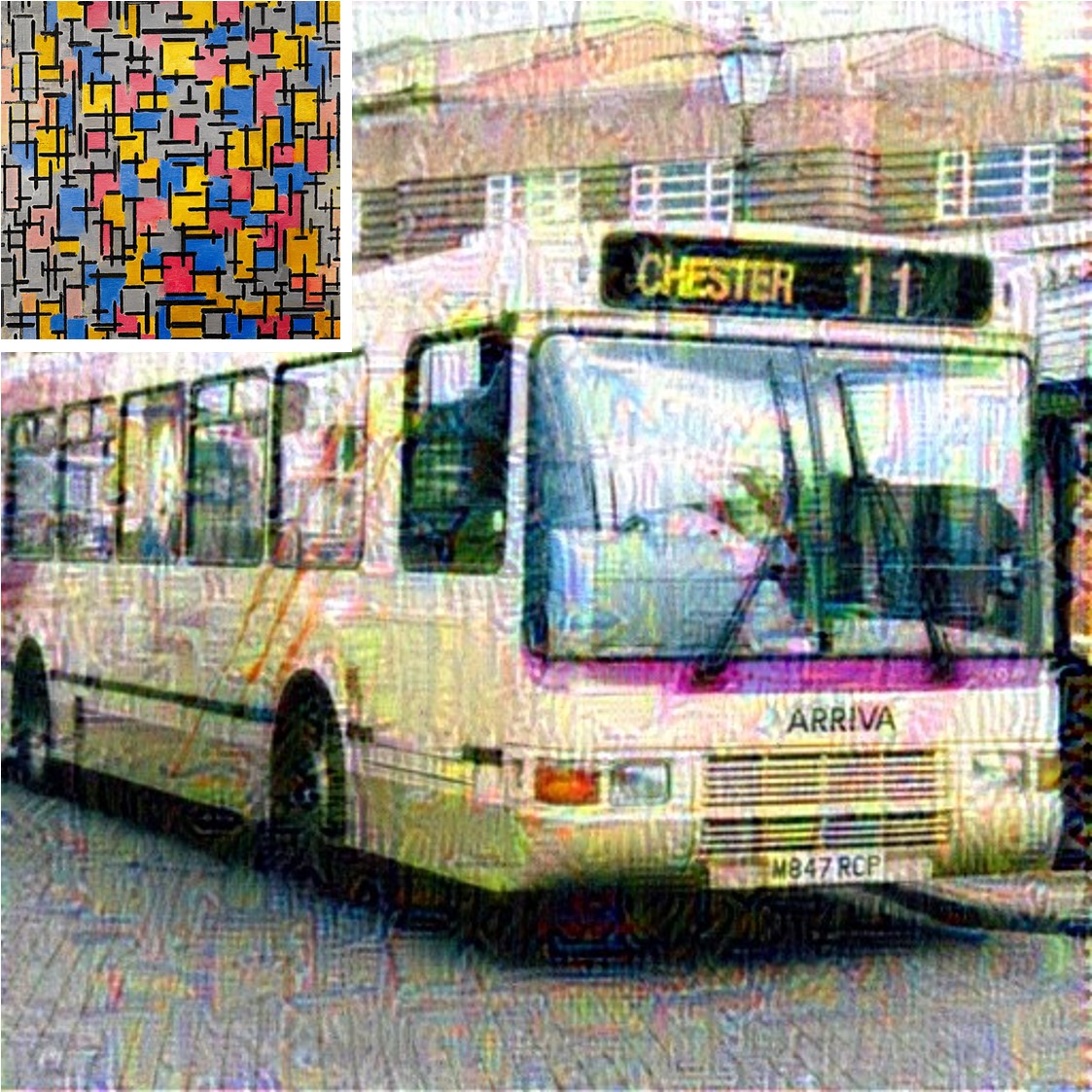} & \includegraphics[width=0.116\textwidth]{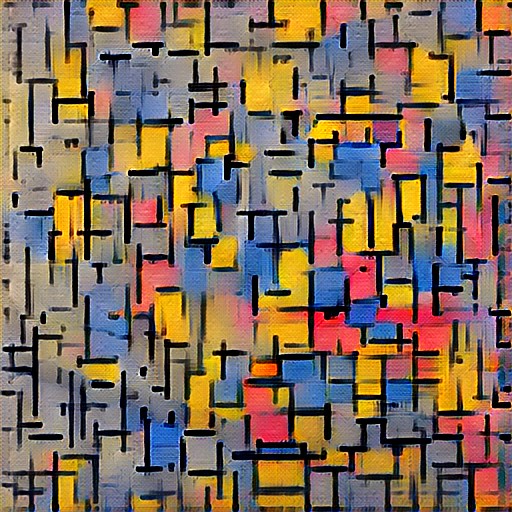} \\

\footnotesize{(a)}&\footnotesize{(b)} &\footnotesize{(c)} & \footnotesize{(d)}
\end{tabular}
}
%\smallskip
\caption{Adversarial example for NST: (a) is the original content and style image pair and (b) is the stylised result of (a) with \cite{Johnson2016perceptual}; (c) is the generated adversarial example and (d) is the stylised result of (c) with the same model as (b).}

\label{fig:adversarial} %% label for entire figure
\end{figure}

\textbf{Adversarial examples.} Several studies have shown that deep classification networks are easily fooled by \emph{adversarial examples} \cite{szegedy2013intriguing,goodfellow2014explaining}, which are generated by applying perturbations to input images (\eg, Figure~\ref{fig:adversarial}(c)). Previous studies on adversarial examples mainly focus on deep classification networks. However, as shown in Figure~\ref{fig:adversarial}, we find that adversarial examples also exist in generative style transfer networks. In Figure~\ref{fig:adversarial}(d), one can hardly recognise the content, which is originally contained in Figure~\ref{fig:adversarial}(c). It reveals the difference between generative networks and the human vision system. The perturbed image is still recognisable to humans but leads to a different result for generative style transfer networks. However, it remains unclear why some perturbations could make such a difference, and whether some similar noised images uploaded by the user could still be stylised into the desired style. Interpreting and understanding adversarial examples in NST could help to avoid some failure cases in stylisation.

%The emergence of adversarial examples reveals the difference between deep neural networks and human vision system. The perturbed result by changing an originally correctly classified image is still recognisable to humans, but leads to a misclassified label for deep neural networks.

\subsection{Three-way Trade-off in Neural Style Transfer}

In the field of NST, there is a three-way trade-off between speed, flexibility and quality. IOB-NST achieves superior performance in quality but is computationally expensive. PSPM-MOB-NST achieves real-time stylisation; however, PSPM-MOB-NST needs to train a separate network for each style, which is not flexible. MSPM-MOB-NST improves the flexibility by incorporating multiple styles into one single model, but it still needs to pre-train a network for a set of target styles. Although ASPM-MOB-NST algorithms successfully transfer arbitrary styles, they are not that satisfying in perceptual quality and speed. The quality of data-driven ASPM quite relies on the diversity of training styles. However, one can hardly cover every style due to the great diversity of artworks. Image transformation based ASPM algorithm transfers arbitrary styles in a learning-free manner, but it is behind others in speed. Another related issue is the problem of hyperparameter tuning. To produce the most visually appealing results, it remains uncertain how to set the value of content and style weights, how to choose layers for computing content and style loss, which optimiser to use and how to set the value of learning rate. Currently, researchers empirically set these hyperparameters; however, one set of hyperparameters does not necessarily work for any style and it is tedious to manually tune these parameters for each combination of content and style images. One of the keys for this problem is a better understanding of the optimisation procedure in NST. A deep understanding of optimisation procedure would help understand how to find the local minima that lead to a high quality.

\section{Discussions and Conclusions}
\label{sect:conclusion}

Over the past several years, NST has continued to become an inspiring research area, motivated by both scientific challenges and industrial demands. A considerable amount of researches have been conducted in the field of NST. Key advances in this field are summarised in Figure~\ref{fig:taxonomy}. A summary of the corresponding style transfer loss functions can be found in Table~\ref{table:lossfunction}. NST is quite a fast-paced area, and we are looking forwarding to more exciting works devoted to advancing the development of this field.

%But to our knowledge, recently there are many researches \cite{shen2018neural,gu2018arbitrary,sheng2018neural,zhang2018neural} that are just announced for publication but without a final PDF version yet. Therefore, we do not include them in our current version.

%\subsection{Overview of Style Transfer Loss Functions}

\begin{table*}
\renewcommand\arraystretch{1.3}
\caption{An overview of major style transfer loss functions.}
%\begin{minipage}{\columnwidth}
\begin{center}
\begin{footnotesize}
\begin{tabular}{  | p{3.3cm}| p{4cm}| p{9.6cm}| }
%\begin{tabular}{  |c |l|l|l|}
    %\toprule[1pt]
    %\hline
    \hline
    \textbf{Paper} &  \textbf{Loss} & \textbf{Description}\\
    \hline
      Gatys \etal \cite{gatys2016image}& \emph{Gram Loss} &  The first proposed style loss based on Gram-based style representations. \\ \hline
      Johnson \etal \cite{Johnson2016perceptual}  & \emph{Perceptual Loss} & Widely adopted content loss based on perceptual similarity. \\ \hline
      \multirow{3}{*}{Berger and Memisevic \cite{berger2016incorporating}} & \multirow{3}{*}{\emph{Transformed Gram Loss}} & Computing \emph{Gram Loss} over horizontally and vertically translated feature representations. More effective at modelling style with symmetric properties, compared with \emph{Gram Loss}.\\ \hline
      \multirow{3}{*}{Li \etal \cite{li2017diverse}}& \multirow{3}{*}{\emph{Mean-substraction Gram Loss}} & Subtracting the mean of feature representations before computing \emph{Gram Loss}. Eliminating large discrepancy in
        scale. Effective at multi-style transfer with one single network. \\ \hline
      \multirow{2}{*}{Zhang and Dana \cite{zhang2017multi}}& \multirow{2}{*}{\emph{Multi-scale Gram Loss}} & Computing \emph{Gram Loss} over multi-scale feature representations. Eliminating a few artefacts. \\ \hline
      \multirow{3}{*}{Li \etal \cite{li2017demystifying}}& \multirow{3}{*}{\emph{MMD Loss with Different Kernels}}& \emph{Gram Loss} is equivalent to \emph{MMD Loss with Second Order Polynomial Kernel}. \emph{MMD Loss with Linear Kernel} is capable of comparable quality with \emph{Gram Loss}, but with lower computational complexity.\\ \hline
      \multirow{2}{*}{Li \etal \cite{li2017demystifying}}& \multirow{2}{*}{\emph{BN Loss}}& Achieving comparable quality with \emph{Gram Loss}, but conceptually clearer in theory.  \\ \hline
      \multirow{2}{*}{Risser \etal \cite{wilmot2017stable}}& \multirow{2}{*}{\emph{Histogram Loss}} & Matching the entire histogram of feature representations. Eliminating instability artefacts, compared with single \emph{Gram Loss}.\\ \hline
      Li \etal \cite{li2017laplacian}& \emph{Laplacian Loss} & Eliminating distorted structures and irregular artefacts. \\ \hline
      \multirow{2}{*}{Li and Wand \cite{li2016combining}}& \multirow{2}{*}{\emph{MRF Loss}}& More effective when the content and style are similar in shape and perspective, compared with \emph{Gram Loss}. \\ \hline
      \multirow{2}{*}{Champandard \cite{champandard2016semantic}}& \multirow{2}{*}{\emph{Semantic Loss}} & Incorporating a segmentation mask over \emph{MRF Loss}. Enabling a more accurate semantic match.  \\  \hline
      \multirow{3}{*}{Li and Wand \cite{li2016precomputed}}&\multirow{3}{*}{\emph{Adversarial Loss}} & Computed based on PatchGAN. Utilising contextual correspondence between patches. More effective at preserving coherent textures in complex images, compared with \emph{Gram Loss}. \\ \hline
      Jing \etal \cite{jing2018stroke}& \emph{Stroke Loss} & Achieving continuous stroke size control while preserving stroke consistency. \\ \hline
      \multirow{2}{*}{Wang \etal \cite{wang2016multimodal}}& \multirow{2}{*}{\emph{Hierarchical Loss}} & Enabling a coarse-to-fine stylisation procedure. Capable of producing large but also subtle strokes for high-resolution content images.\\ \hline
      \multirow{2}{*}{Liu \etal \cite{liu2017depthaware}}& \multirow{2}{*}{\emph{Depth Loss}} & Preserving depth maps of content images. Effective at retaining spatial layout and structure of content images, compared with single \emph{Gram Loss}. \\ \hline
      \multirow{3}{*}{Ruder \etal \cite{ruder2016artistic}}& \multirow{3}{*}{\emph{Temporal Consistency Loss}} & Designed for video style transfer. Penalising the deviations along point trajectories based on optical flow. Capable of maintaining temporal consistency among stylised video frames. \\ \hline
      \multirow{2}{*}{Chen \etal \cite{chen2018stereoscopic}} & \multirow{2}{*}{\emph{Disparity Loss}} & Designed for stereoscopic style transfer. Penalising bidirectional disparity. Capable of consistent strokes for different views. \\ \hline

    %\midrule[1pt]
     \end{tabular}
     \end{footnotesize}
\end{center}
%\bigskip\centering
\footnotesize
%\begin{flushleft}
\smallskip
%\end{flushleft}
\label{table:lossfunction}
%\end{minipage}
\end{table*}

%\subsection{Significance for the Larger Vision Community}

%\cite{he2018deep,fan2018decouple}

During the period of preparing this review, we are also delighted to find that related researches on NST also bring new inspirations for other areas \cite{ulyanov2017deep,upchurch2017deep,he2018deep,fan2018decouple,atapour2018real} and accelerate the development of a wider vision community.
%\textbf{1)}
For the area of \emph{Image Reconstruction}, inspired by NST, Ulyanov \etal \cite{ulyanov2017deep} propose a novel deep image prior, which replaces the manually-designed total variation regulariser in \cite{mahendran2015understanding} with a randomly initialised deep neural network. Given a task-dependent loss function $\mathcal{L}$, an image $I_o$ and a fixed uniform noise $z$ as inputs, their algorithm can be formulated as:
\begin{equation}
\theta^*=\mathop {\arg \min }\limits_{\theta} \mathcal{L}(g_{\theta^*}(z),I_o), \ I^* = g_{\theta^*}(z).
\label{eq:deepprior}
\end{equation}
One can easily notice that Equation (\ref{eq:deepprior}) is very similar to Equation (\ref{eq:fastloss}). The process in \cite{ulyanov2017deep} is equivalent with the training process of MOB-NST when there is only one available image in the training set, but replacing $I_c$ with $z$ and $\mathcal{L}_{total}$ with $\mathcal{L}$. In other words, $g$ in \cite{ulyanov2017deep} is trained to overfit one single sample. %\\
%\textbf{2)}
Inspired by NST, Upchurch \etal \cite{upchurch2017deep} propose a deep feature interpolation technique and provide a new baseline for the area of \emph{Image Transformation} (\eg, face aging and smiling). Upon the procedure of IOB-NST algorithm \cite{gatys2016image}, they add an extra step which is interpolating in the VGG feature space. In this way, their algorithm successfully changes image contents in a learning-free manner. %\\
%\textbf{3)}
Another field closely related to NST is \emph{Face Photo-sketch Synthesis}. For example, \cite{chen2018face} exploits style transfer to generate shadings and textures for final face sketches. Similarly, for the area of \emph{Face Swapping}, the idea of MOB-NST algorithm \cite{ulyanov2016texture} can be directly applied to build a feed-forward \emph{Face-Swap} algorithm \cite{korshunova2017fast}. %\\
%Similarly, Korshunova \etal \cite{korshunova2017fast} apply the idea of feed-forward Fast NST algorithm \cite{ulyanov2016texture}, NST \cite{korshunova2017fast} Fast \emph{Face-Swap}  \\
%\textbf{4)}
NST also provides a new way for \emph{Domain Adaption}, as is validated in the work of Atapour-Abarghouei and Breckon \cite{atapour2018real}. They apply style transfer technique to translate images from different domains so as to improve the generalisation capabilities of their \emph{Monocular Depth Estimation} model.
%\cite{atapour2018real} depth estimation single image depth estimation?
%There are also some works closely related to Neural Style Transfer, \cite{upchurch2017deep}\cite{ulyanov2017deep}
%\textbf{An Overview of Style Transfer Loss Functions}
%Gram
%Gram average
%histogram
%BN ... polynomial linear ...
%markov
%semantic
%local adversarial
%depth

%The area is so fast many papers just anounced and without pdf need to be added many arbitrary style algorithms

%only review to time 2018.3. the following papers will be subsequatially added

%other ASPM \cite{shen2018neural,gu2018arbitrary,sheng2018neural,zhang2018neural}

%In conclusion, this paper gives the first comprehensive survey of past developments in Neural Style Transfer, covering the taxonomy of current methods, their improvements and extensions, evaluation methodology as well as existing challenges and corresponding possible solutions. Moreover, three application domains of Neural Style Transfer are reviewed, including social communication, user-assisted creation tools and production tools for entertainment applications.

%\section{Future Work}
%\label{sect:futurework}

Despite the great progress in recent years, the area of NST is far from a mature state. Currently, the first stage of NST is to refine and optimise recent NST algorithms, aiming to perfectly imitate varieties of styles. This stage involves two technical directions. The first one is to reduce failure cases and improve stylised quality on a wider variety of style and content images.
Although there is not an explicit restriction on the type of styles, NST does have styles it is particularly good at and also some certain styles it is weak in. For example, NST typically performs well in producing irregular style elements (\eg, paintings), as demonstrated in many NST papers \cite{gatys2016image,Johnson2016perceptual,dumoulin2016learned,li2017universal}; however, for some styles with regular elements such as low-poly styles \cite{zhang2015low,gai2016artistic} and pixelator styles \cite{gerstner2012pixelated}, NST generally produces distorted and irregular results due to the property of CNN-based image reconstruction.
For content images, previous NST papers usually use natural images as content to demonstrate their proposed algorithms; however, given abstract images (\eg, sketches and cartoons) as input content, NST typically does not combine enough style elements to match the content \cite{huang2018multimodal}, since a pre-trained classification network could not extract proper image content from these abstract images.
The other technical direction of the first stage lies in deriving more extensions from general NST algorithms. For example, as the emergence of 3D vision techniques, it is promising to study 3D surface stylisation, which is to directly optimise and produce 3D objects for both photorealistic and non-photorealistic stylisation.
After moving beyond the first stage, a further trend of NST is to not just imitate human-created art with NST techniques, but rather to create a new form of AI-created art under the guidance of underlying aesthetic principles.
The first step towards this direction has been taken, \ie, using current NST methods \cite{chen2017stylebank,dumoulin2016learned,wang2016multimodal} to combine different styles. For example, in \cite{wang2016multimodal}, Wang \etal successfully utilise their proposed algorithm to produce a new style which fuses the coarse texture distortions of one style with the fine brush strokes of another style image.

\ifCLASSOPTIONcaptionsoff
  \newpage
\fi

%%%%%%%%%%%%%%%%%%%%%%%%%%%%%%%%%%%%%%%%%%%%%%%%%%%%%%%%%%%%%%%%%%%%%%%%%%%%%%%%%%%%%%%%%%%%%%%%%%%
%{\small\bibliographystyle{ieee}\bibliography{MYRE.bib}}

\bibliographystyle{IEEEtran}
\bibliography{MYRE.bib}

\end{document}